\documentclass{article} %
\usepackage{iclr2021_conference,times}

\usepackage[utf8]{inputenc} %
\usepackage[T1]{fontenc}    %
\usepackage{hyperref}       %
\usepackage{url}            %
\usepackage{booktabs}       %
\usepackage{amsfonts}       %
\usepackage{nicefrac}       %
\usepackage{microtype}      %
\usepackage{multirow}
\usepackage{adjustbox}
\usepackage{caption}
\usepackage{graphicx}
\usepackage{rotating}
\usepackage{enumitem}
\usepackage{wrapfig}
\usepackage{subcaption}
\usepackage{color,colortbl}
\graphicspath{{fig/}}

\usepackage{amsmath,amsfonts,bm}

\def\Figref#1{Figure~\ref{#1}}

\def\Secref#1{Section~\ref{#1}}

\def\eqref#1{equation~\ref{#1}}

\def\1{\bm{1}}

\DeclareMathAlphabet{\mathsfit}{\encodingdefault}{\sfdefault}{m}{sl}
\SetMathAlphabet{\mathsfit}{bold}{\encodingdefault}{\sfdefault}{bx}{n}

\newcommand{\minisection}[1]{\textbf{#1}\hspace{0.2em}}
\definecolor{Gray}{gray}{0.9}

\title{Tent: Fully Test-Time Adaptation \\ by Entropy Minimization}

\author{%
  Dequan Wang$^1$\thanks{Equal contribution. $^\dagger$Work done at Adobe Research; the author is now at DeepMind.}~,
  Evan Shelhamer$^{2*\dagger}$,
  Shaoteng Liu$^1$,
  Bruno Olshausen$^1$,
  Trevor Darrell$^1$\\
  \texttt{dqwang@cs.berkeley.edu},
  \texttt{shelhamer@google.com}\\
  UC Berkeley$^1$ {\quad} Adobe Research$^2$
}

\iclrfinalcopy
\begin{document}

\maketitle

\begin{abstract}
A model must adapt itself to generalize to new and different data during testing.
In this setting of fully test-time adaptation the model has only the test data and its own parameters.
We propose to adapt by test entropy minimization (tent\footnote{Please see the project page at \url{https://github.com/DequanWang/tent} for the code and more.}): we optimize the model for confidence as measured by the entropy of its predictions.
Our method estimates normalization statistics and optimizes channel-wise affine transformations to update online on each batch.
Tent reduces generalization error for image classification on corrupted ImageNet and CIFAR-10/100 and reaches a new state-of-the-art error on ImageNet-C.
Tent handles source-free domain adaptation on digit recognition from SVHN to MNIST/MNIST-M/USPS, on semantic segmentation from GTA to Cityscapes, and on the VisDA-C benchmark.
These results are achieved in one epoch of test-time optimization without altering training.
\end{abstract}

\section{Introduction}

Deep networks can achieve high accuracy on training and testing data from the same distribution, as evidenced by tremendous benchmark progress \citep{krizhevsky2012imagenet,simonyan2014very,he2016deep}.
However, generalization to new and different data is limited \citep{hendrycks2019benchmarking,recht2019do-imagenet,geirhos2018generalisation}.
Accuracy suffers when the training (source) data differ from the testing (target) data, a condition known as \emph{dataset shift} \citep{quionero2009dataset}.
Models can be sensitive to shifts during testing that were not known during training, whether natural variations or corruptions, such as unexpected weather or sensor degradation.
Nevertheless, it can be necessary to deploy a model on different data distributions, so adaptation is needed.

During testing, the model must adapt given only its parameters and the target data.
This \emph{fully test-time adaptation} setting cannot rely on source data or supervision.
Neither is practical when the model first encounters new testing data, before it can be collected and annotated, as inference must go on.
Real-world usage motivates fully test-time adaptation by data, computation, and task needs:

\begin{enumerate}[leftmargin=4mm]
\item Availability. A model might be distributed without source data for bandwidth, privacy, or profit.
\item Efficiency. It might not be computationally practical to (re-)process source data during testing.  %
\item Accuracy. A model might be too inaccurate without adaptation to serve its purpose.
\end{enumerate}

\begin{figure}[t]
\begin{minipage}{0.48\textwidth}
\centering
\includegraphics[width=0.95\linewidth]{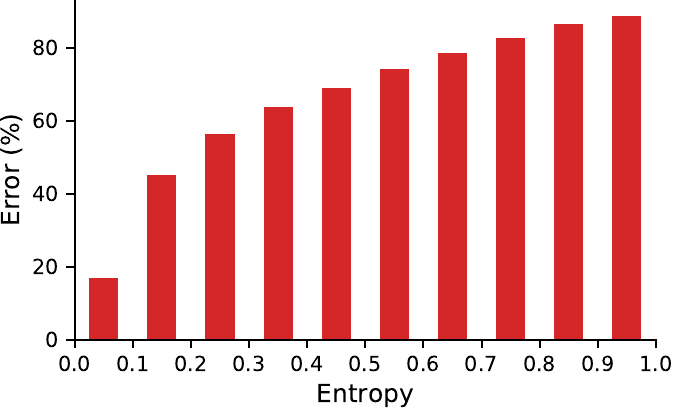}
\caption{%
Predictions with lower entropy have lower error rates on corrupted CIFAR-100-C.
Certainty can serve as supervision during testing.
}
\label{fig:error-entropy}
\end{minipage}
\hfill
\begin{minipage}{0.48\textwidth}
\centering
\vspace{-1mm}
\includegraphics[width=0.95\linewidth]{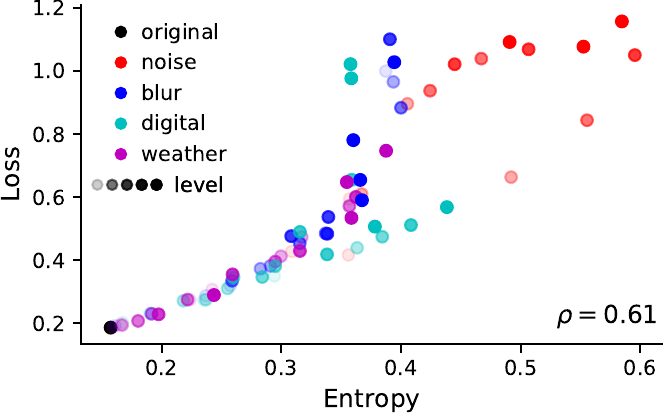}
\caption{%
More corruption causes more loss and entropy on CIFAR-100-C.
Entropy can estimate the degree of shift without training data or labels.
}
\label{fig:corrupt-loss-entropy}
\end{minipage}
\end{figure}

To adapt during testing we minimize the entropy of model predictions.
We call this objective the test entropy and name our method \emph{tent} after it.
We choose entropy for its connections to error and shift.
Entropy is related to error, as more confident predictions are all-in-all more correct (\Figref{fig:error-entropy}).
Entropy is related to shifts due to corruption, as more corruption results in more entropy, with a strong rank correlation to the loss for image classification as the level of corruption increases (\Figref{fig:corrupt-loss-entropy}).

To minimize entropy, tent normalizes and transforms inference on target data by estimating statistics and optimizing affine parameters batch-by-batch.
This choice of low-dimensional, channel-wise feature modulation is efficient to adapt during testing, even for online updates.
Tent does not restrict or alter model training: it is independent of the source data given the model parameters.
If the model can be run, it can be adapted.
Most importantly, tent effectively reduces not just entropy but error.

Our results evaluate generalization to corruptions for image classification, to domain shift for digit recognition, and to simulation-to-real shift for semantic segmentation.
For context with more data and optimization, we evaluate methods for robust training, domain adaptation, and self-supervised learning given the labeled source data.
Tent can achieve less error given only the target data, and it improves on the state-of-the-art for the ImageNet-C benchmark.
Analysis experiments support our entropy objective, check sensitivity to the amount of data and the choice of parameters for adaptation, and back the generality of tent across architectures.

\minisection{Our contributions}
\begin{itemize}[leftmargin=4mm]
\item We highlight the setting of fully test-time adaptation with only target data and no source data.
To emphasize practical adaptation during inference we benchmark with offline and online updates.
\item We examine entropy as an adaptation objective and propose tent: a test-time entropy minimization scheme to reduce generalization error by reducing the entropy of model predictions on test data.
\item For robustness to corruptions, tent reaches $44.0\%$ error on ImageNet-C, better than the state-of-the-art for robust training ($50.2\%$) and the strong baseline of test-time normalization ($49.9\%$).
\item For domain adaptation, tent is capable of online and source-free adaptation for digit classification and semantic segmentation, and can even rival methods that use source data and more optimization. %
\end{itemize}

\section{Setting: Fully Test-Time Adaptation}
\label{sec:ftta}

Adaptation addresses generalization from source %
to target. %
A model $f_\theta(x)$ with parameters $\theta$ trained on source data and labels $x^s, y^s$ %
may not generalize when tested on shifted target data $x^t$. %
Table \ref{tab:settings} summarizes adaptation settings, their required data, and types of losses.
Our fully test-time adaptation setting uniquely requires only the model $f_\theta$ and unlabeled target data $x^t$ for adaptation during inference.

Existing adaptation settings extend training given more data and supervision.
Transfer learning by fine-tuning \citep{donahue2014decaf,yosinski2014transferable} needs target labels to (re-)train with a supervised loss $L(x^t, y^t)$.
Without target labels, our setting denies this supervised training.
Domain adaptation (DA) \citep{quionero2009dataset,saenko2010adapting,ganin2015unsupervised,tzeng2015simultaneous} needs both the source and target data to train with a cross-domain loss $L(x^s, x^t)$.
Test-time training (TTT) \citep{sun2019test} adapts during testing but first alters training to jointly optimize its supervised loss $L(x^s, y^s)$ and self-supervised loss $L(x^s)$.
Without source, our setting denies joint training across domains (DA) or losses (TTT).
Existing settings have their purposes, but do not cover all practical cases when source, target, or supervision are not simultaneously available.

Unexpected target data during testing requires test-time adaptation.
TTT and our setting adapt the model by optimizing an unsupervised loss during testing $L(x^t)$.
During training, TTT jointly optimizes this same loss on source data $L(x^s)$ with a supervised loss $L(x^s, y^s)$, to ensure the parameters $\theta$ are shared across losses for compatibility with adaptation by $L(x^t)$.
Fully test-time adaptation is independent of the training data and training loss given the parameters $\theta$. %
By not changing training, our setting has the potential to require less data and computation for adaptation.

\begin{table}[t]
\caption{
Adaptation settings differ by their data and therefore losses during training and testing.
Of the source $^s$ and target $^t$ data $x$ and labels $y$, our fully test-time setting only needs the target data $x^t$.
}
\label{tab:settings}
\begin{center}
\begin{tabular}{lcccc}
\bf setting & \bf source data & \bf target data & \bf train loss & \bf test loss \\
\toprule
fine-tuning & - & $x^t, y^t$ & $L(x^t, y^t)$ & - \\
domain adaptation & $x^s$, $y^s$ & $x^t$ & $L(x^s, y^s)$ + $L(x^s, x^t)$ & - \\
test-time training & $x^s$, $y^s$ & $x^t$ & $L(x^s, y^s)$ + $L(x^s)$ & $L(x^t)$ \\
fully test-time adaptation & - & $x^t$ & - & $L(x^t)$ \\
\bottomrule
\end{tabular}
\end{center}
\end{table}

\begin{figure}[t]
\centering
\vspace{-4mm}
\begin{minipage}{0.49\textwidth}
\includegraphics[width=\linewidth]{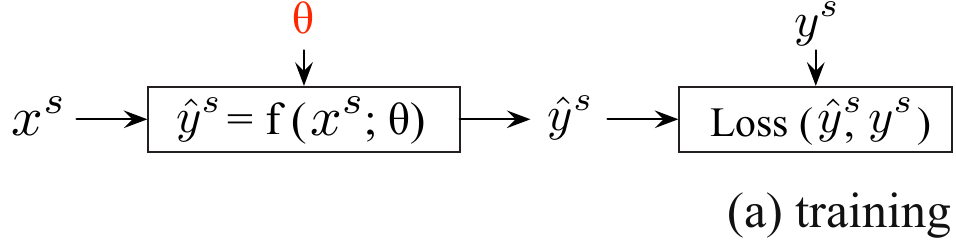}
\end{minipage}%
\hfill%
\vline%
\hfill%
\begin{minipage}{0.49\textwidth}
\centering
\includegraphics[width=\linewidth]{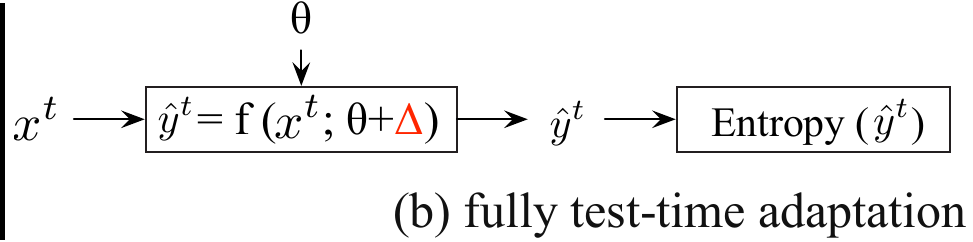}
\end{minipage}
\caption{%
Method overview.
Tent does not alter training (a), but minimizes the entropy of predictions during testing (b)
over a constrained modulation $\textcolor{red}{\Delta}$, given the parameters $\theta$ and target data $x^t$.
}
\label{fig:problem}
\end{figure}

\section{Method: Test Entropy Minimization via Feature Modulation}
\label{sec:minent}

We optimize the model during testing to minimize the entropy of its predictions by modulating its features.
We call our method \emph{tent} for test entropy.
Tent requires a compatible model, an objective to minimize (\Secref{sec:objective}), and parameters to optimize over (\Secref{sec:modulation}) to fully define the algorithm (Section \Secref{sec:algo}).
\Figref{fig:problem} outlines our method for fully test-time adaptation.

The model to be adapted must be trained for the supervised task, probabilistic, and differentiable.
No supervision is provided during testing, so the model must already be trained.
Measuring the entropy of predictions requires a distribution over predictions, so the model must be probabilistic.
Gradients are required for fast iterative optimization, so the model must be differentiable.
Typical deep networks for supervised learning satisfy these model requirements.

\subsection{Entropy Objective}
\label{sec:objective}

Our test-time objective $L(x_t)$ is to minimize the entropy $H(\hat{y})$ of model predictions $\hat{y} = f_\theta(x^t)$.
In particular, we measure the Shannon entropy \citep{shannon1948a-mathematical}, $H(\hat{y}) = -\sum_c p(\hat{y}_c) \log p(\hat{y}_c)$ for the probability $\hat{y}_c$ of class $c$.
Note that optimizing a single prediction has a trivial solution: assign all probability to the most probable class.
We prevent this by jointly optimizing batched predictions over parameters that are shared across the batch.

Entropy is an unsupervised objective because it only depends on predictions and not annotations.
However, as a measure of the predictions it is directly related to the supervised task and model.

In contrast, proxy tasks for self-supervised learning are not directly related to the supervised task.
Proxy tasks derive a self-supervised label $y'$ from the input $x_t$ without the task label $y$.
Examples of these proxies include rotation prediction \citep{gidaris2018unsupervised}, context prediction \citep{doersch2015unsupervised}, and cross-channel auto-encoding \citep{zhang2020split-brain}.
Too much progress on a proxy task could interfere with performance on the supervised task, and self-supervised adaptation methods have to limit or mix updates accordingly \citep{sun2019test,sun2019unsupervised}.
As such, care is needed to choose a proxy compatible with the domain and task, to design the architecture for the proxy model, and to balance optimization between the task and proxy objectives.
Our entropy objective does not need such efforts.

\subsection{Modulation Parameters}
\label{sec:modulation}

The model parameters $\theta$ are a natural choice for test-time optimization, and these are the choice of prior work for train-time entropy minimization \citep{grandvalet2005semi,dhillon2020baseline,carlucci2017autodial}.
However, $\theta$ is the only representation of the training/source data in our setting, and altering $\theta$ could cause the model to diverge from its training.
Furthermore, $f$ can be nonlinear and $\theta$ can be high dimensional, making optimization too sensitive and inefficient for test-time usage.

\begin{figure}[t]
\centering
\includegraphics[width=\linewidth]{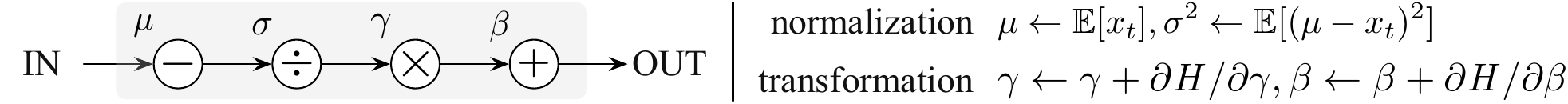}
\caption{
Tent modulates features during testing by estimating normalization statistics $\mu, \sigma$ and optimizing transformation parameters $\gamma, \beta$.
Normalization and transformation apply channel-wise scales and shifts to the features.
The statistics and parameters are updated on target data without use of source data.
In practice, adapting $\gamma, \beta$ is efficient because they make up <$1\%$ of model parameters.
}
\label{fig:arch}
\end{figure}

For stability and efficiency, we instead only update feature modulations that are linear (scales and shifts), and low-dimensional (channel-wise). 
\Figref{fig:arch} shows the two steps of our modulations: normalization by statistics and transformation by parameters.
Normalization centers and standardizes the input $x$ into $\bar{x} = (x - \mu)/\sigma$ by its mean $\mu$ and standard deviation $\sigma$.
Transformation turns $\bar{x}$ into the output $x' = \gamma\bar{x} + \beta$ by affine parameters for scale $\gamma$ and shift $\beta$.
Note that the statistics $\mu, \sigma$ are estimated from the data while the parameters $\gamma, \beta$ are optimized by the loss.

For implementation, we simply repurpose the normalization layers of the source model. 
We update their normalization statistics and affine parameters for all layers and channels during testing.

\subsection{Algorithm}
\label{sec:algo}

\minisection{Initialization}
The optimizer collects the affine transformation parameters $\left\{\gamma_{l,k}, \beta_{l,k}\right\}$ for each normalization layer $l$ and channel $k$ in the source model.
The remaining parameters $\theta \setminus \left\{\gamma_{l,k}, \beta_{l,k}\right\}$ are fixed.
The normalization statistics $\left\{\mu_{l,k}, \sigma_{l,k}\right\}$ from the source data are discarded.

\minisection{Iteration}
Each step updates the normalization statistics and transformation parameters on a batch of data.
The normalization statistics are estimated for each layer in turn, during the forward pass.
The transformation parameters $\gamma, \beta$ are updated by the gradient of the prediction entropy $\nabla H(\hat{y})$, during the backward pass.
Note that the transformation update follows the prediction for the current batch, and so it only affects the next batch (unless forward is repeated).
This needs just one gradient per point of additional computation, so we use this scheme by default for efficiency.

\minisection{Termination}
For online adaptation, no termination is necessary, and iteration continues as long as there is test data.
For offline adaptation, the model is first updated and then inference is repeated.
Adaptation may of course continue by updating for multiple epochs.
\section{Experiments}
\label{sec:experiments}

We evaluate tent for corruption robustness on CIFAR-10/CIFAR-100 and ImageNet, and for domain adaptation on digit adaptation from SVHN to MNIST/MNIST-M/USPS.
Our implementation is in PyTorch \citep{paszke2019pytorch} with the \texttt{pycls} library~\citep{Radosavovic2019}.

\minisection{Datasets}
We run on image classification datasets for corruption and domain adaptation conditions.
For large-scale experiments we choose ImageNet \citep{russakovsky2015imagenet}, with 1,000 classes, a training set of 1.2 million, and a validation set of 50,000.
For experiments at an accessible scale we choose CIFAR-10/CIFAR-100 \citep{krizhevsky2009learning}, with 10/100 classes, a training set of 50,000, and a test set of 10,000.
For domain adaptation we choose SVHN \citep{netzer2011reading} as source and MNIST \citep{lecun1998gradient-based}/MNIST-M \citep{ganin2015unsupervised}/USPS \citep{hull1994database} as targets, with ten classes for the digits 0--9.
SVHN has color images of house numbers from street views with a training set of 73,257 and test set of 26,032.
MNIST/MNIST-M/USPS have handwritten digits with a training sets of 60,000/60,000/7,291 and test sets of 10,000/10,000/2,007.

\minisection{Models}
For corruption we use residual networks \citep{he2016deep} with 26 layers (R-26) on CIFAR-10/100 and 50 layers (R-50) on ImageNet.
For domain adaptation we use the R-26 architecture. %
For fair comparison, all methods in each experimental condition share the same architecture.

Our networks are equipped with batch normalization \citep{ioffe2015batch}.
For the source model without adaptation, the normalization statistics are estimated during training on the source data.
For all test-time adaptation methods, we estimate these statistics during testing on the target data, as done in concurrent work on adaptation by normalization \citep{schneider2020improving,nado2020evaluating}.

\minisection{Optimization}
We optimize the modulation parameters $\gamma, \beta$ following the training hyperparameters for the source model with few changes.
On ImageNet we optimize by SGD with momentum; on other datasets we optimize by Adam \citep{kingma2015adam}. %
We lower the batch size (BS) to reduce memory usage for inference, then lower the learning rate (LR) by the same factor to compensate \citep{goyal2017accurate}.
On ImageNet, we set BS = 64 and LR = 0.00025, and on other datasets we set BS = 128 and LR = 0.001.%
We control for ordering by shuffling and sharing the order across methods.

\minisection{Baselines}
We compare to domain adaptation, self-supervision, normalization, and pseudo-labeling:
\begin{itemize}[noitemsep,leftmargin=4mm]
\item source applies the trained classifier to the test data without adaptation,
\item adversarial domain adaptation (RG) reverses the gradients of a domain classifier on source and target to optimize for a domain-invariant representation \citep{ganin2015unsupervised},
\item self-supervised domain adaptation (UDA-SS) jointly trains self-supervised rotation and position tasks on source and target to optimize for a shared representation \citep{sun2019unsupervised},
\item test-time training (TTT) jointly trains for supervised and self-supervised tasks on source, then keeps training the self-supervised task on target during testing \citep{sun2019test},
\item test-time normalization (BN) updates batch normalization statistics \citep{ioffe2015batch} on the target data during testing \citep{schneider2020improving,nado2020evaluating},
\item pseudo-labeling (PL) tunes a confidence threshold, assigns predictions over the threshold as labels, and then optimizes the model to these pseudo-labels before testing \citep{lee2013pseudo}.
\end{itemize}

Only test-time normalization (BN), pseudo-labeling (PL), and tent (ours) are fully test-time adaptation methods.
See \Secref{sec:ftta} for an explanation and contrast with domain adaptation and test-time training.

\begin{table}[t]
\begin{minipage}{0.49\textwidth}
\caption{%
Corruption benchmark on CIFAR-10-C and CIFAR-100-C for the highest severity. %
Tent has least error, with less optimization than domain adaptation (RG, UDA-SS) and test-time training (TTT), %
and improves on test-time norm (BN).
}
\label{tab:corruption}
\begin{center}
\resizebox{0.98\textwidth}{!}{
\begin{tabular}{lcccc}
\multirow{2}{*}{{\bf Method}} & \multirow{2}{*}{{\bf Source}} & \multirow{2}{*}{{\bf Target}} & \multicolumn{2}{c}{{\bf Error (\%)}} \\
& & & C10-C & C100-C \\
\toprule
Source        & train  &       & 40.8       & 67.2       \\
RG            & train  & train & 18.3       & 38.9       \\

UDA-SS        & train  & train & 16.7       & 47.0       \\
TTT           & train  & test  & 17.5       & 45.0       \\
BN            &        & test  & 17.3       & 42.6       \\
PL            &        & test  & 15.7       & 41.2       \\
Tent (ours)   &        & test  & {\bf 14.3} & {\bf 37.3} \\ %
\bottomrule
\end{tabular}
}
\end{center}
\end{minipage}%
\hfill%
\begin{minipage}{0.49\textwidth}
\centering
\includegraphics[width=\linewidth]{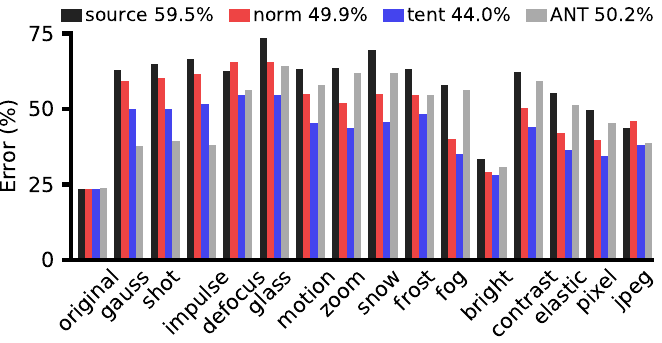}
\captionof{figure}{
Corruption benchmark on ImageNet-C: error for each type averaged over severity levels.
Tent improves on the prior state-of-the-art, adversarial noise training \citep{rusak2020ant}, by fully test-time adaptation \emph{without altering training}.
}
\label{fig:in-c}
\end{minipage}
\end{table}

\subsection{Robustness to Corruptions}
\label{sec:corrupt}

To benchmark robustness to corruption, we make use of common image corruptions (see Appendix \ref{app:corruption} for examples).  %
The CIFAR-10/100 and ImageNet datasets are turned into the CIFAR-10/100-C and ImageNet-C corruption benchmarks by duplicating their test/validation sets and applying 15 types of corruptions at five severity levels \citep{hendrycks2019benchmarking}.

\minisection{Tent improves more with less data and computation.}
Table \ref{tab:corruption} reports errors averaged over corruption types at the severest level of corruption.
On CIFAR-10/100-C we compare all methods, including those that require joint training across domains or losses, given the convenient sizes of these datasets.
Adaptation is offline for fair comparison with offline baselines.
Tent improves on the fully test-time adaptation baselines (BN, PL) but also the domain adaptation (RG, UDA-SS) and test-time training (TTT) methods that need several epochs of optimization on source and target.

\minisection{Tent consistently improves across corruption types.}
Figure \ref{fig:in-c} plots the error for each corruption type averaged over corruption levels on ImageNet-C.
We compare the most efficient methods---source, normalization, and tent---given the large scale of the source data (>1 million images) needed by other methods and the 75 target combinations of corruption types and levels.
Tent and BN adapt online to rival the efficiency of inference without adaptation.
Tent reaches the least error for most corruption types without increasing the error on the original data.

\minisection{Tent reaches a new state-of-the-art without altering training.}
The state-of-the-art methods for robustness extend training with adversarial noise (ANT) \citep{rusak2020ant} for $50.2\%$ error or mixtures of data augmentations (AugMix) \citep{hendrycks2020augmix} for $51.7\%$ error.
Combined with stylization from external images (SIN) \citep{geirhos2019texture}, ANT+SIN reaches $47.4\%$.
Tent reaches a new state-of-the-art of $44.0\%$ by online adaptation and $42.3\%$ by offline adaptation.
It improves on ANT for all types except noise, on which ANT is trained.
This requires just one gradient per test point, without more optimization on the training set (ANT, AugMix) or use of external images (SIN). 
Among fully test-time adaptation methods, tent reduces the error beyond test-time normalization for $18\%$ relative improvement.
In concurrent work, \cite{schneider2020improving} report $49.3\%$ error for test-time normalization, for which tent still gives $14\%$ relative improvement.

\begin{table}[t]
\caption{
Digit domain adaptation from SVHN to MNIST/MNIST-M/USPS.
Source-free adaptation is not only feasible, but more efficient.
Tent always improves on normalization (BN), and in 2/3 cases achieves less error than domain adaptation (RG, UDA-SS) without joint training on source \& target.
}
\label{tab:digit-adapt}
\begin{center}
\vspace{-2mm}
\begin{tabular}{lcccccc}
\multirow{2}{*}{{\bf Method}} & \multirow{2}{*}{{\bf Source}} & \multirow{2}{*}{{\bf Target}} & {\bf Epochs} & \multicolumn{3}{c}{{\bf Error (\%)}} \\
& & & Source + Target & {MNIST} & {MNIST-M} & {USPS} \\
\toprule
Source        & train  &       & -       & 18.2         & 39.7       & 19.3       \\
RG            & train  & train & 10 + 10 & 15.0         & 33.4       & 18.9       \\
UDA-SS        & train  & train & 10 + 10 & 11.1         & {\bf 22.2} & 18.4       \\  %
BN            &        & test  & 0 + 1  & 15.7         & 39.7       & 18.0       \\
Tent (ours)   &        & test  & 0 + 1  & 10.0         & 37.0       & 16.3       \\ %
Tent (ours)   &        & test  & 0 + 10 & {\bf 8.2}    & 36.8       & {\bf 14.4} \\ %
\bottomrule
\end{tabular}
\end{center}
\vspace{-2mm}
\end{table}

\subsection{Source-Free Domain Adaptation}
\label{sec:da}

We benchmark digit adaptation \citep{ganin2015unsupervised,tzeng2015simultaneous,tzeng2017adversarial,shu2018dirt} for
shifts from SVHN to MNIST/MNIST-M/USPS.
Recall that unsupervised domain adaptation makes use the labeled source data and unlabeled target data, while our fully test-time adaptation setting denies use of source data.
Adaptation is offline for fair comparison with offline baselines.

\minisection{Tent adapts to target without source.}
Table \ref{tab:digit-adapt} reports the target errors for domain adaptation and fully test-time adaptation methods.
Test-time normalization (BN) marginally improves, while adversarial domain adaptation (RG) and self-supervised domain adaptation (UDA-SS) improve more by joint training on source and target. %
Tent always has lower error than the source model and BN, and it achieves the lowest error in 2/3 cases, even in just one epoch and without use of source data. %

While encouraging for fully test-time adaptation, unsupervised domain adaptation remains necessary for the highest accuracy and harder shifts.
For SVHN-to-MNIST, DIRT-T \citep{shu2018dirt} achieves a remarkable $0.6\%$ error
\footnote{%
We exclude DIRT-T from our experiments because of incomparable differences in architecture and model selection.
DIRT-T tunes with labeled target data, but we do not.
Please refer to \citet{shu2018dirt} for more detail.
}.
For MNIST-to-SVHN, a difficult shift with source-only error of $71.3\%$, DIRT-T reaches $45.5\%$ and UDA-SS reaches $38.7\%$.
Tent fails on this shift and increases error to $79.8\%$.
In this case success presently requires joint optimization over source and target.

\minisection{Tent needs less computation, but still improves with more.}
Tent adapts efficiently on target data alone with just one gradient per point.
RG \& UDA-SS also use the source data (SVHN train), which is ${\sim}7\times$ the size of the target data (MNIST test), and optimize for $10$ epochs.
Tent adapts with ${\sim}80\times$ less computation.  %
With more updates, tent reaches $8.2\%$ error in 10 epochs and $6.5\%$ in 100 epochs. %
With online updates, tent reaches $12.5\%$ error in one epoch and $8.4\%$ error in 10 epochs.

\minisection{Tent scales to semantic segmentation.}
To show scalability to large models and inputs, we evaluate semantic segmentation (pixel-wise classification) on a domain shift from a simulated source to a real target.
The source is GTA \citep{richter2017playing}, a video game in an urban environment, and the target is Cityscapes \citep{cordts2016cityscapes}, an urban autonomous driving dataset.
The model is HRNet-W18, a fully convolutional network \citep{shelhamer2017fully} with high-resolution architecture \citep{wang2020deep}.
The target intersection-over-union scores (higher is better) are source $28.8\%$, BN $31.4\%$, and tent $35.8\%$ with offline optimization by Adam.
For \emph{adaptation to a single image}, tent reaches $36.4\%$ in 10 iterations with episodic optimization.
See the appendix for a qualitative example (Appendix \ref{app:da}).

\minisection{Tent scales to the VisDA-C challenge.}
To show adaptation on a more difficult benchmark, we evaluate on the VisDA-C challenge \citep{peng2017visda}.
The task is object recognition for 12 classes where the source data is synthesized by rendering 3D models and the target data is collected from real scenes.
The validation error for our source model (ResNet-50, pretrained on ImageNet) is $56.1\%$, while tent reaches $45.6\%$, and improves to $39.6\%$ by updating all layers except for the final classifier as done by \cite{liang2020source}.
Although offline source-free adaptation by model adaptation \citep{li2020model} or SHOT \citep{liang2020source} can reach lower error with more computation and tuning, tent can adapt online during testing. %

\begin{figure}[t]
\centering
\begin{minipage}{0.49\textwidth}
\includegraphics[width=\linewidth]{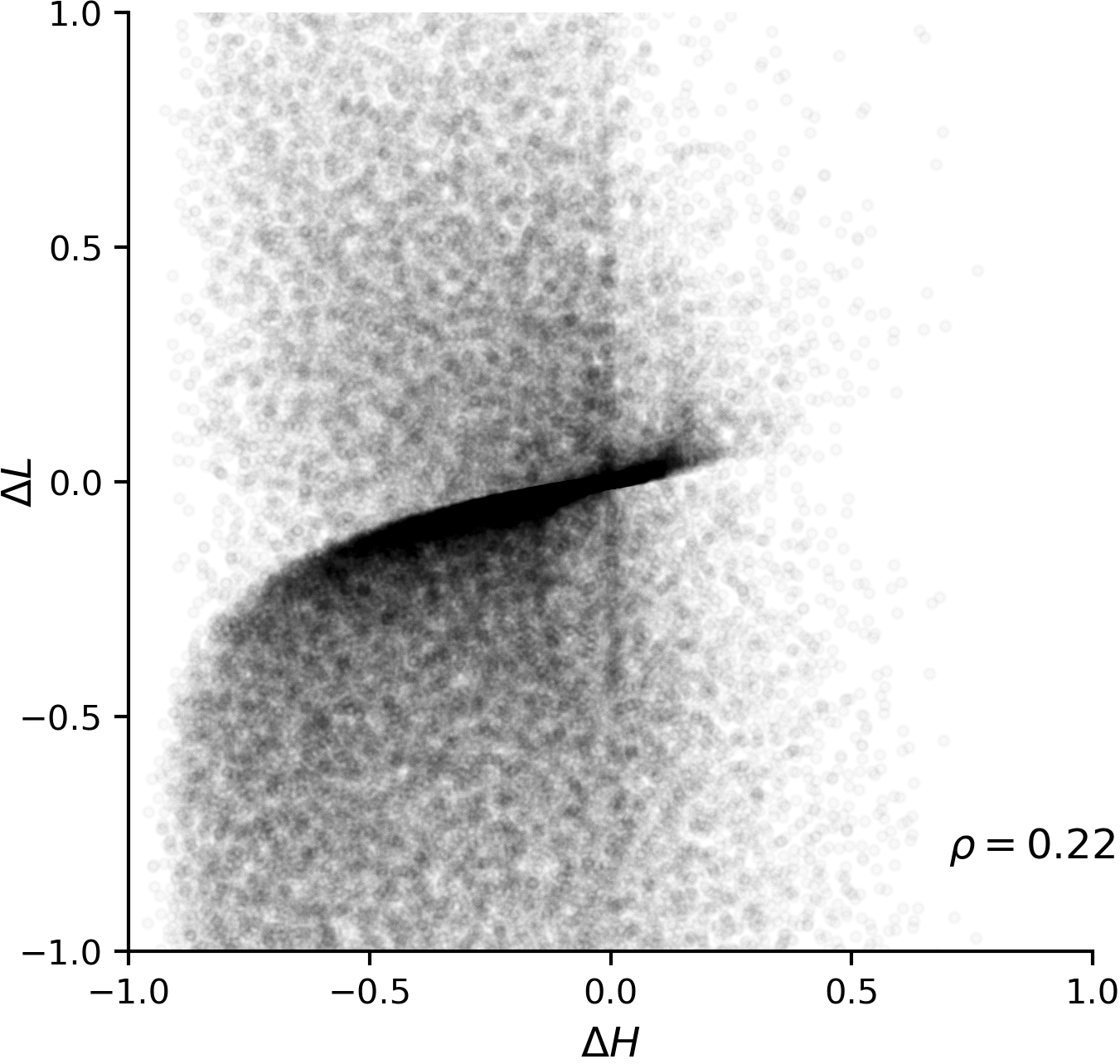}
\caption{
Tent reduces the entropy and loss.
We plot changes in entropy $\Delta H$ and loss $\Delta L$ for all of CIFAR-100-C.
Change in entropy rank-correlates with change in loss: note the dark diagonal and the rank correlation coefficient of $0.22$.
}
\label{fig:deltas}
\end{minipage}%
\hfill%
\begin{minipage}{0.49\textwidth}
\centering
\includegraphics[width=0.98\linewidth]{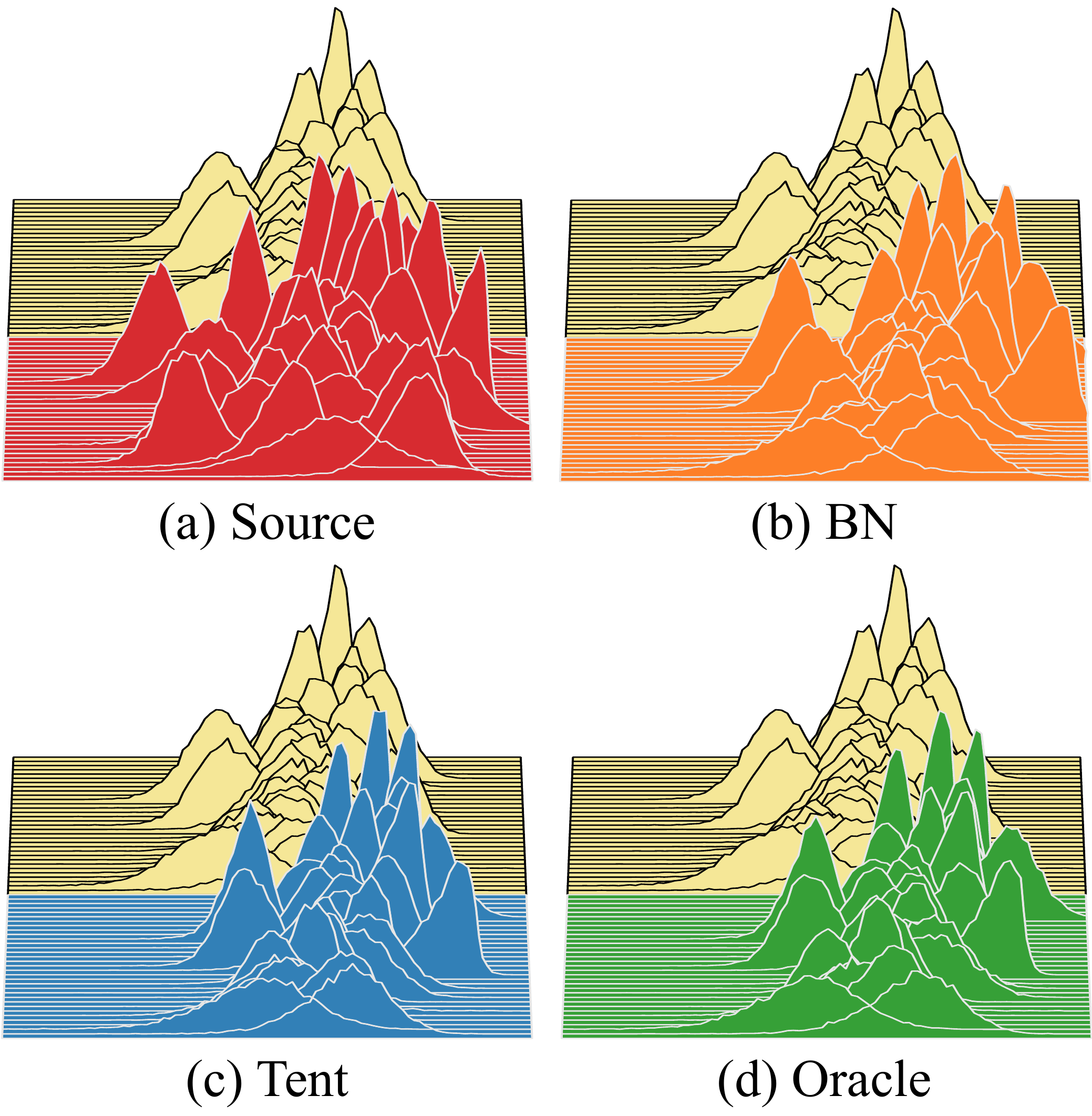}
\vspace{0.1mm}
\caption{
Adapted features on CIFAR-100-C with Gaussian noise (front) and reference features without corruption (back).
Corruption shifts features away from the reference, but BN reduces the shifts.
Tent instead shifts features more, and closer to an oracle that optimizes on target labels.
}
\label{fig:features}
\end{minipage}
\end{figure}

\subsection{Analysis}
\label{sec:analysis}

\minisection{Tent reduces entropy and error.}
\Figref{fig:deltas} verifies tent does indeed reduce the entropy and the task loss (softmax cross-entropy).
We plot changes in entropy and loss on CIFAR-100-C for all 75 corruption type/level combinations.
Both axes are normalized by the maximum entropy of a prediction ($\log100$) and clipped to $\pm1$.
Most points have lower entropy and error after adaptation.

\minisection{Tent needs feature modulation.}
We ablate the normalization and transformation steps of feature modulation.
Not updating normalization increases errors, and can fail to improve over BN and PL.
Not updating transformation parameters reduces the method to test-time normalization.
Updating only the last layer of the model can improve but then degrades with further optimization.
Updating the full model parameters $\theta$ never improves over the unadapted source model.

\minisection{Tent generalizes across target data.}
Adaptation could be limited to the points used for updates.
We check that adaptation generalizes across points by adapting on target train and not target test.
Test errors drop: CIFAR-100-C error goes from 37.3\% to 34.2\% and SVHN-to-MNIST error goes from 8.2\% to 6.5\%.
(Train is larger than test; when subsampling to the same size errors differ by <$0.1\%$.)
Therefore the adapted modulation is not point specific but general.

\minisection{Tent modulation differs from normalization.}
Modulation normalizes and transforms features. 
We examine the combined effect.
\Figref{fig:features} contrasts adapted features on corrupted data against reference features on uncorrupted data.
We plot features from the source model, normalization, tent, and an oracle that optimizes on the target labels.
Normalization makes features more like the reference, but tent does not.
Instead, tent makes features more like the oracle.
This suggests a different and task-specific effect.
See the appendix for visualizations of more layers (Appendix \ref{sec:app-features}).

\minisection{Tent adapts alternative architectures.}
Tent is architecture agnostic in principle. 
To gauge its generality in practice, we evaluate new architectures based on self-attention (SAN) \citep{zhao2020exploring} and equilibrium solving (MDEQ) \citep{bai2020multiscale} for corruption robustness on CIFAR-100-C.
Table \ref{tab:alt-arch} shows that tent reduces error with the same settings as convolutional residual networks.

\begin{table}[h]
\caption{
Tent adapts alternative architectures on CIFAR-100-C without tuning.
Results are error (\%).
}
\label{tab:alt-arch}
\vspace{-2mm}
\begin{center}
\begin{tabular}{ccccccccc}
\multicolumn{3}{c}{{\bf SAN-10 (pair)}} & \multicolumn{3}{c}{{\bf SAN-10 (patch)}} & \multicolumn{3}{c}{{\bf MDEQ (large)}} \\
Source & BN & Tent & Source & BN & Tent & Source & BN & Tent \\
\toprule
55.3 & 39.7 & {\bf 36.7} & 48.0 & 31.8 & {\bf 29.2} & 53.3 & 44.9 & {\bf 41.7} \\
\bottomrule
\end{tabular}
\end{center}
\vspace{-2mm}
\end{table}

\section{Related Work}

We relate tent to existing adaptation, entropy minimization, and feature modulation methods.

\minisection{Train-Time Adaptation}
Domain adaptation jointly optimizes on source and target by cross-domain losses $L(x^s, x^t)$ to mitigate shift.
These losses optimize feature alignment \citep{gretton2009covariate,sun2017correlation}, adversarial invariance \citep{ganin2015unsupervised,tzeng2017adversarial}, or shared proxy tasks \citep{sun2019unsupervised}.
Transduction \citep{gammerman1998learning,joachims1999transductive,zhou2004learning} jointly optimizes on train and test to better fit specific test instances.
While effective in their settings, neither applies when joint use of source/train and target/test is denied.
Tent adapts on target alone.

Recent ``source-free'' methods \citep{li2020model,kundu2020universal,liang2020source} also adapt without source data.
\citet{li2020model,kundu2020universal} rely on generative modeling and optimize multiple models with multiple losses.
\citet{kundu2020universal,liang2020source} also alter training.
Tent does not need generative modeling, nor does it alter training, and so it can deployed more generally to adapt online with much more computational efficiency.
SHOT \citep{liang2020source} adapts by information maximization (entropy minimization and diversity regularization), but differs in its other losses and its parameterization.
These source-free methods optimize offline with multiple losses for multiple epochs, which requires more tuning and computation than tent, but may achieve more accuracy with more computation.
Tent optimizes online with just one loss and an efficient parameterization of modulation to emphasize fully test-time adaptation during inference.
We encourage examination of each of these works on the frontier of adaptation without source data. 

\cite{chidlovskii2016domain} are the first to motivate adaptation without source data for legal, commercial, or technical concerns.
They adapt predictions by applying denoising auto-encoders while we adapt models by entropy minimization.
We share their motivations, but the methods and experiments differ.

\minisection{Test-Time Adaptation}
Tent adapts by test-time optimization and normalization to update the model.
Test-time adaptation of predictions, through which harder and uncertain cases are adjusted based on easier and certain cases \citep{jain2011online}, provides inspiration for certainty-based model adaptation schemes like our own.

Test-time training (TTT) \citep{sun2019test} also optimizes during testing, but differs in its loss and must alter training.
TTT relies on a proxy task, such as recognizing rotations of an image, and so its loss depends on the choice of proxy.
(Indeed, its authors caution that the proxy must be ``both well-defined and non-trivial in the new domain'').
TTT alters training to optimize this proxy loss on source before adapting to target.
Tent adapts without proxy tasks and without altering training.

Normalizing feature statistics is common for domain adaptation \citep{gretton2009covariate,sun2017correlation}. %
For batch normalization \cite{li2017revisiting,carlucci2017autodial} separate source and target statistics during training.
\cite{schneider2020improving,nado2020evaluating} estimate target statistics during testing to improve generalization.
Tent builds on test-time normalization to further reduce generalization error.

\minisection{Entropy Minimization}
Entropy minimization is a key regularizer for domain adaptation \citep{carlucci2017autodial,shu2018dirt,saito2019semi,roy2019unsupervised}, semi-supervised learning \citep{grandvalet2005semi,lee2013pseudo,berthelot2019mixmatch}, and few-shot learning \citep{dhillon2020baseline}.
Regularizing entropy penalizes decisions at high densities in the data distribution to improve accuracy for distinct classes \citep{grandvalet2005semi}.
These methods regularize entropy during training in concert with other supervised and unsupervised losses on additional data.
Tent is the first to minimize entropy during testing, for adaptation to dataset shifts, without other losses or data.
Entropic losses are common; our contribution is to exhibit entropy \emph{as the sole loss} for fully test-time adaptation.

\minisection{Feature Modulation}
Modulation makes a model vary with its input.
We optimize modulations that are simpler than the full model for stable and efficient adaptation.
We modulate channel-wise affine transformations, for their effectiveness in tandem with normalization \citep{ioffe2015batch,wu2018group}, and for their flexibility in conditioning for different tasks \citep{perez2018film}.
These normalization and conditioning methods optimize the modulation during training by a supervised loss, but keep it fixed during testing.
We optimize the modulation during testing by an unsupervised loss, so that it can adapt to different target data.
\section{Discussion}

Tent reduces generalization error on shifted data by test-time entropy minimization.
In minimizing entropy, the model adapts itself to feedback from its own predictions. 
This is truly self-supervised self-improvement.
Self-supervision of this sort is totally defined by the supervised task, unlike proxy tasks designed to extract more supervision from the data, and yet it remarkably still reduces error.
Nevertheless, errors due to corruption and other shifts remain, and therefore more adaptation is needed.
Next steps should pursue test-time adaptation on more and harder types of shift, over more general parameters, and by more effective and efficient losses.

\minisection{Shifts}
Tent reduces error for a variety of shifts including image corruptions, %
simple changes in appearance for digits, %
and simulation-to-real discrepancies. %
These shifts are popular as standardized benchmarks, but other real-world shifts exist.
For instance, the CIFAR 10.1 and ImageNetV2 test sets \citep{recht2018cifar,recht2019do-imagenet}, made by reproducing the dataset collection procedures, entail natural but unknown shifts.
Although error is higher on both sets, indicating the presence of shift, tent does not improve generalization.
Adversarial shifts \citep{szegedy2014intriguing} also threaten real-world usage, and attackers keep adapting to defenses.
While adversarial training \citep{madry2018towards} makes a difference, test-time adaptation could help counter such test-time attacks.

\minisection{Parameters}
Tent modulates the model by normalization and transformation, but much of the model stays fixed.
Test-time adaptation could update more of the model, but the issue is to identify parameters that are both expressive and reliable, and this may interact with the choice of loss.
TTT adapts multiple layers of features shared by supervised and self-supervised models and SHOT adapts all but the last layer(s) of the model.
These choices depend on the model architecture, the loss, and tuning.
For tent modulation is reliable, but the larger shift on VisDA is better addressed by the SHOT parameterization.
Jointly adapting the input could be a more general alternative.
If a model can adapt itself on target, then perhaps its input gradients might optimize spatial transformations or image translations to reduce shift without source data. 

\minisection{Losses}
Tent minimizes entropy.
For more adaptation, is there an effective loss for general but episodic test-time optimization?
Entropy is general across tasks but limited in scope.
It needs batches for optimization, and cannot update episodically on one point at a time.
TTT can do so, but only with the right proxy task.
For less computation, is there an efficient loss for more local optimization?
Tent and TTT both require full (re-)computation of the model for updates because they depend on its predictions.
If the loss were instead defined on the representation, then updates would require less forward and backward computation. 
Returning to entropy specifically, this loss may interact with calibration \citep{guo2017calibration}, as better uncertainty estimation could drive better adaptation.

We hope that the fully test-time adaptation setting can promote new methods for equipping a model to adapt itself, just as tent yields a new model with every update.

\subsection*{Acknowledgments}
We thank
Eric Tzeng for discussions on domain adaptation,
Bill Freeman for comments on the experiments,  
Yu Sun for consultations on test-time training, and
Kelsey Allen for feedback on the exposition.
We thank the anonymous reviewers of ICLR 2021 for their feedback, 
which certainly improved the latest adaptation of the paper.

\newpage
\small

{
\bibliographystyle{iclr2021_conference}
\bibliography{minent}

\begin{thebibliography}{66}
\providecommand{\natexlab}[1]{#1}
\providecommand{\url}[1]{\texttt{#1}}
\expandafter\ifx\csname urlstyle\endcsname\relax
  \providecommand{\doi}[1]{doi: #1}\else
  \providecommand{\doi}{doi: \begingroup \urlstyle{rm}\Url}\fi

\bibitem[Bai et~al.(2020)Bai, Koltun, and Kolter]{bai2020multiscale}
Shaojie Bai, Vladlen Koltun, and J~Zico Kolter.
\newblock Multiscale deep equilibrium models.
\newblock \emph{arXiv preprint arXiv:2006.08656}, 2020.

\bibitem[Berthelot et~al.(2019)Berthelot, Carlini, Goodfellow, Papernot,
  Oliver, and Raffel]{berthelot2019mixmatch}
David Berthelot, Nicholas Carlini, Ian Goodfellow, Nicolas Papernot, Avital
  Oliver, and Colin~A Raffel.
\newblock Mixmatch: A holistic approach to semi-supervised learning.
\newblock In \emph{NeurIPS}, 2019.

\bibitem[Carlucci et~al.(2017)Carlucci, Porzi, Caputo, Ricci, and
  Bulo]{carlucci2017autodial}
Fabio~Maria Carlucci, Lorenzo Porzi, Barbara Caputo, Elisa Ricci, and
  Samuel~Rota Bulo.
\newblock Autodial: Automatic domain alignment layers.
\newblock In \emph{2017 IEEE International Conference on Computer Vision
  (ICCV)}, pp.\  5077--5085. IEEE, 2017.

\bibitem[Chidlovskii et~al.(2016)Chidlovskii, Clinchant, and
  Csurka]{chidlovskii2016domain}
Boris Chidlovskii, Stephane Clinchant, and Gabriela Csurka.
\newblock Domain adaptation in the absence of source domain data.
\newblock In \emph{SIGKDD}, pp.\  451--460, 2016.

\bibitem[Cordts et~al.(2016)Cordts, Omran, Ramos, Rehfeld, Enzweiler, Benenson,
  Franke, Roth, and Schiele]{cordts2016cityscapes}
Marius Cordts, Mohamed Omran, Sebastian Ramos, Timo Rehfeld, Markus Enzweiler,
  Rodrigo Benenson, Uwe Franke, Stefan Roth, and Bernt Schiele.
\newblock The cityscapes dataset for semantic urban scene understanding.
\newblock In \emph{CVPR}, 2016.

\bibitem[Dhillon et~al.(2020)Dhillon, Chaudhari, Ravichandran, and
  Soatto]{dhillon2020baseline}
Guneet~Singh Dhillon, Pratik Chaudhari, Avinash Ravichandran, and Stefano
  Soatto.
\newblock A baseline for few-shot image classification.
\newblock In \emph{ICLR}, 2020.

\bibitem[Doersch et~al.(2015)Doersch, Gupta, and
  Efros]{doersch2015unsupervised}
Carl Doersch, Abhinav Gupta, and Alexei~A Efros.
\newblock Unsupervised visual representation learning by context prediction.
\newblock In \emph{ICCV}, 2015.

\bibitem[{Donahue} et~al.(2014){Donahue}, {Jia}, {Vinyals}, {Hoffman}, {Zhang},
  {Tzeng}, and {Darrell}]{donahue2014decaf}
J.~{Donahue}, Y.~{Jia}, O.~{Vinyals}, J.~{Hoffman}, N.~{Zhang}, E.~{Tzeng}, and
  T.~{Darrell}.
\newblock Decaf: A deep convolutional activation feature for generic visual
  recognition.
\newblock In \emph{ICML}, 2014.

\bibitem[Gammerman et~al.(1998)Gammerman, Vovk, and
  Vapnik]{gammerman1998learning}
A~Gammerman, V~Vovk, and V~Vapnik.
\newblock Learning by transduction.
\newblock In \emph{UAI}, 1998.

\bibitem[Ganin \& Lempitsky(2015)Ganin and Lempitsky]{ganin2015unsupervised}
Yaroslav Ganin and Victor Lempitsky.
\newblock Unsupervised domain adaptation by backpropagation.
\newblock In \emph{ICML}, 2015.

\bibitem[Geirhos et~al.(2018)Geirhos, Temme, Rauber, Sch{\"u}tt, Bethge, and
  Wichmann]{geirhos2018generalisation}
Robert Geirhos, Carlos~RM Temme, Jonas Rauber, Heiko~H Sch{\"u}tt, Matthias
  Bethge, and Felix~A Wichmann.
\newblock Generalisation in humans and deep neural networks.
\newblock In \emph{NeurIPS}, 2018.

\bibitem[Geirhos et~al.(2019)Geirhos, Rubisch, Michaelis, Bethge, Wichmann, and
  Brendel]{geirhos2019texture}
Robert Geirhos, Patricia Rubisch, Claudio Michaelis, Matthias Bethge, Felix~A.
  Wichmann, and Wieland Brendel.
\newblock Imagenet-trained {CNN}s are biased towards texture; increasing shape
  bias improves accuracy and robustness.
\newblock In \emph{International Conference on Learning Representations}, 2019.

\bibitem[Gidaris et~al.(2018)Gidaris, Singh, and
  Komodakis]{gidaris2018unsupervised}
Spyros Gidaris, Praveer Singh, and Nikos Komodakis.
\newblock Unsupervised representation learning by predicting image rotations.
\newblock In \emph{ICLR}, 2018.

\bibitem[Goyal et~al.(2017)Goyal, Doll{\'a}r, Girshick, Noordhuis, Wesolowski,
  Kyrola, Tulloch, Jia, and He]{goyal2017accurate}
Priya Goyal, Piotr Doll{\'a}r, Ross Girshick, Pieter Noordhuis, Lukasz
  Wesolowski, Aapo Kyrola, Andrew Tulloch, Yangqing Jia, and Kaiming He.
\newblock Accurate, large minibatch sgd: training imagenet in 1 hour.
\newblock \emph{arXiv preprint arXiv:1706.02677}, 2017.

\bibitem[Grandvalet \& Bengio(2005)Grandvalet and Bengio]{grandvalet2005semi}
Yves Grandvalet and Yoshua Bengio.
\newblock Semi-supervised learning by entropy minimization.
\newblock In \emph{NeurIPS}, 2005.

\bibitem[Gretton et~al.(2009)Gretton, Smola, Huang, Schmittfull, Borgwardt, and
  Sch{\"o}lkopf]{gretton2009covariate}
A.~Gretton, AJ. Smola, J.~Huang, M.~Schmittfull, KM. Borgwardt, and
  B.~Sch{\"o}lkopf.
\newblock Covariate shift and local learning by distribution matching.
\newblock In \emph{Dataset Shift in Machine Learning}, pp.\  131--160. MIT
  Press, Cambridge, MA, USA, 2009.

\bibitem[Guo et~al.(2017)Guo, Pleiss, Sun, and Weinberger]{guo2017calibration}
Chuan Guo, Geoff Pleiss, Yu~Sun, and Kilian~Q Weinberger.
\newblock On calibration of modern neural networks.
\newblock In \emph{ICML}, 2017.

\bibitem[He et~al.(2016)He, Zhang, Ren, and Sun]{he2016deep}
Kaiming He, Xiangyu Zhang, Shaoqing Ren, and Jian Sun.
\newblock Deep residual learning for image recognition.
\newblock In \emph{CVPR}, June 2016.

\bibitem[Hendrycks \& Dietterich(2019)Hendrycks and
  Dietterich]{hendrycks2019benchmarking}
Dan Hendrycks and Thomas Dietterich.
\newblock Benchmarking neural network robustness to common corruptions and
  perturbations.
\newblock In \emph{ICLR}, 2019.

\bibitem[Hendrycks et~al.(2020)Hendrycks, Mu, Cubuk, Zoph, Gilmer, and
  Lakshminarayanan]{hendrycks2020augmix}
Dan Hendrycks, Norman Mu, Ekin~D Cubuk, Barret Zoph, Justin Gilmer, and Balaji
  Lakshminarayanan.
\newblock Augmix: A simple data processing method to improve robustness and
  uncertainty.
\newblock In \emph{ICLR}, 2020.

\bibitem[Hull(1994)]{hull1994database}
Jonathan~J. Hull.
\newblock A database for handwritten text recognition research.
\newblock \emph{TPAMI}, 1994.

\bibitem[Ioffe \& Szegedy(2015)Ioffe and Szegedy]{ioffe2015batch}
Sergey Ioffe and Christian Szegedy.
\newblock Batch normalization: Accelerating deep network training by reducing
  internal covariate shift.
\newblock In \emph{ICML}, 2015.

\bibitem[Jain \& Learned-Miller(2011)Jain and Learned-Miller]{jain2011online}
Vidit Jain and Erik Learned-Miller.
\newblock Online domain adaptation of a pre-trained cascade of classifiers.
\newblock In \emph{CVPR}, 2011.

\bibitem[Joachims(1999)]{joachims1999transductive}
Thorsten Joachims.
\newblock Transductive inference for text classification using support vector
  machines.
\newblock In \emph{ICML}, 1999.

\bibitem[Kingma \& Ba(2015)Kingma and Ba]{kingma2015adam}
Diederik Kingma and Jimmy Ba.
\newblock Adam: A method for stochastic optimization.
\newblock In \emph{ICLR}, 2015.

\bibitem[Krizhevsky et~al.(2012)Krizhevsky, Sutskever, and
  Hinton]{krizhevsky2012imagenet}
A.~Krizhevsky, I.~Sutskever, and G.~Hinton.
\newblock Imagenet classification with deep convolutional neural networks.
\newblock \emph{NeurIPS}, 25, 2012.

\bibitem[Krizhevsky(2009)]{krizhevsky2009learning}
Alex Krizhevsky.
\newblock Learning multiple layers of features from tiny images.
\newblock Technical report, University of Toronto, 2009.

\bibitem[Kundu et~al.(2020)Kundu, Venkat, Babu, et~al.]{kundu2020universal}
Jogendra~Nath Kundu, Naveen Venkat, R~Venkatesh Babu, et~al.
\newblock Universal source-free domain adaptation.
\newblock In \emph{CVPR}, pp.\  4544--4553, 2020.

\bibitem[LeCun et~al.(1998)LeCun, Bottou, Bengio, and
  Haffner]{lecun1998gradient-based}
Y.~LeCun, L.~Bottou, Y.~Bengio, and P.~Haffner.
\newblock Gradient-based learning applied to document recognition.
\newblock \emph{Proceedings of the IEEE}, 86\penalty0 (11):\penalty0
  2278--2324, 1998.

\bibitem[Lee(2013)]{lee2013pseudo}
Dong-Hyun Lee.
\newblock Pseudo-label: The simple and efficient semi-supervised learning
  method for deep neural networks.
\newblock In \emph{ICML Workshop on challenges in representation learning},
  2013.

\bibitem[Li et~al.(2020)Li, Jiao, Cao, Wong, and Wu]{li2020model}
Rui Li, Qianfen Jiao, Wenming Cao, Hau-San Wong, and Si~Wu.
\newblock Model adaptation: Unsupervised domain adaptation without source data.
\newblock In \emph{CVPR}, June 2020.

\bibitem[Li et~al.(2017)Li, Wang, Shi, Liu, and Hou]{li2017revisiting}
Yanghao Li, Naiyan Wang, Jianping Shi, Jiaying Liu, and Xiaodi Hou.
\newblock Revisiting batch normalization for practical domain adaptation.
\newblock In \emph{ICLRW}, 2017.

\bibitem[Liang et~al.(2020)Liang, Hu, and Feng]{liang2020source}
Jian Liang, Dapeng Hu, and Jiashi Feng.
\newblock Do we really need to access the source data? source hypothesis
  transfer for unsupervised domain adaptation.
\newblock In \emph{ICML}, 2020.

\bibitem[Madry et~al.(2018)Madry, Makelov, Schmidt, Tsipras, and
  Vladu]{madry2018towards}
Aleksander Madry, Aleksandar Makelov, Ludwig Schmidt, Dimitris Tsipras, and
  Adrian Vladu.
\newblock Towards deep learning models resistant to adversarial attacks.
\newblock In \emph{International Conference on Learning Representations}, 2018.

\bibitem[Nado et~al.(2020)Nado, Padhy, Sculley, D'Amour, Lakshminarayanan, and
  Snoek]{nado2020evaluating}
Zachary Nado, Shreyas Padhy, D~Sculley, Alexander D'Amour, Balaji
  Lakshminarayanan, and Jasper Snoek.
\newblock Evaluating prediction-time batch normalization for robustness under
  covariate shift.
\newblock \emph{arXiv preprint arXiv:2006.10963}, 2020.

\bibitem[Netzer et~al.(2011)Netzer, Wang, Coates, Bissacco, Wu, and
  Ng]{netzer2011reading}
Yuval Netzer, Tao Wang, Adam Coates, Alessandro Bissacco, Bo~Wu, and Andrew~Y
  Ng.
\newblock Reading digits in natural images with unsupervised feature learning.
\newblock \emph{NeurIPS Workshop on Deep Learning and Unsupervised Feature
  Learning}, 2011.

\bibitem[Paszke et~al.(2019)Paszke, Gross, Massa, Lerer, Bradbury, Chanan,
  Killeen, Lin, Gimelshein, Antiga, et~al.]{paszke2019pytorch}
Adam Paszke, Sam Gross, Francisco Massa, Adam Lerer, James Bradbury, Gregory
  Chanan, Trevor Killeen, Zeming Lin, Natalia Gimelshein, Luca Antiga, et~al.
\newblock Pytorch: An imperative style, high-performance deep learning library.
\newblock In \emph{NeurIPS}, 2019.

\bibitem[Peng et~al.(2017)Peng, Usman, Kaushik, Hoffman, Wang, and
  Saenko]{peng2017visda}
Xingchao Peng, Ben Usman, Neela Kaushik, Judy Hoffman, Dequan Wang, and Kate
  Saenko.
\newblock {VisDA}: The visual domain adaptation challenge.
\newblock \emph{arXiv preprint arXiv:1710.06924}, 2017.

\bibitem[Perez et~al.(2018)Perez, Strub, De~Vries, Dumoulin, and
  Courville]{perez2018film}
Ethan Perez, Florian Strub, Harm De~Vries, Vincent Dumoulin, and Aaron
  Courville.
\newblock Film: Visual reasoning with a general conditioning layer.
\newblock In \emph{AAAI}, 2018.

\bibitem[Quionero-Candela et~al.(2009)Quionero-Candela, Sugiyama, Schwaighofer,
  and Lawrence]{quionero2009dataset}
Joaquin Quionero-Candela, Masashi Sugiyama, Anton Schwaighofer, and Neil~D
  Lawrence.
\newblock \emph{Dataset shift in machine learning}.
\newblock MIT Press, Cambridge, MA, USA, 2009.

\bibitem[Radosavovic et~al.(2019)Radosavovic, Johnson, Xie, Lo, and
  Doll{\'a}r]{Radosavovic2019}
Ilija Radosavovic, Justin Johnson, Saining Xie, Wan-Yen Lo, and Piotr
  Doll{\'a}r.
\newblock On network design spaces for visual recognition.
\newblock In \emph{ICCV}, 2019.

\bibitem[Recht et~al.(2018)Recht, Roelofs, Schmidt, and
  Shankar]{recht2018cifar}
Benjamin Recht, Rebecca Roelofs, Ludwig Schmidt, and Vaishaal Shankar.
\newblock Do cifar-10 classifiers generalize to cifar-10?
\newblock \emph{arXiv preprint arXiv:1806.00451}, 2018.

\bibitem[Recht et~al.(2019)Recht, Roelofs, Schmidt, and
  Shankar]{recht2019do-imagenet}
Benjamin Recht, Rebecca Roelofs, Ludwig Schmidt, and Vaishaal Shankar.
\newblock Do {I}mage{N}et classifiers generalize to {I}mage{N}et?
\newblock In \emph{ICML}, 2019.

\bibitem[Richter et~al.(2017)Richter, Hayder, and Koltun]{richter2017playing}
Stephan~R Richter, Zeeshan Hayder, and Vladlen Koltun.
\newblock Playing for benchmarks.
\newblock In \emph{ICCV}, 2017.

\bibitem[Roy et~al.(2019)Roy, Siarohin, Sangineto, Bulo, Sebe, and
  Ricci]{roy2019unsupervised}
Subhankar Roy, Aliaksandr Siarohin, Enver Sangineto, Samuel~Rota Bulo, Nicu
  Sebe, and Elisa Ricci.
\newblock Unsupervised domain adaptation using feature-whitening and consensus
  loss.
\newblock In \emph{CVPR}, 2019.

\bibitem[Rusak et~al.(2020)Rusak, Schott, Zimmermann, Bitterwolf, Bringmann,
  Bethge, and Brendel]{rusak2020ant}
Evgenia Rusak, Lukas Schott, Roland~S Zimmermann, Julian Bitterwolf, Oliver
  Bringmann, Matthias Bethge, and Wieland Brendel.
\newblock A simple way to make neural networks robust against diverse image
  corruptions.
\newblock In \emph{ECCV}, 2020.

\bibitem[Russakovsky et~al.(2015)Russakovsky, Deng, Su, Krause, Satheesh, Ma,
  Huang, Karpathy, Khosla, Bernstein, et~al.]{russakovsky2015imagenet}
Olga Russakovsky, Jia Deng, Hao Su, Jonathan Krause, Sanjeev Satheesh, Sean Ma,
  Zhiheng Huang, Andrej Karpathy, Aditya Khosla, Michael Bernstein, et~al.
\newblock {ImageNet} large scale visual recognition challenge.
\newblock \emph{IJCV}, 2015.

\bibitem[Saenko et~al.(2010)Saenko, Kulis, Fritz, and
  Darrell]{saenko2010adapting}
Kate Saenko, Brian Kulis, Mario Fritz, and Trevor Darrell.
\newblock Adapting visual category models to new domains.
\newblock In \emph{European conference on computer vision}, pp.\  213--226.
  Springer, 2010.

\bibitem[Saito et~al.(2019)Saito, Kim, Sclaroff, Darrell, and
  Saenko]{saito2019semi}
Kuniaki Saito, Donghyun Kim, Stan Sclaroff, Trevor Darrell, and Kate Saenko.
\newblock Semi-supervised domain adaptation via minimax entropy.
\newblock In \emph{ICCV}, 2019.

\bibitem[Schneider et~al.(2020)Schneider, Rusak, Eck, Bringmann, Brendel, and
  Bethge]{schneider2020improving}
Steffen Schneider, Evgenia Rusak, Luisa Eck, Oliver Bringmann, Wieland Brendel,
  and Matthias Bethge.
\newblock Improving robustness against common corruptions by covariate shift
  adaptation.
\newblock \emph{arXiv preprint arXiv:2006.16971}, 2020.

\bibitem[Shannon(1948)]{shannon1948a-mathematical}
C.E. Shannon.
\newblock A mathematical theory of communication.
\newblock \emph{Bell system technical journal}, 27, 1948.

\bibitem[Shelhamer et~al.(2017)Shelhamer, Long, and
  Darrell]{shelhamer2017fully}
Evan Shelhamer, Jonathan Long, and Trevor Darrell.
\newblock Fully convolutional networks for semantic segmentation.
\newblock \emph{PAMI}, 2017.

\bibitem[Shu et~al.(2018)Shu, Bui, Narui, and Ermon]{shu2018dirt}
Rui Shu, Hung~H Bui, Hirokazu Narui, and Stefano Ermon.
\newblock A dirt-t approach to unsupervised domain adaptation.
\newblock In \emph{ICLR}, 2018.

\bibitem[Simonyan \& Zisserman(2015)Simonyan and Zisserman]{simonyan2014very}
Karen Simonyan and Andrew Zisserman.
\newblock Very deep convolutional networks for large-scale image recognition.
\newblock In \emph{ICLR}, 2015.

\bibitem[Sun et~al.(2017)Sun, Feng, and Saenko]{sun2017correlation}
Baochen Sun, Jiashi Feng, and Kate Saenko.
\newblock Correlation alignment for unsupervised domain adaptation.
\newblock In \emph{Domain Adaptation in Computer Vision Applications}, pp.\
  153--171. Springer, 2017.

\bibitem[Sun et~al.(2019{\natexlab{a}})Sun, Tzeng, Darrell, and
  Efros]{sun2019unsupervised}
Yu~Sun, Eric Tzeng, Trevor Darrell, and Alexei~A Efros.
\newblock Unsupervised domain adaptation through self-supervision.
\newblock \emph{arXiv preprint arXiv:1909.11825}, 2019{\natexlab{a}}.

\bibitem[Sun et~al.(2019{\natexlab{b}})Sun, Wang, Liu, Miller, Efros, and
  Hardt]{sun2019test}
Yu~Sun, Xiaolong Wang, Zhuang Liu, John Miller, Alexei~A Efros, and Moritz
  Hardt.
\newblock Test-time training for out-of-distribution generalization.
\newblock \emph{arXiv preprint arXiv:1909.13231}, 2019{\natexlab{b}}.

\bibitem[Szegedy et~al.(2014)Szegedy, Zaremba, Sutskever, Bruna, Erhan,
  Goodfellow, and Fergus]{szegedy2014intriguing}
Christian Szegedy, Wojciech Zaremba, Ilya Sutskever, Joan Bruna, Dumitru Erhan,
  Ian Goodfellow, and Rob Fergus.
\newblock Intriguing properties of neural networks.
\newblock 2014.

\bibitem[Tzeng et~al.(2015)Tzeng, Hoffman, Darrell, and
  Saenko]{tzeng2015simultaneous}
Eric Tzeng, Judy Hoffman, Trevor Darrell, and Kate Saenko.
\newblock Simultaneous deep transfer across domains and tasks.
\newblock In \emph{ICCV}, 2015.

\bibitem[Tzeng et~al.(2017)Tzeng, Hoffman, Saenko, and
  Darrell]{tzeng2017adversarial}
Eric Tzeng, Judy Hoffman, Kate Saenko, and Trevor Darrell.
\newblock Adversarial discriminative domain adaptation.
\newblock In \emph{CVPR}, 2017.

\bibitem[Wang et~al.(2020)Wang, Sun, Cheng, Jiang, Deng, Zhao, Liu, Mu, Tan,
  Wang, et~al.]{wang2020deep}
Jingdong Wang, Ke~Sun, Tianheng Cheng, Borui Jiang, Chaorui Deng, Yang Zhao,
  Dong Liu, Yadong Mu, Mingkui Tan, Xinggang Wang, et~al.
\newblock Deep high-resolution representation learning for visual recognition.
\newblock \emph{PAMI}, 2020.

\bibitem[Wu \& He(2018)Wu and He]{wu2018group}
Yuxin Wu and Kaiming He.
\newblock Group normalization.
\newblock In \emph{ECCV}, 2018.

\bibitem[Yosinski et~al.(2014)Yosinski, Clune, Bengio, and
  Lipson]{yosinski2014transferable}
Jason Yosinski, Jeff Clune, Yoshua Bengio, and Hod Lipson.
\newblock How transferable are features in deep neural networks?
\newblock In \emph{NeurIPS}, 2014.

\bibitem[Zhang et~al.(2017)Zhang, Isola, and Efros]{zhang2020split-brain}
Richard Zhang, Phillip Isola, and Alexei~A Efros.
\newblock Split-brain autoencoders: Unsupervised learning by cross-channel
  prediction.
\newblock In \emph{CVPR}, 2017.

\bibitem[Zhao et~al.(2020)Zhao, Jia, and Koltun]{zhao2020exploring}
Hengshuang Zhao, Jiaya Jia, and Vladlen Koltun.
\newblock Exploring self-attention for image recognition.
\newblock In \emph{CVPR}, 2020.

\bibitem[Zhou et~al.(2004)Zhou, Bousquet, Lal, Weston, and
  Sch{\"o}lkopf]{zhou2004learning}
Dengyong Zhou, Olivier Bousquet, Thomas~Navin Lal, Jason Weston, and Bernhard
  Sch{\"o}lkopf.
\newblock Learning with local and global consistency.
\newblock \emph{NeurIPS}, 2004.

\end{thebibliography}
}

\newpage
\appendix

\section*{Appendix}

This supplement summarizes the image corruptions used in our experiments, highlights a qualitative example of instance-wise adaptation for semantic segmentation, and visualizes feature shifts across more layers.

\section{Robustness to Corruptions}
\label{app:corruption}

\begin{figure}[h]
    \centering
    \includegraphics[width=\linewidth]{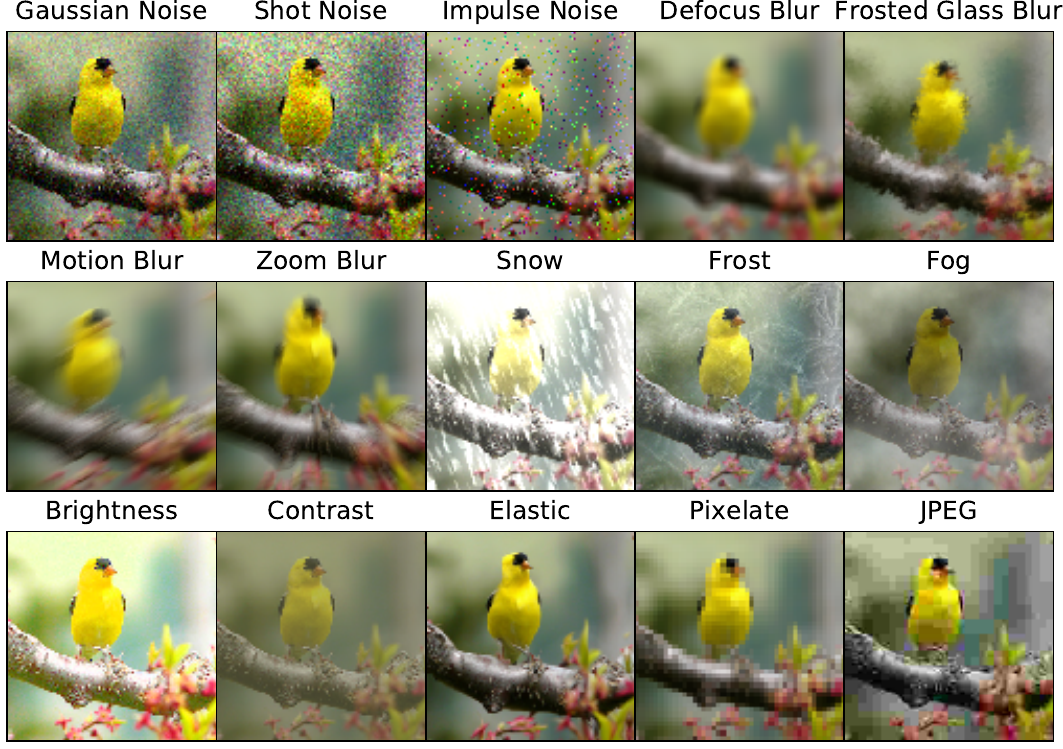}
\vspace{0.5mm}
\caption{
Examples of each corruption type in the image corruptions benchmark. 
While synthetic, this set of corruptions aims to represent natural factors of variation like noise, blur, weather, and digital imaging effects.
This figure is reproduced from \cite{hendrycks2019benchmarking}.
}
\label{fig:corruptions}
\end{figure}

In Section \ref{sec:corrupt} we evaluate methods on a common image corruptions benchmark.
Table \ref{tab:corruption} reports errors on the most severe level of corruption, level 5, and Figure \ref{fig:in-c} reports errors for each corruption type averaged across each of the levels 1--5.
We summarize these corruptions types by example in Figure \ref{fig:corruptions}.

\section{Source-Free Adaptation for Semantic Segmentation}
\label{app:da}

\begin{figure}[h]
\minipage{0.49\textwidth}
    \centering
    \includegraphics[width=\linewidth]{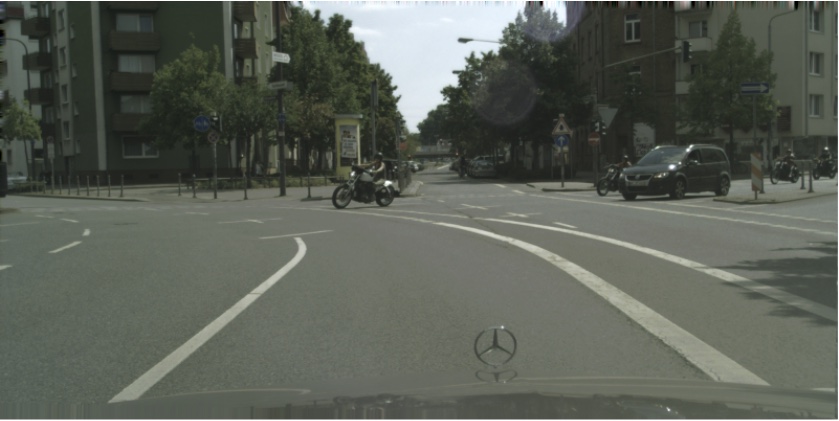}\\
    image\\
    \includegraphics[width=\linewidth]{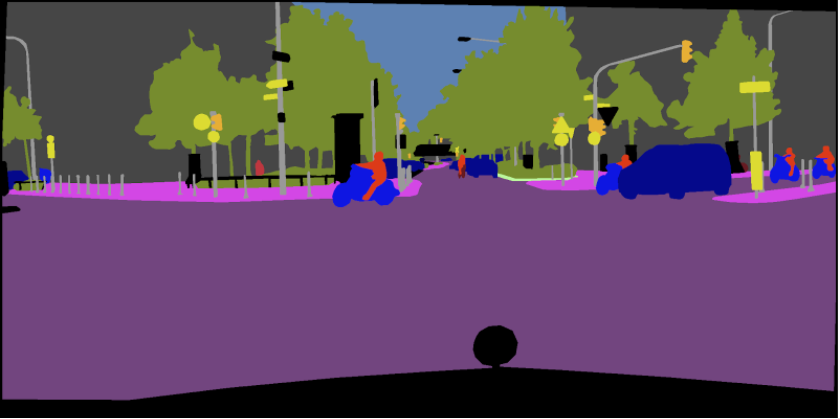}\\
    label\\
    \includegraphics[width=\linewidth]{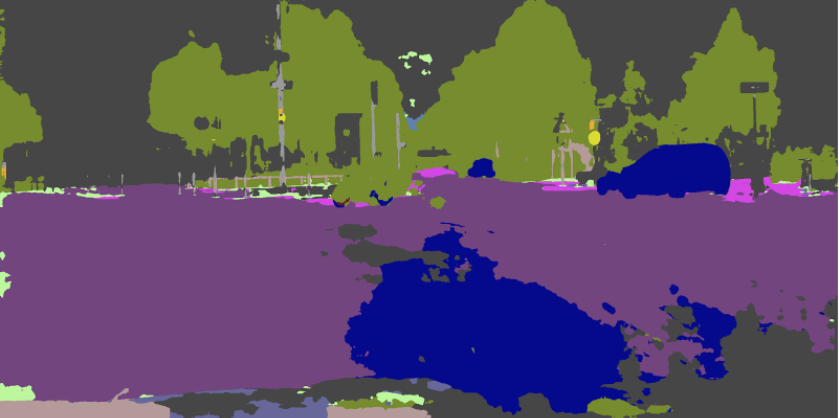}\\
    source-only\\
\endminipage\hfill
\minipage{0.49\textwidth}
    \centering
    \includegraphics[width=\linewidth]{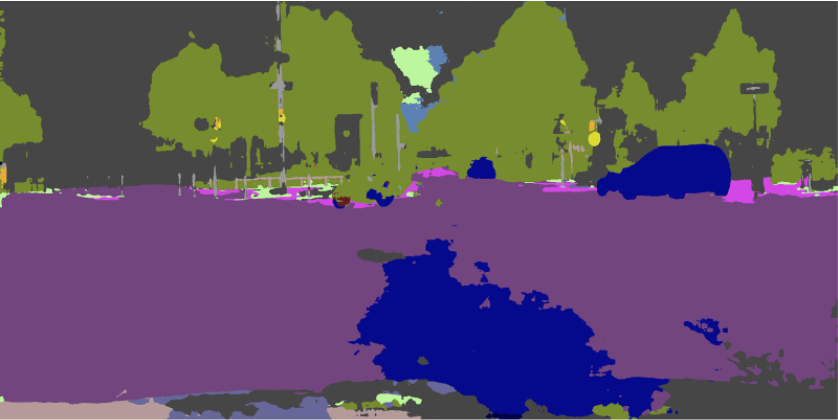}\\
    tent, iteration 1\\
    \includegraphics[width=\linewidth]{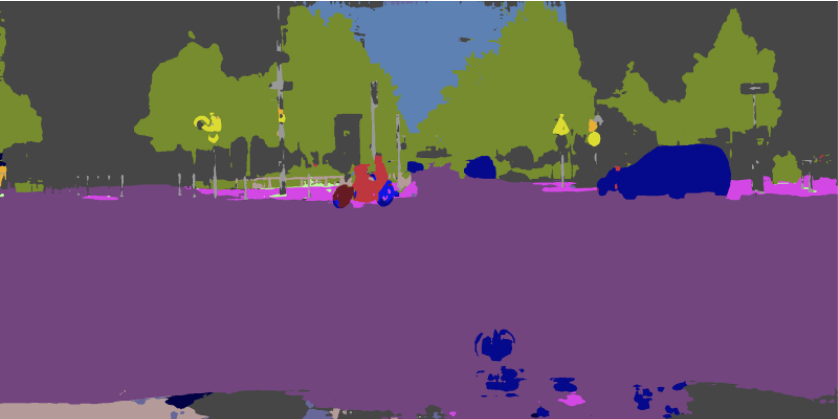}\\
    tent, iteration 5\\
    \includegraphics[width=\linewidth]{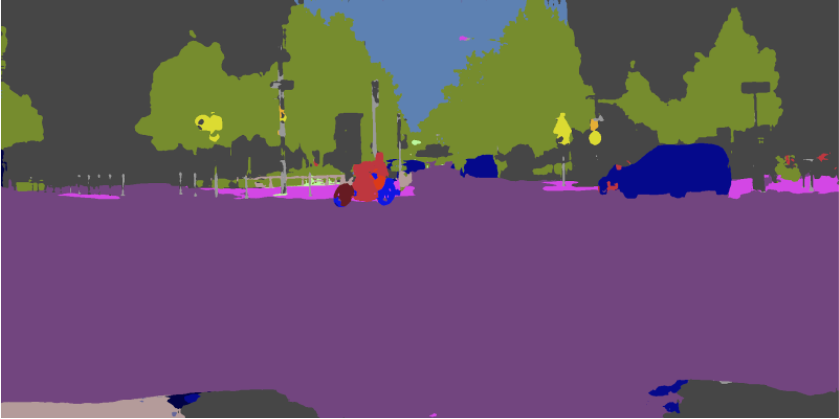}\\
    tent, iteration 10\\
\endminipage
\vspace{0.5mm}
\caption{
Adaptation for semantic segmentation with simulation-to-real shift from GTA \cite{richter2017playing} to Cityscapes \cite{cordts2016cityscapes}.
Tent only uses the target data, and optimizes over a single image as a dataset of pixel-wise predictions.
This episodic optimization in effect fits a custom model to each image of the target domain.
In only 10 iterations our method suppresses noise (see the completion of the street segment, in purple) and recovers missing classes (see the motorcycle and rider, center).
}
\label{fig:sim2real-da}
\end{figure}

Figure \ref{fig:sim2real-da} shows a qualitative result on source-free adaptation for semantic segmentation (pixel-wise classification) with simulation-to-real (sim-to-real) shift.

For this sim-to-real condition, the source data is simulated while the target data is real.
Our source data is GTA \cite{richter2017playing}, a visually-sophisticated video game set in an urban environment, and our target data is Cityscapes \cite{cordts2016cityscapes}, an urban autonomous driving dataset.
The supervised model is HRnet-W18, a fully convolutional network \cite{shelhamer2017fully} in the high-resolution network family \cite{wang2020deep}.
For this qualitative example, we run tent on a single image for multiple iterations, because an image is in effect a batch of pixels.
This demonstrates adaptation to a target \emph{instance}, without any further access to the target \emph{domain} through usage of multiple images from the target distribution.

\clearpage

\section{Feature Shifts across Layers and Methods}
\label{sec:app-features}

\begin{figure}[h]
\centering
\adjustbox{max width=0.95\columnwidth}{
\begin{tabular}{l c c c c}
& \resizebox{0.8\textwidth}{!}{(a) Source} & \resizebox{0.55\textwidth}{!}{(b) BN} & \resizebox{0.63\textwidth}{!}{(c) Tent} & \resizebox{0.8\textwidth}{!}{(d) Oracle} \\
& & & & \\
\resizebox{0.12\textwidth}{!}{\rotatebox{90}{ Layer 2}} &
\includegraphics[trim=55 55 55 55, clip, keepaspectratio=false, width=222mm, height=111mm]{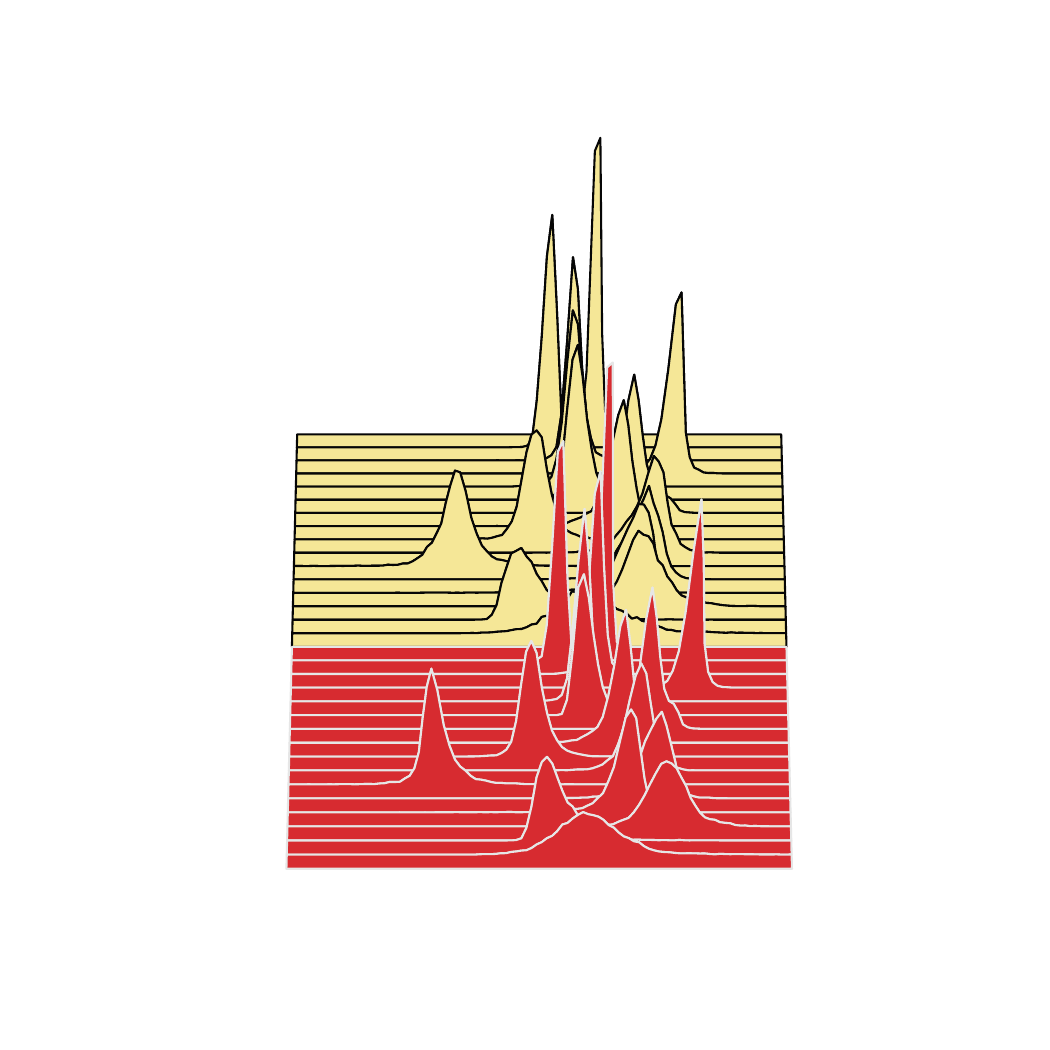} &
\includegraphics[trim=55 55 55 55, clip, keepaspectratio=false, width=222mm, height=111mm]{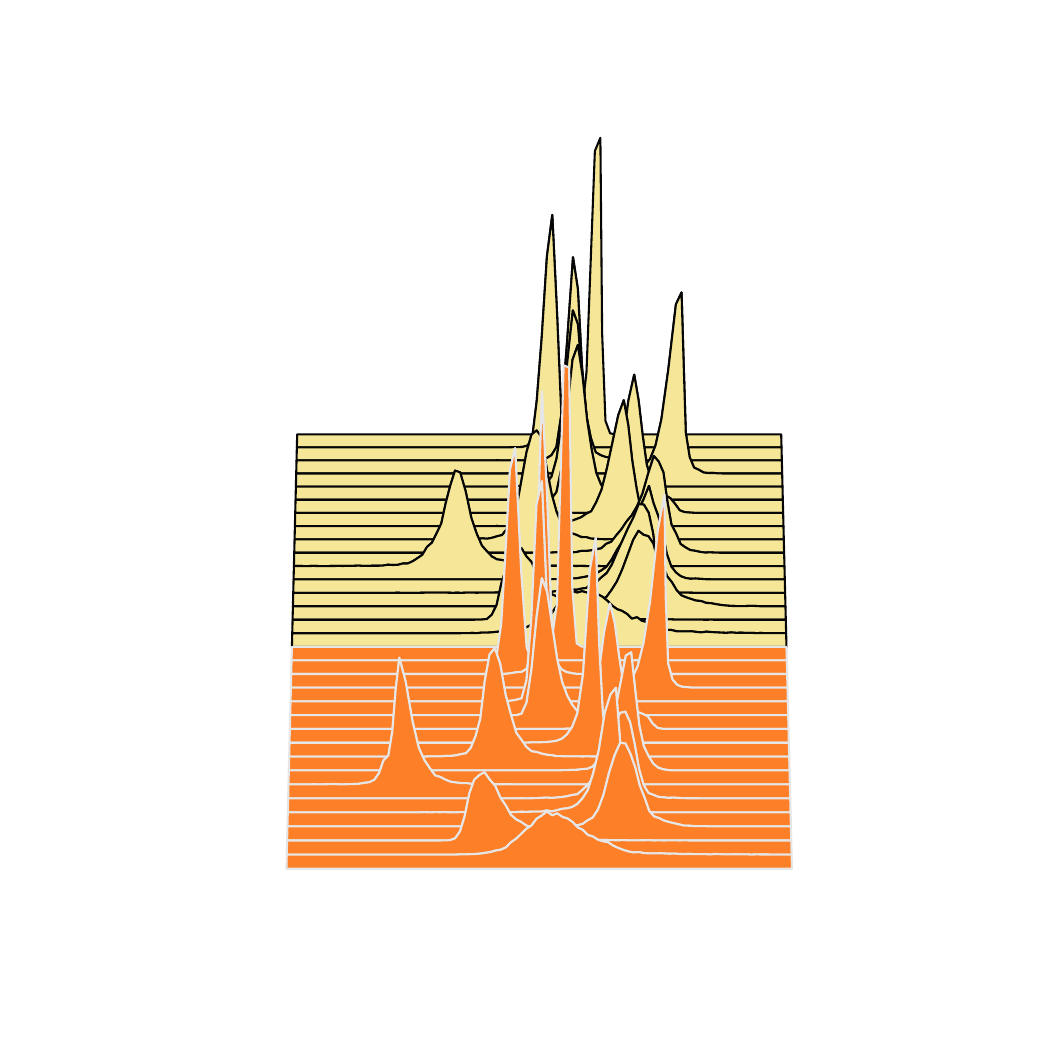} &
\includegraphics[trim=55 55 55 55, clip, keepaspectratio=false, width=222mm, height=111mm]{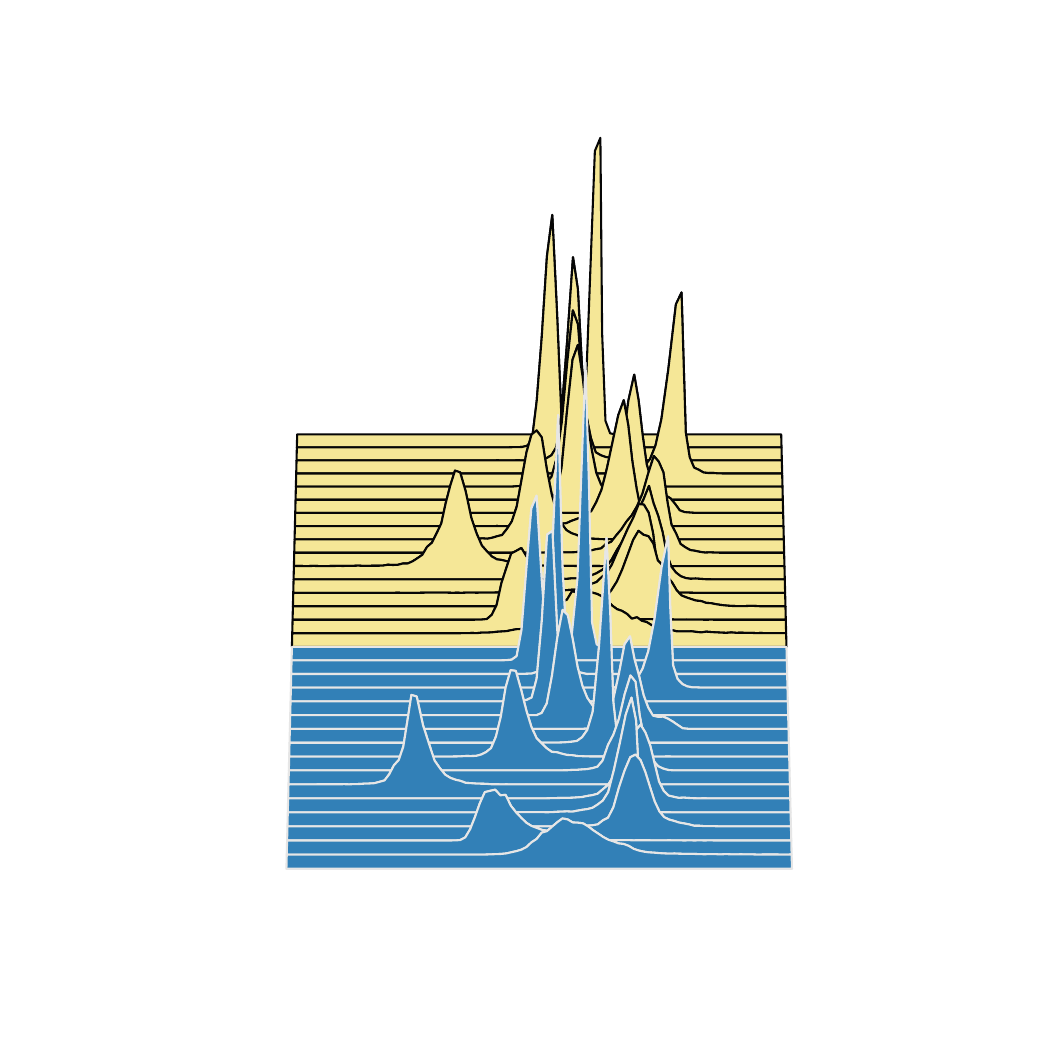} &
\includegraphics[trim=55 55 55 55, clip, keepaspectratio=false, width=222mm, height=111mm]{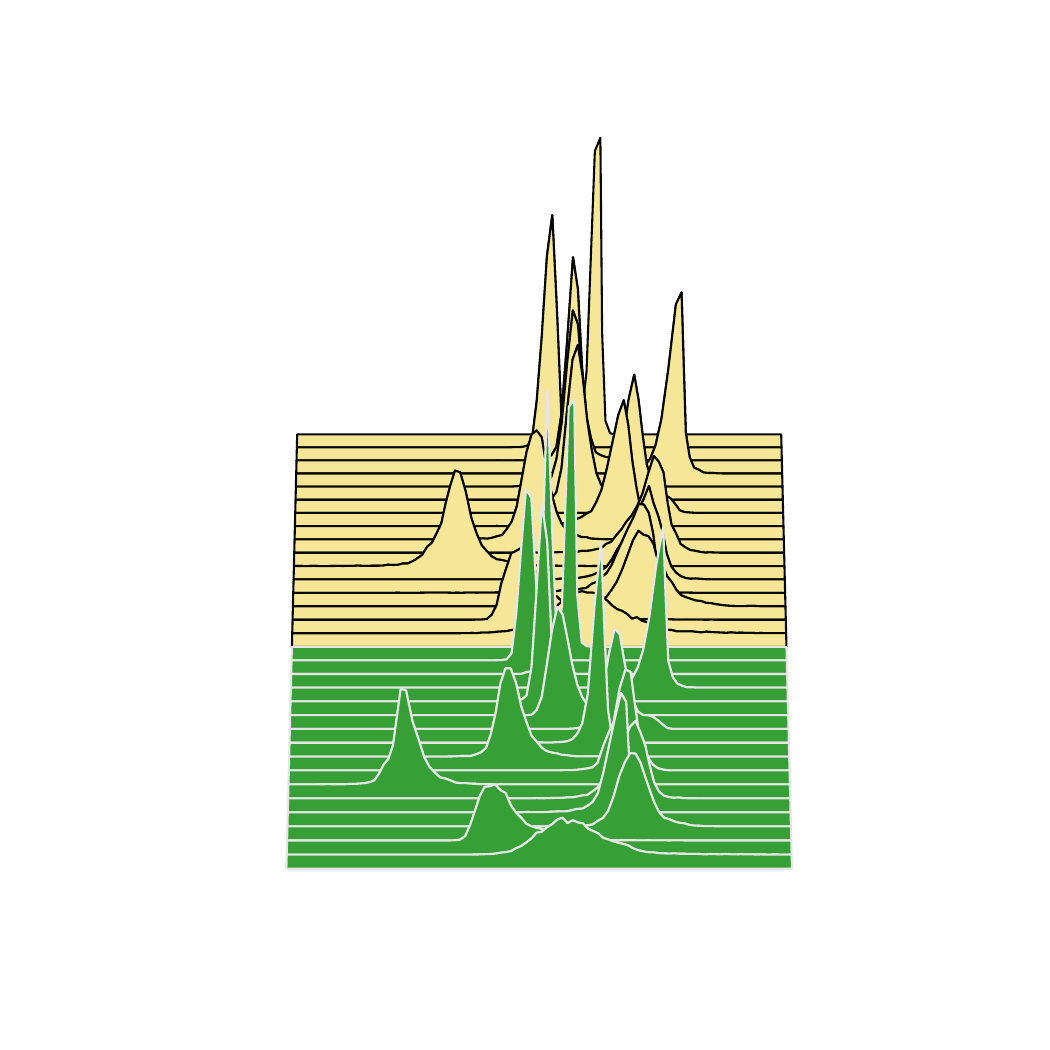} \\
\resizebox{0.12\textwidth}{!}{\rotatebox{90}{ Layer 5}} &
\includegraphics[trim=55 55 55 55, clip, keepaspectratio=false, width=222mm, height=111mm]{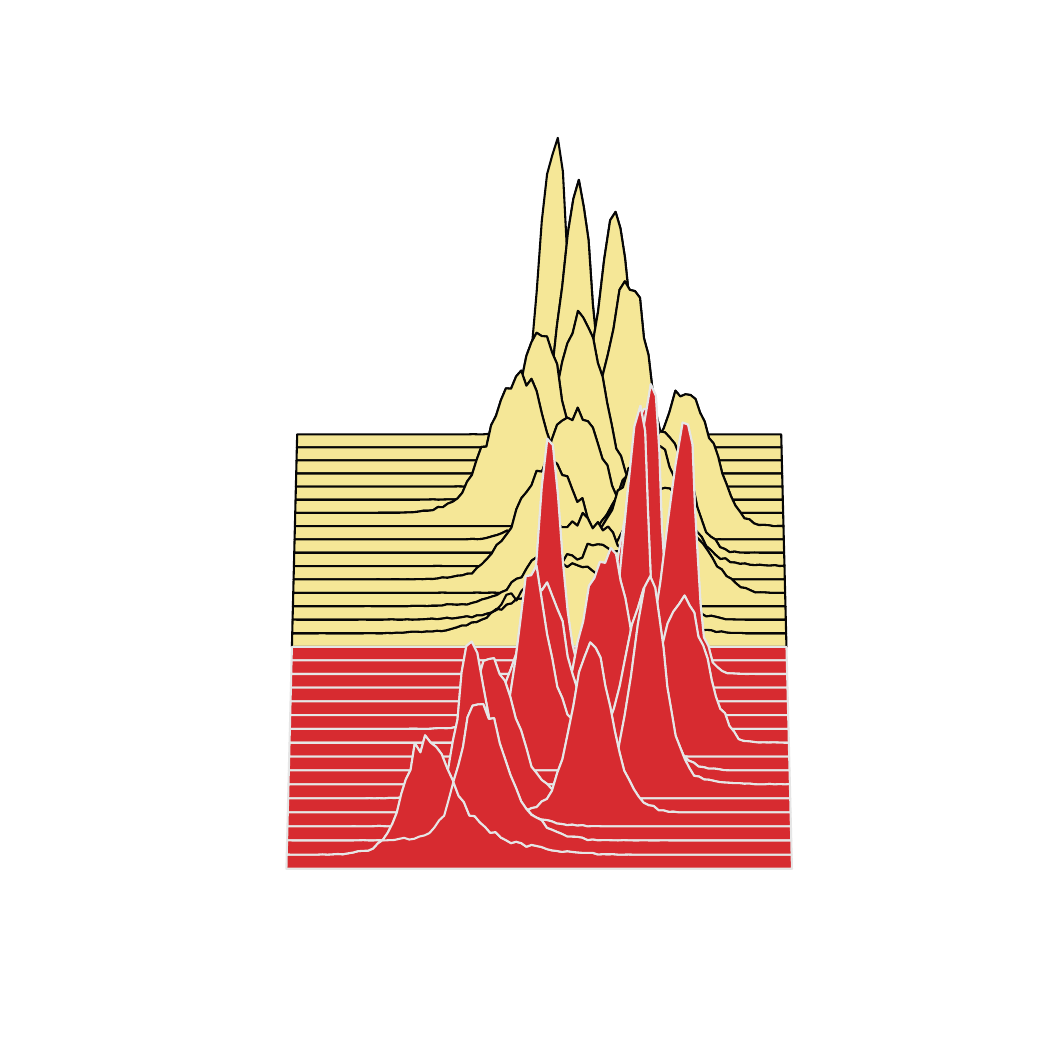} &
\includegraphics[trim=55 55 55 55, clip, keepaspectratio=false, width=222mm, height=111mm]{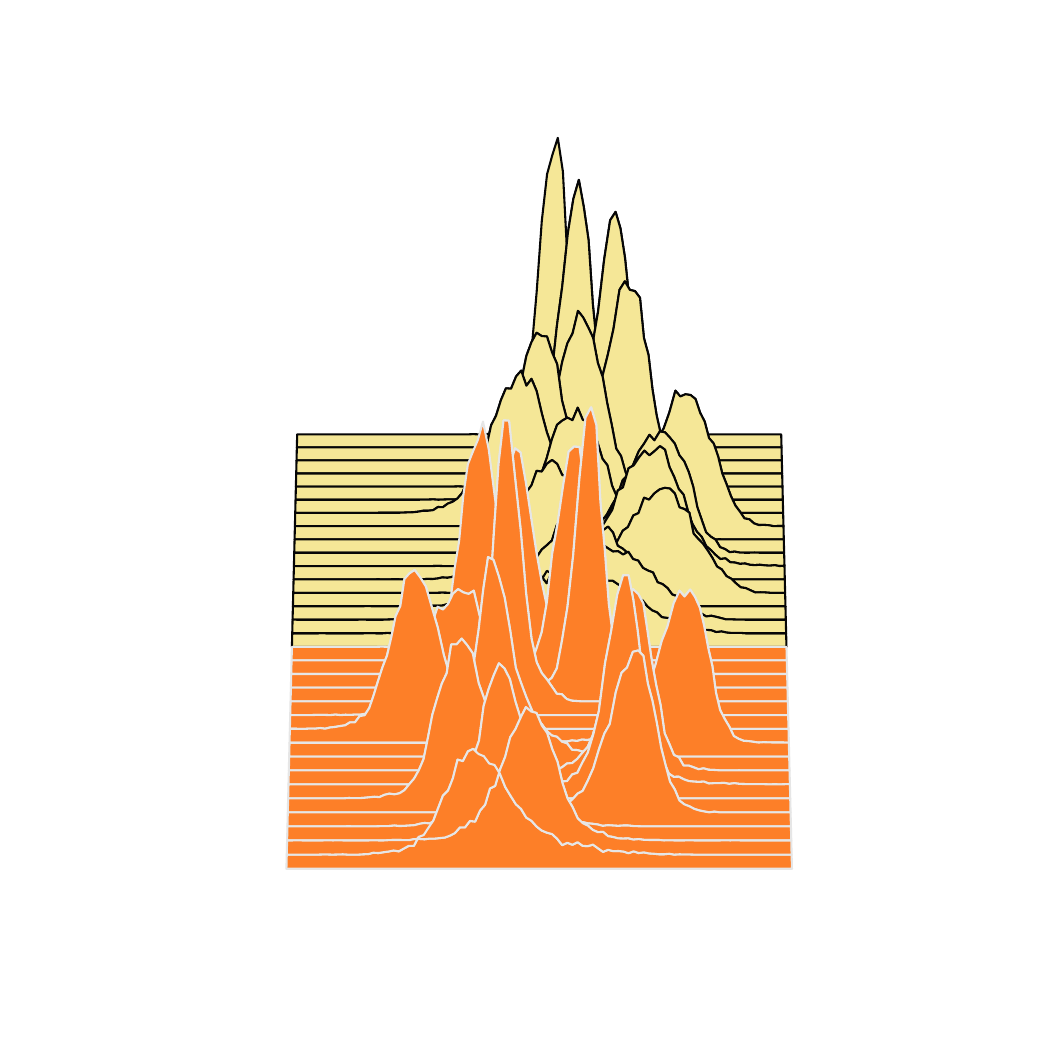} &
\includegraphics[trim=55 55 55 55, clip, keepaspectratio=false, width=222mm, height=111mm]{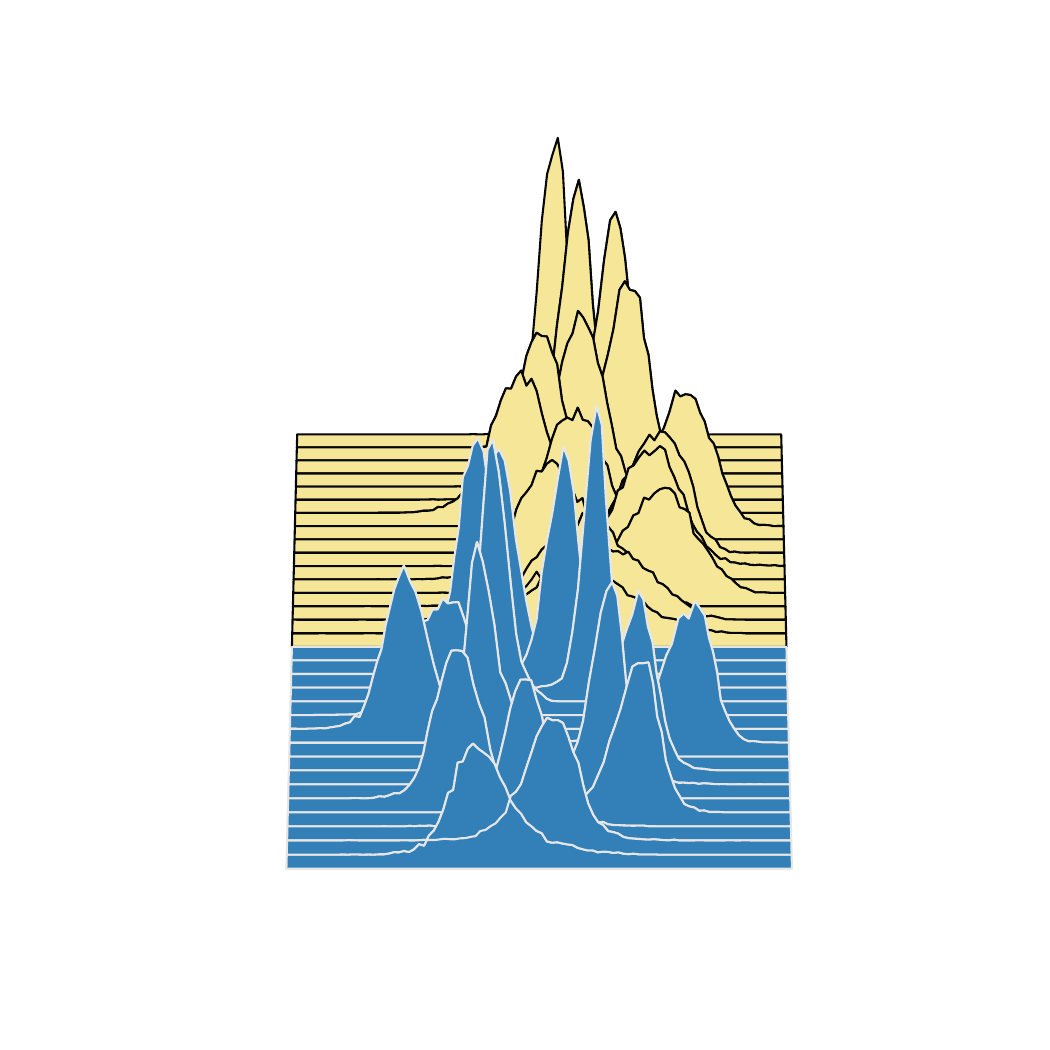} &
\includegraphics[trim=55 55 55 55, clip, keepaspectratio=false, width=222mm, height=111mm]{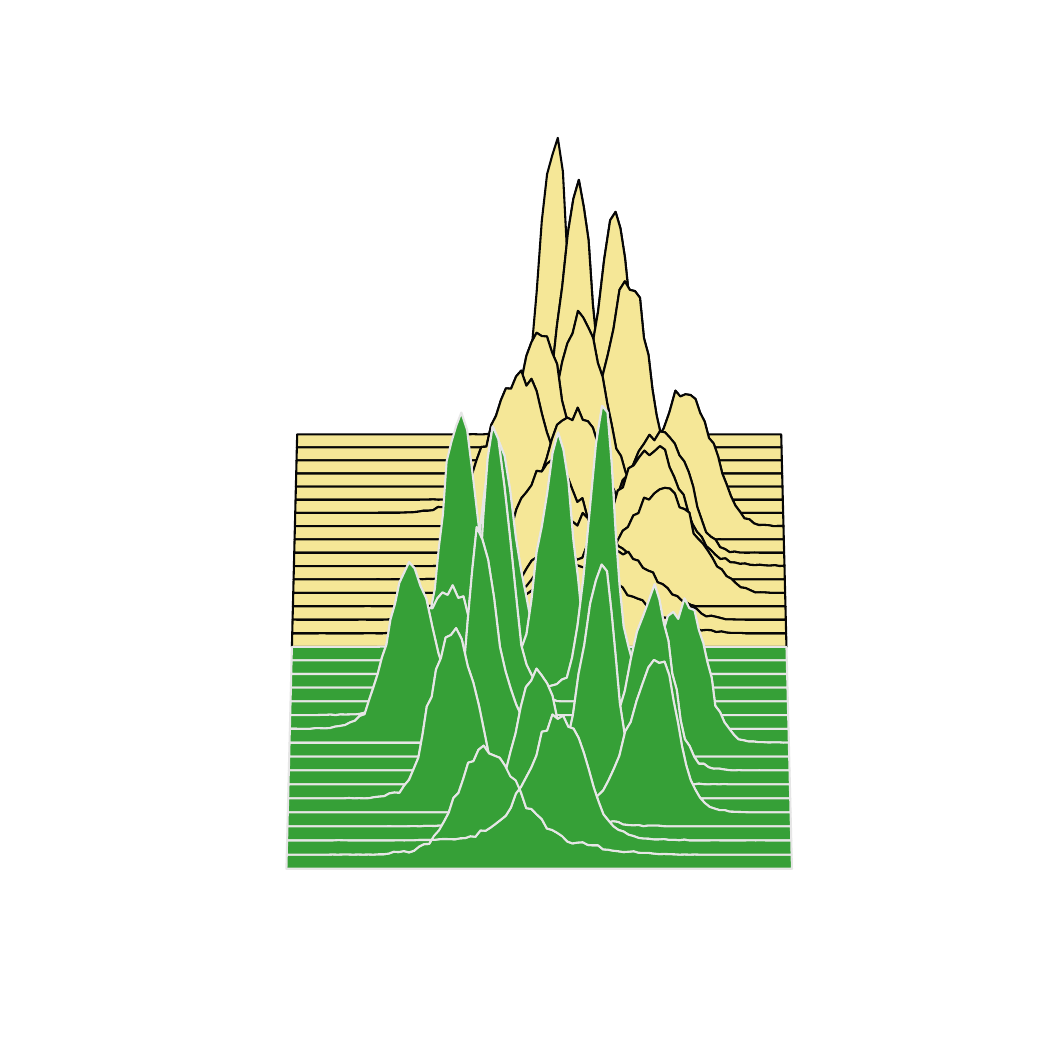} \\
\resizebox{0.12\textwidth}{!}{\rotatebox{90}{ Layer 8}} &
\includegraphics[trim=55 55 55 55, clip, keepaspectratio=false, width=222mm, height=111mm]{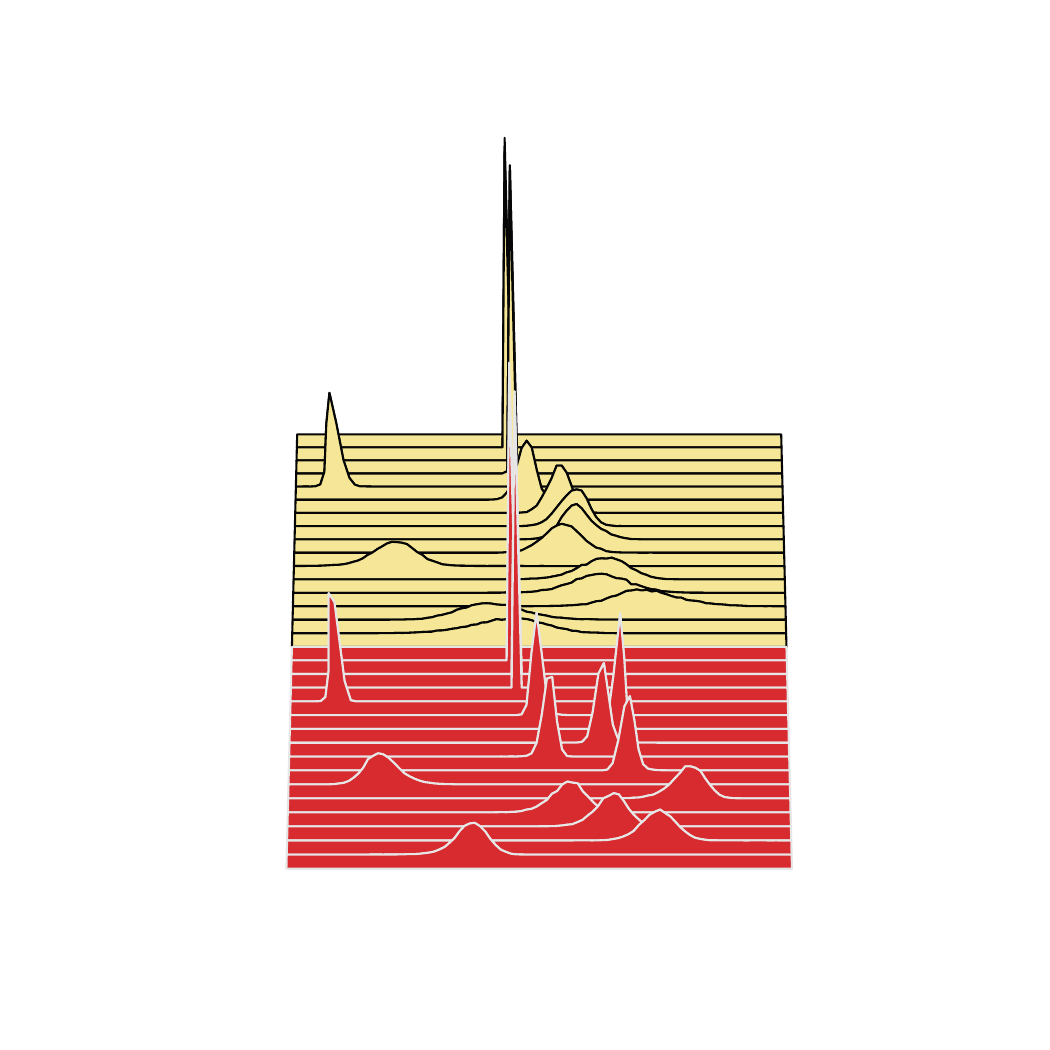} &
\includegraphics[trim=55 55 55 55, clip, keepaspectratio=false, width=222mm, height=111mm]{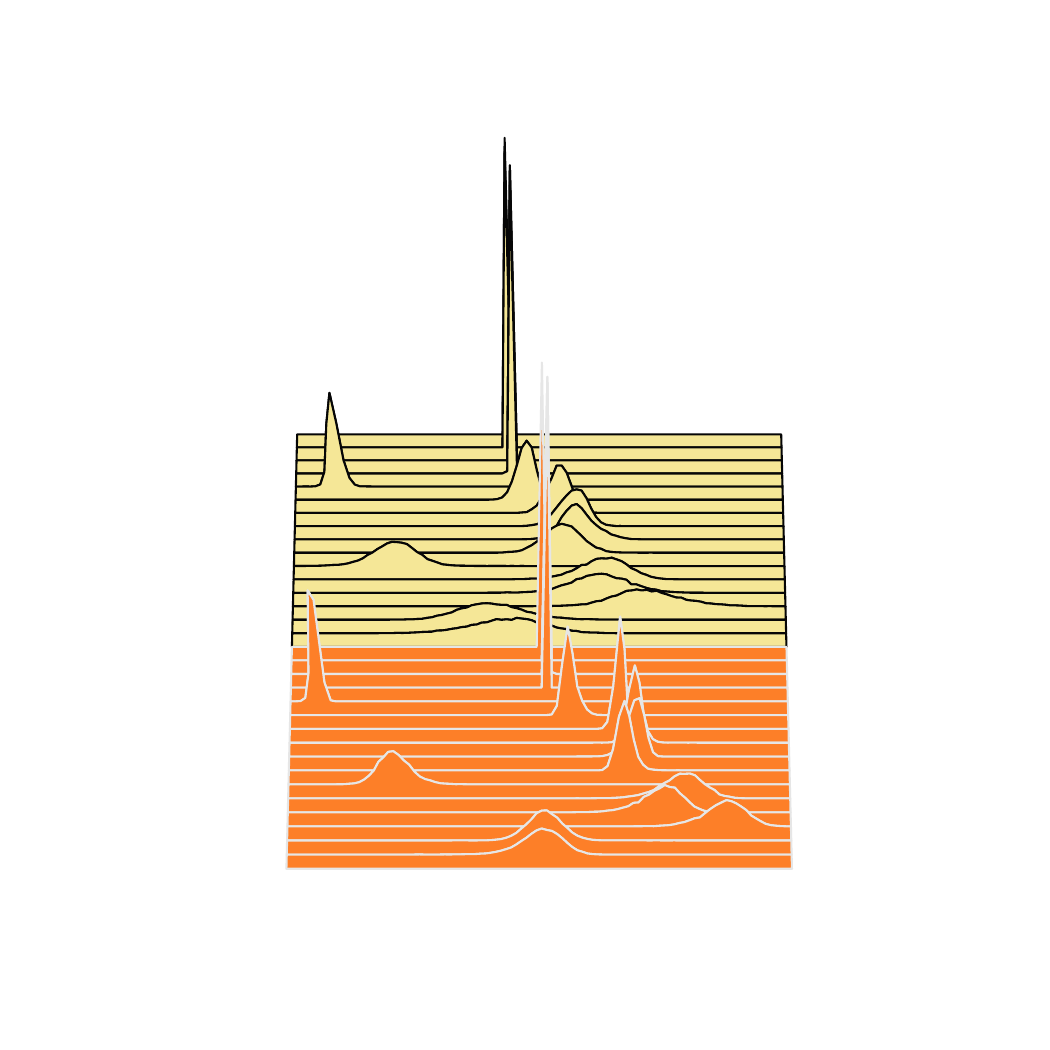} &
\includegraphics[trim=55 55 55 55, clip, keepaspectratio=false, width=222mm, height=111mm]{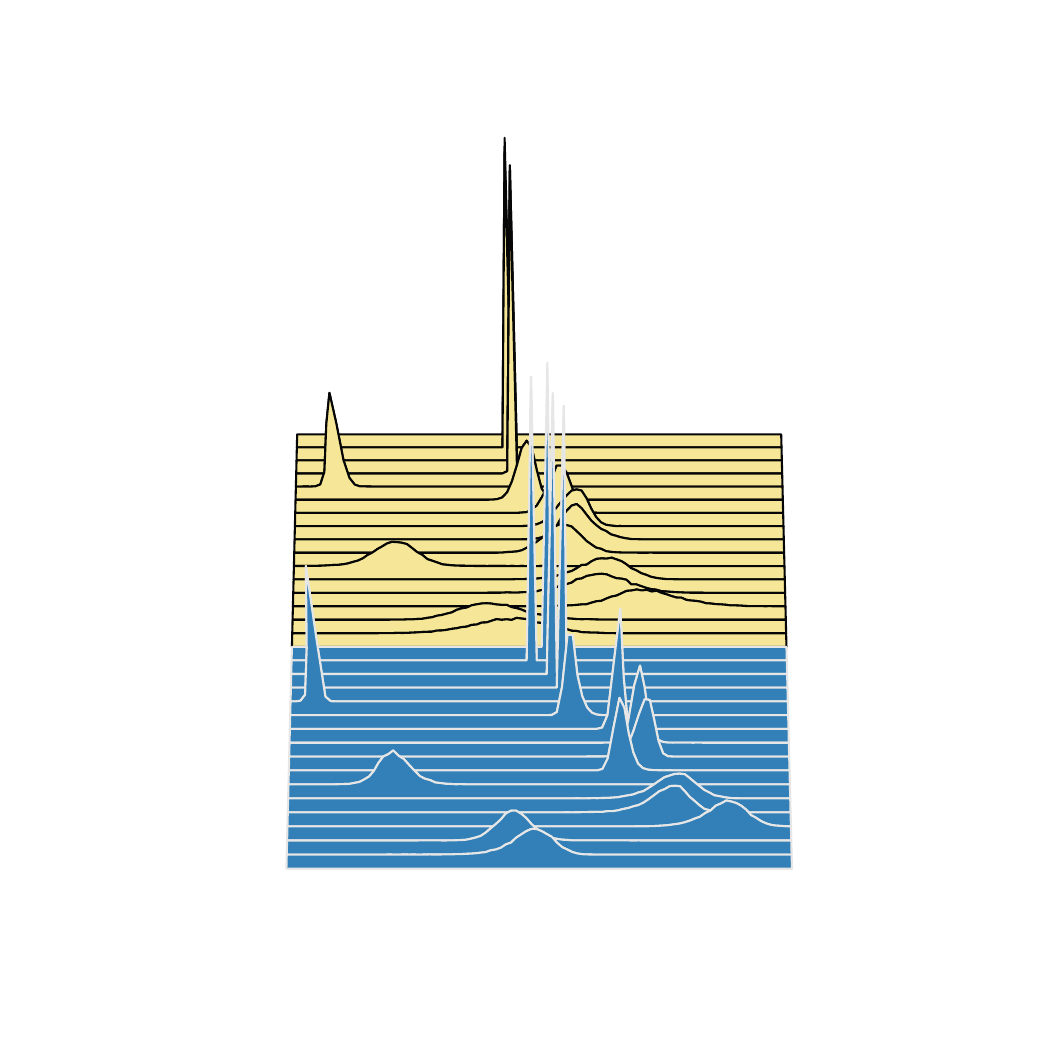} &
\includegraphics[trim=55 55 55 55, clip, keepaspectratio=false, width=222mm, height=111mm]{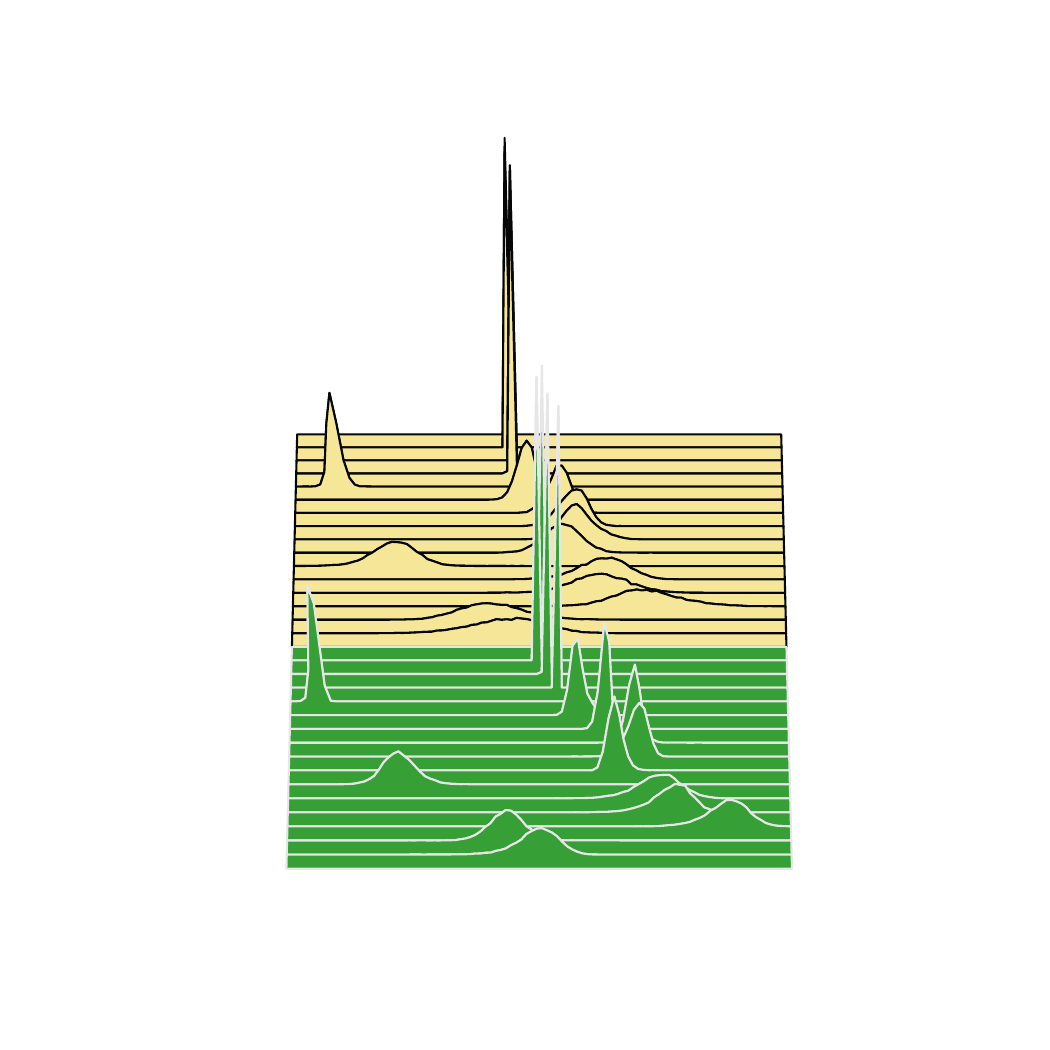} \\
\resizebox{0.12\textwidth}{!}{\rotatebox{90}{ Layer 11}} &
\includegraphics[trim=120 55 65 55, clip, keepaspectratio=false, width=222mm, height=111mm]{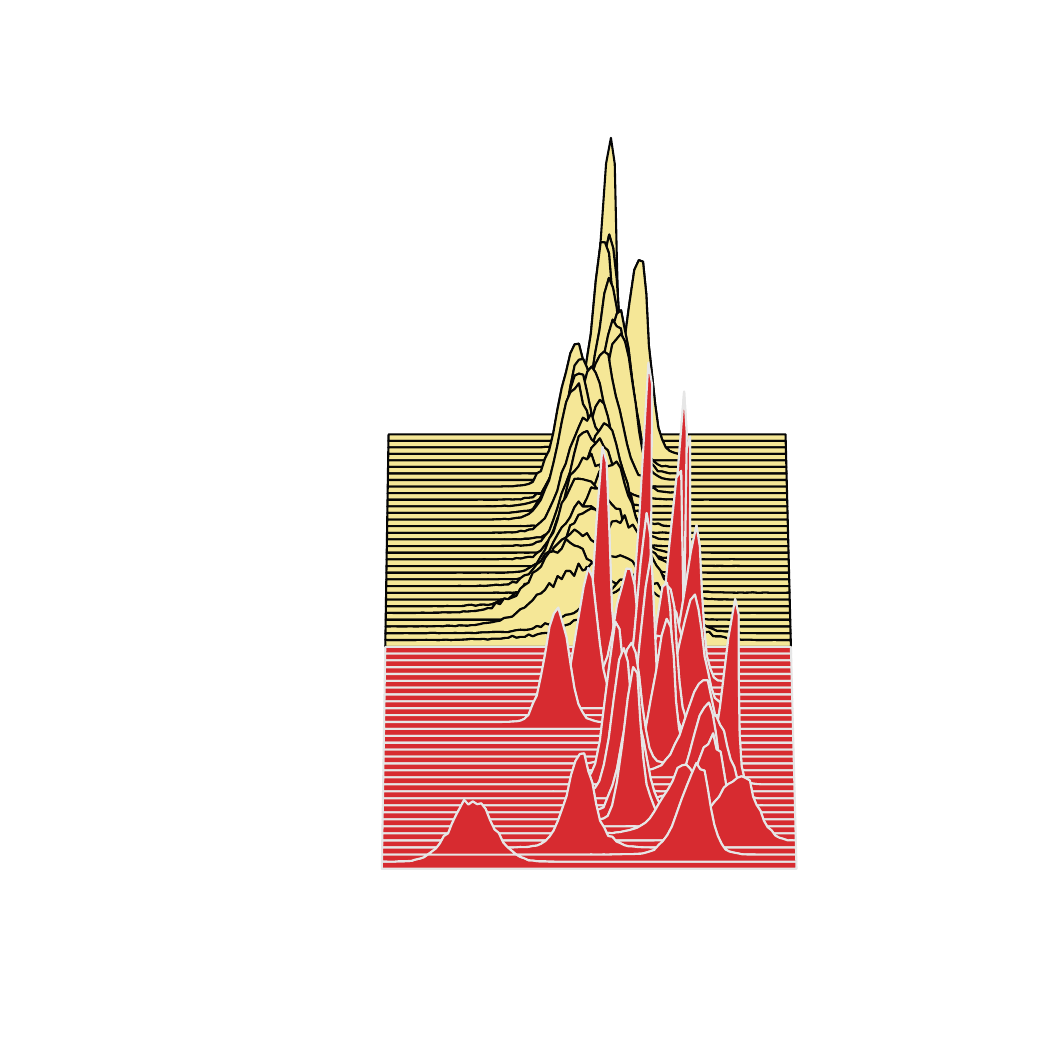} &
\includegraphics[trim=120 55 65 55, clip, keepaspectratio=false, width=222mm, height=111mm]{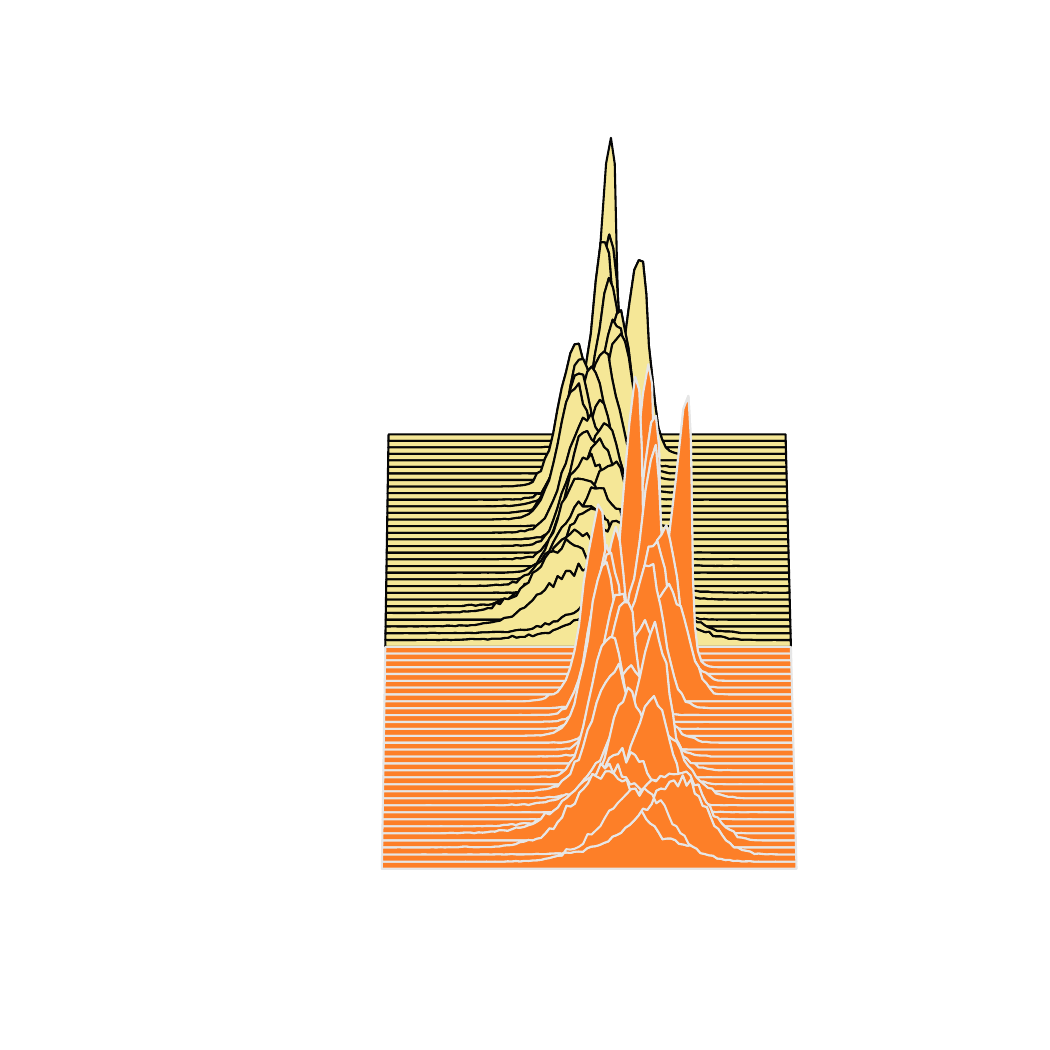} &
\includegraphics[trim=120 55 65 55, clip, keepaspectratio=false, width=222mm, height=111mm]{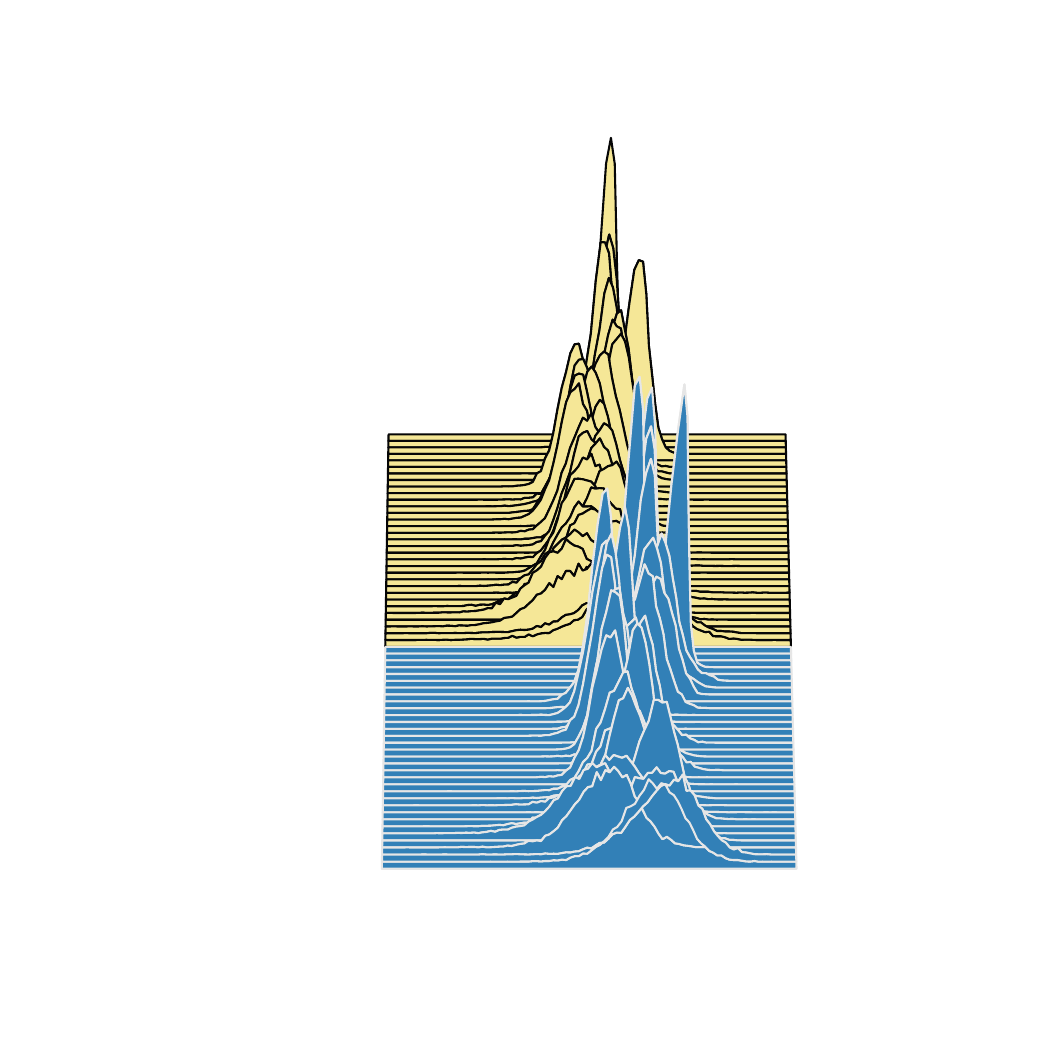} &
\includegraphics[trim=120 55 65 55, clip, keepaspectratio=false, width=222mm, height=111mm]{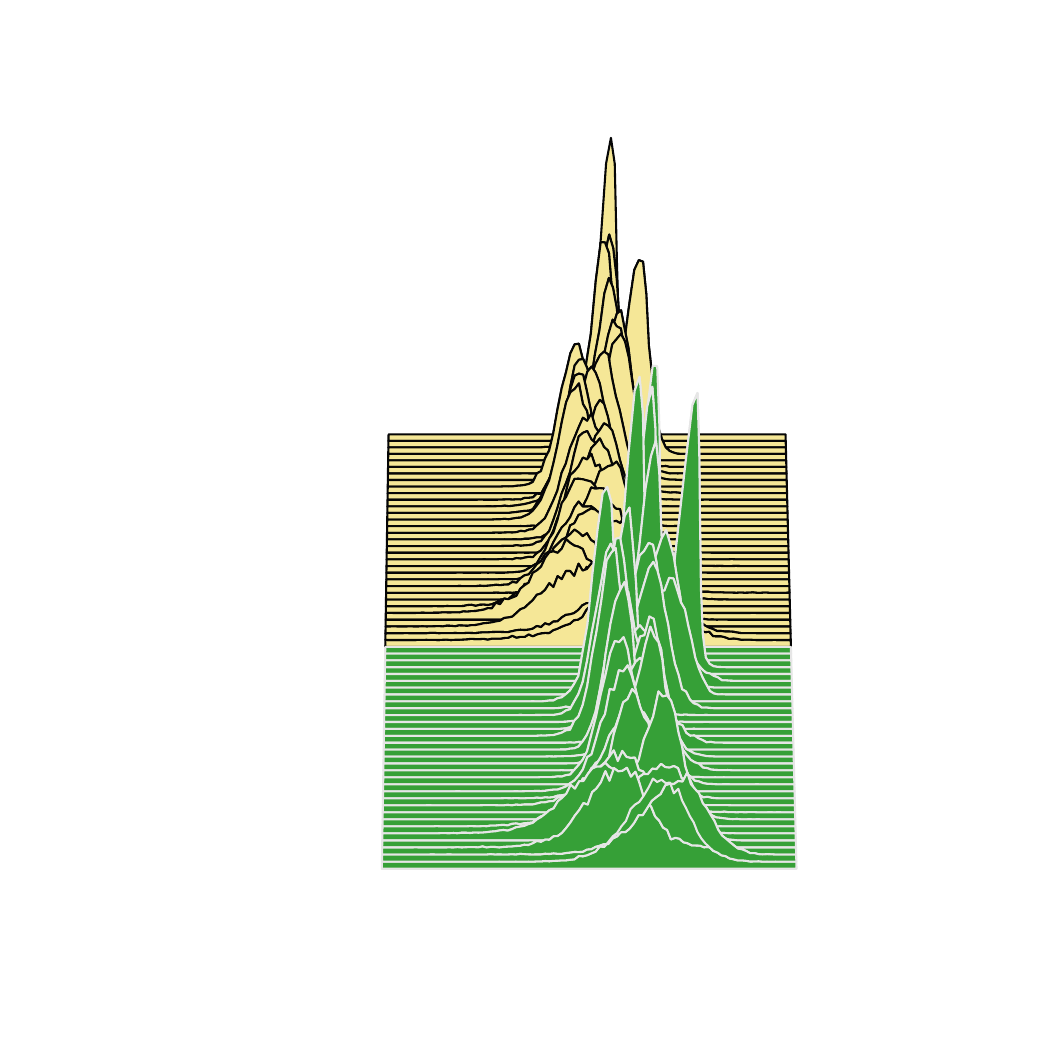} \\
\resizebox{0.12\textwidth}{!}{\rotatebox{90}{ Layer 14}} &
\includegraphics[trim=55 55 55 55, clip, keepaspectratio=false, width=222mm, height=111mm]{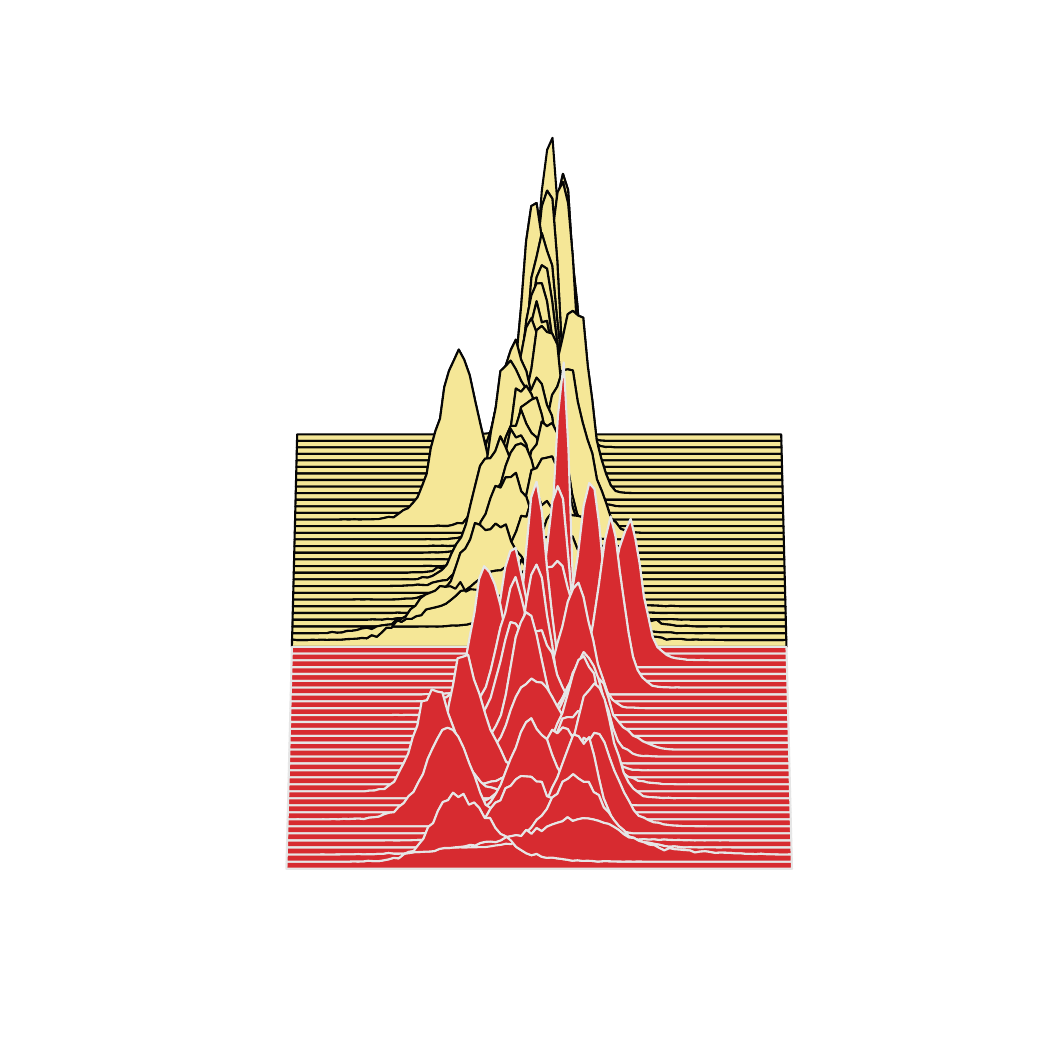} &
\includegraphics[trim=55 55 55 55, clip, keepaspectratio=false, width=222mm, height=111mm]{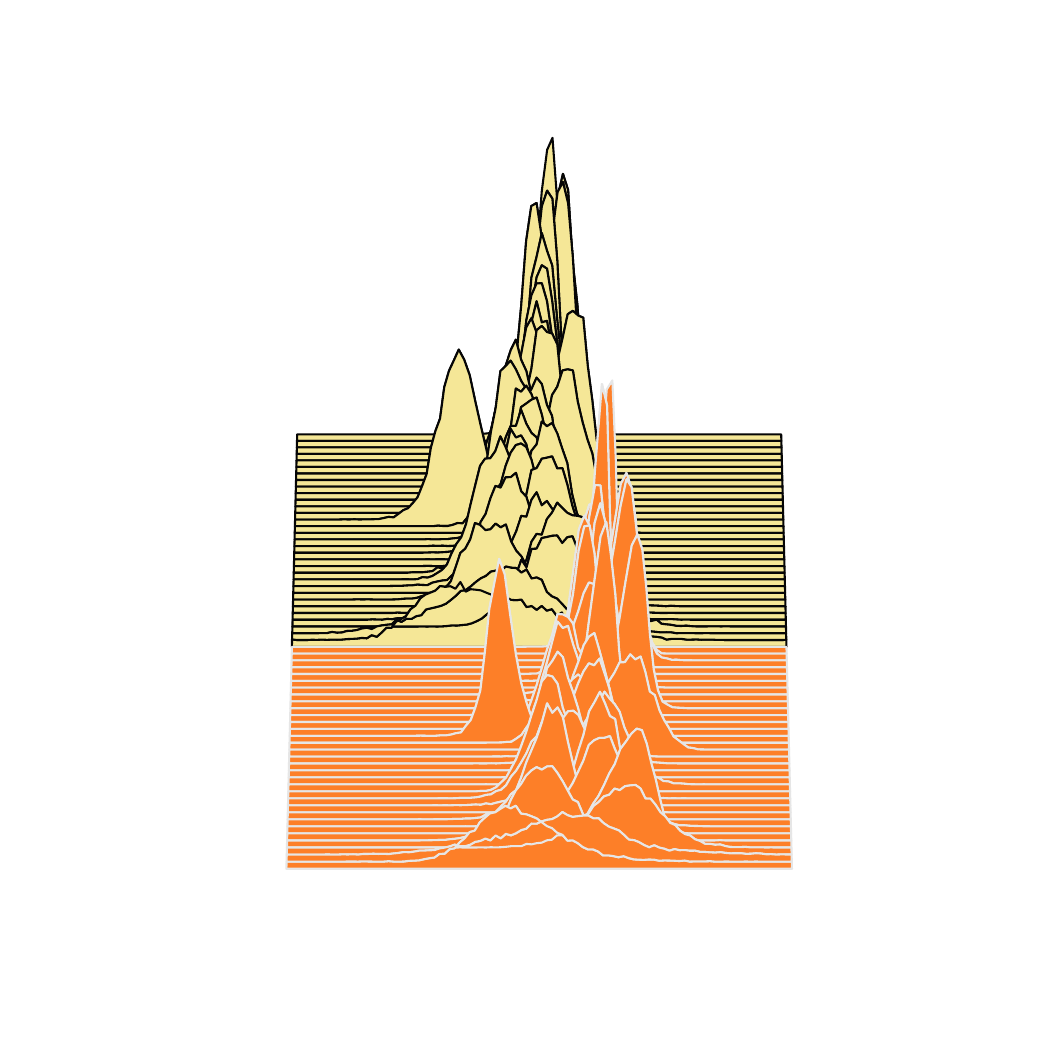} &
\includegraphics[trim=55 55 55 55, clip, keepaspectratio=false, width=222mm, height=111mm]{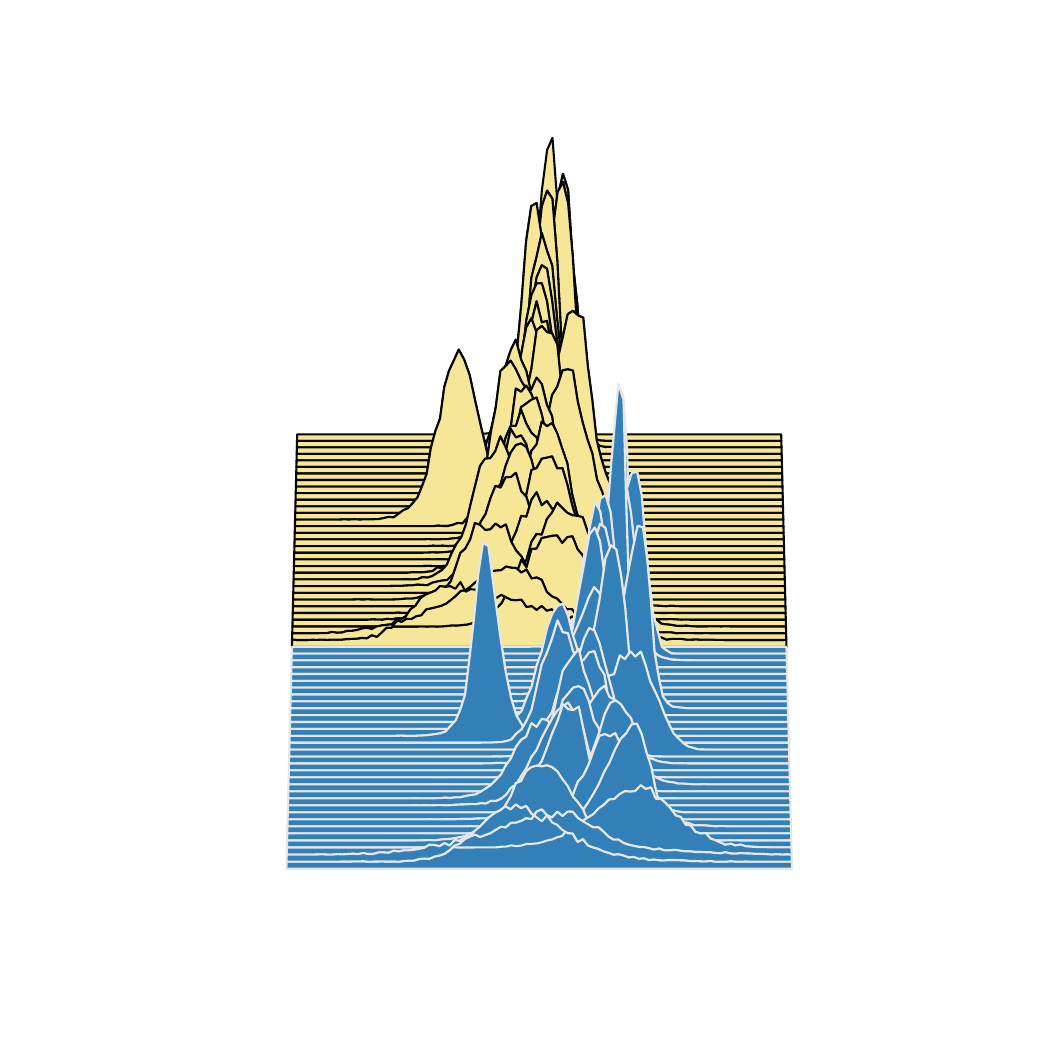} &
\includegraphics[trim=55 55 55 55, clip, keepaspectratio=false, width=222mm, height=111mm]{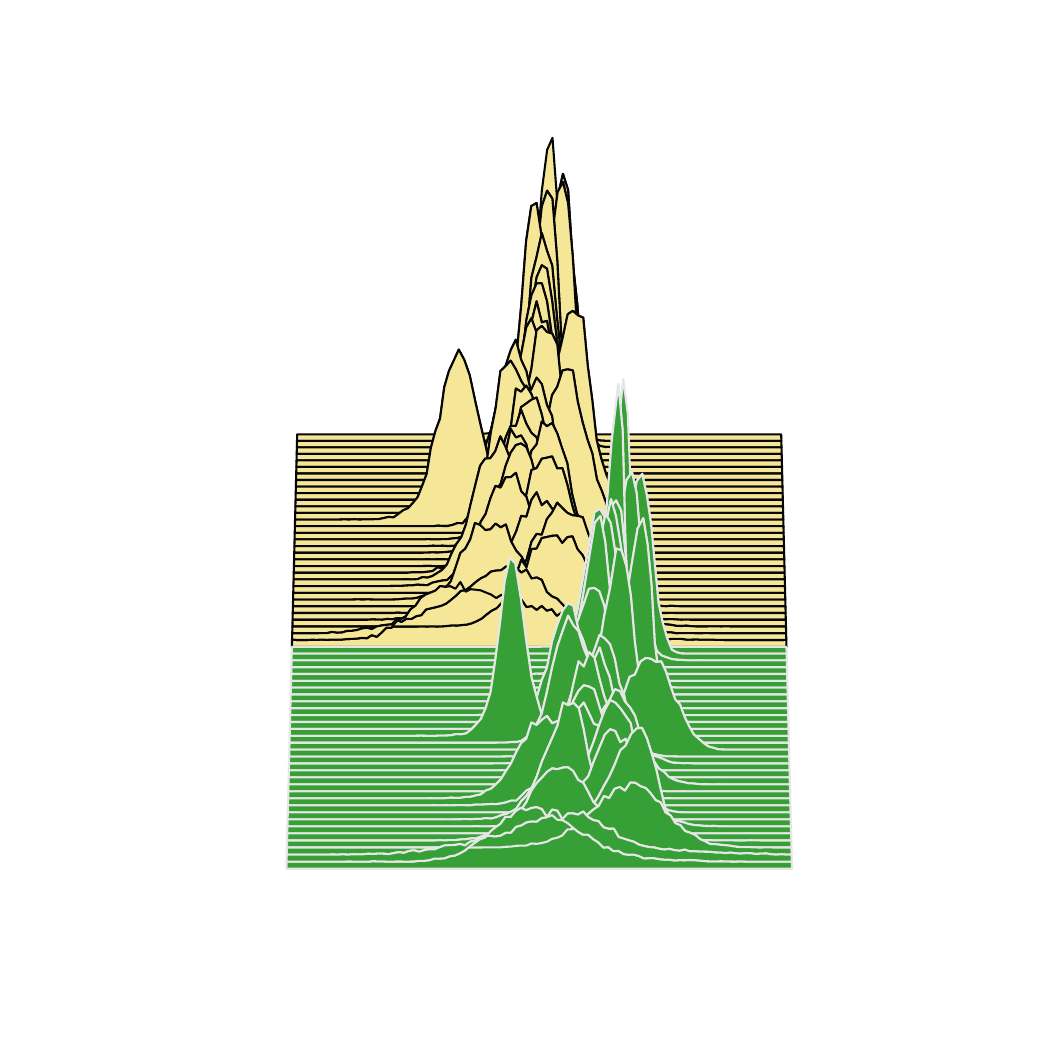} \\
\resizebox{0.12\textwidth}{!}{\rotatebox{90}{ Layer 18}} &
\includegraphics[trim=55 55 55 55, clip, keepaspectratio=false, width=222mm, height=111mm]{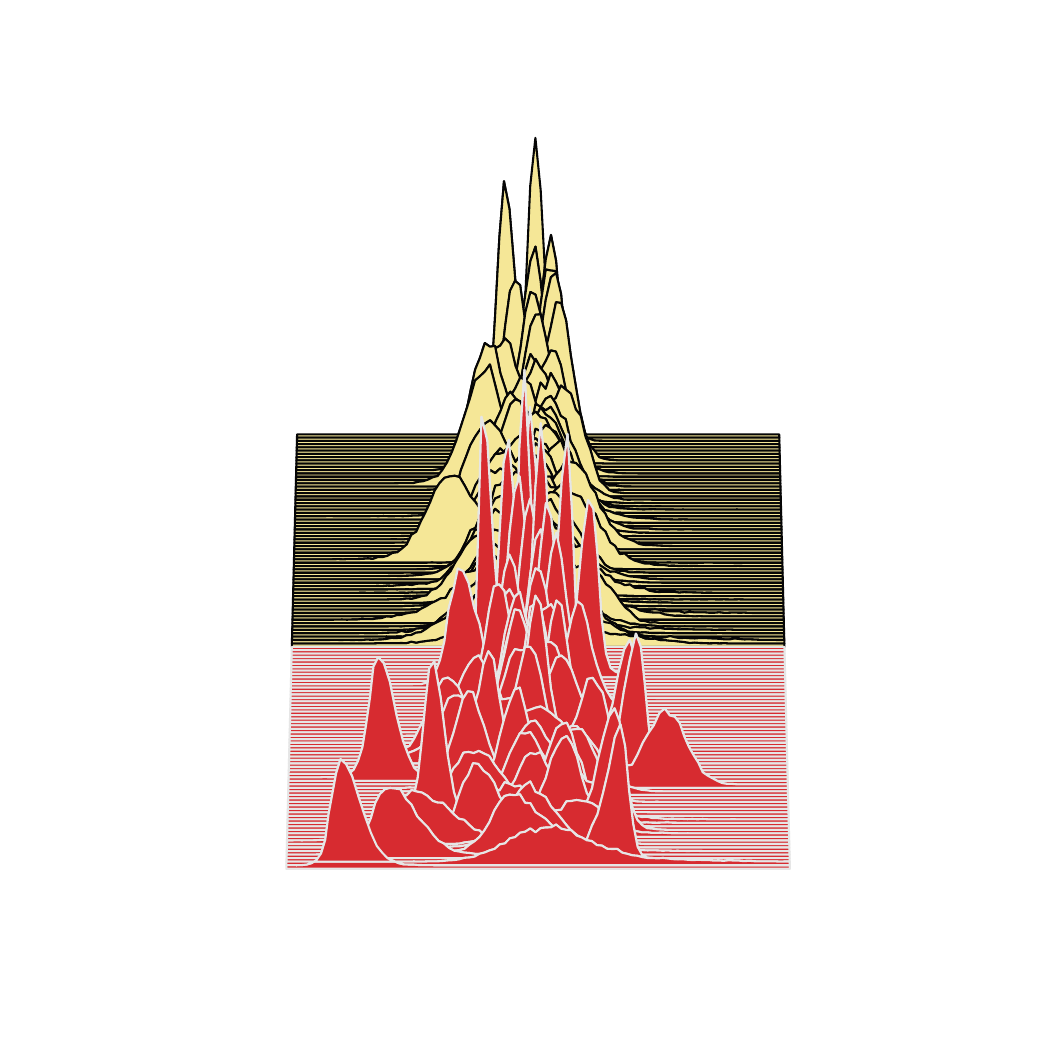} &
\includegraphics[trim=55 55 55 55, clip, keepaspectratio=false, width=222mm, height=111mm]{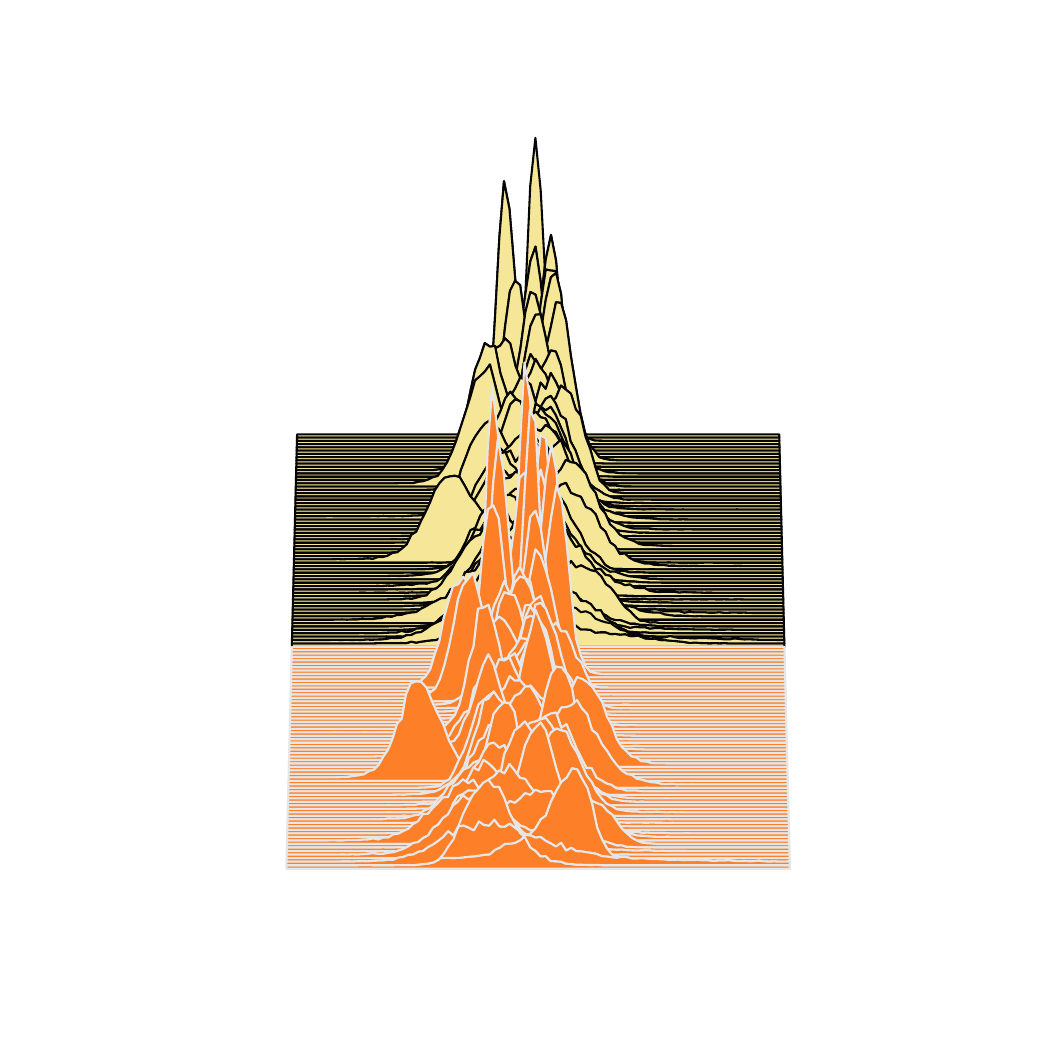} &
\includegraphics[trim=55 55 55 55, clip, keepaspectratio=false, width=222mm, height=111mm]{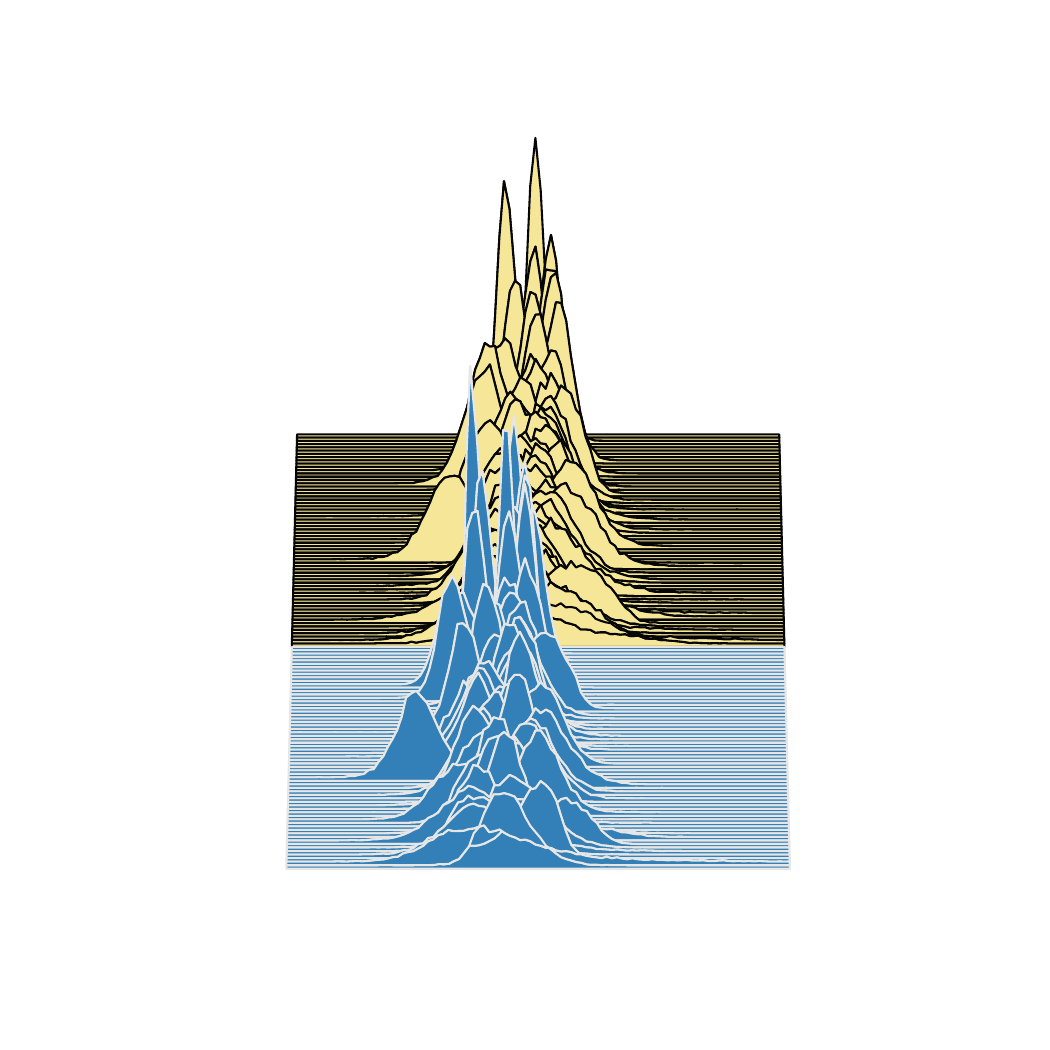} &
\includegraphics[trim=55 55 55 55, clip, keepaspectratio=false, width=222mm, height=111mm]{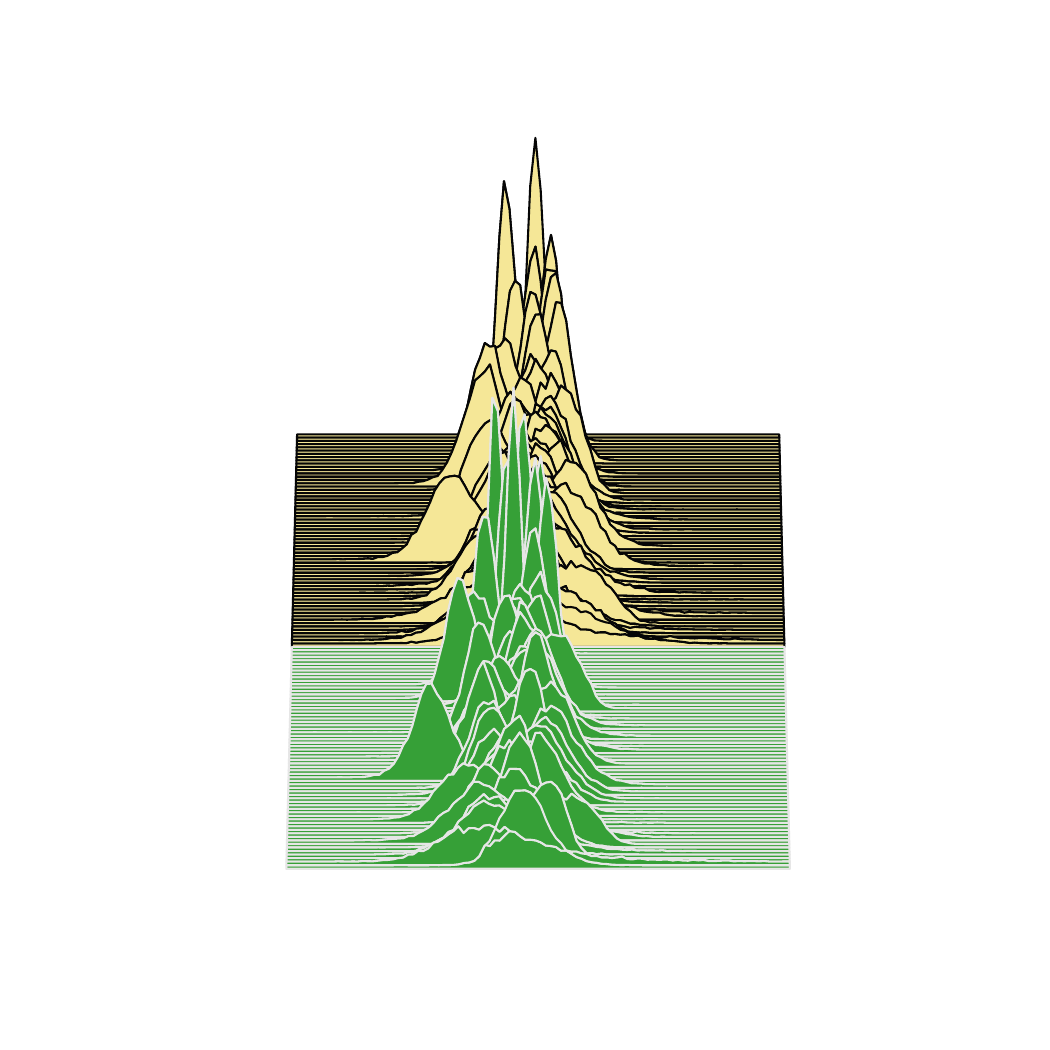} \\
\resizebox{0.12\textwidth}{!}{\rotatebox{90}{ Layer 20}} &
\includegraphics[trim=55 55 55 55, clip, keepaspectratio=false, width=222mm, height=111mm]{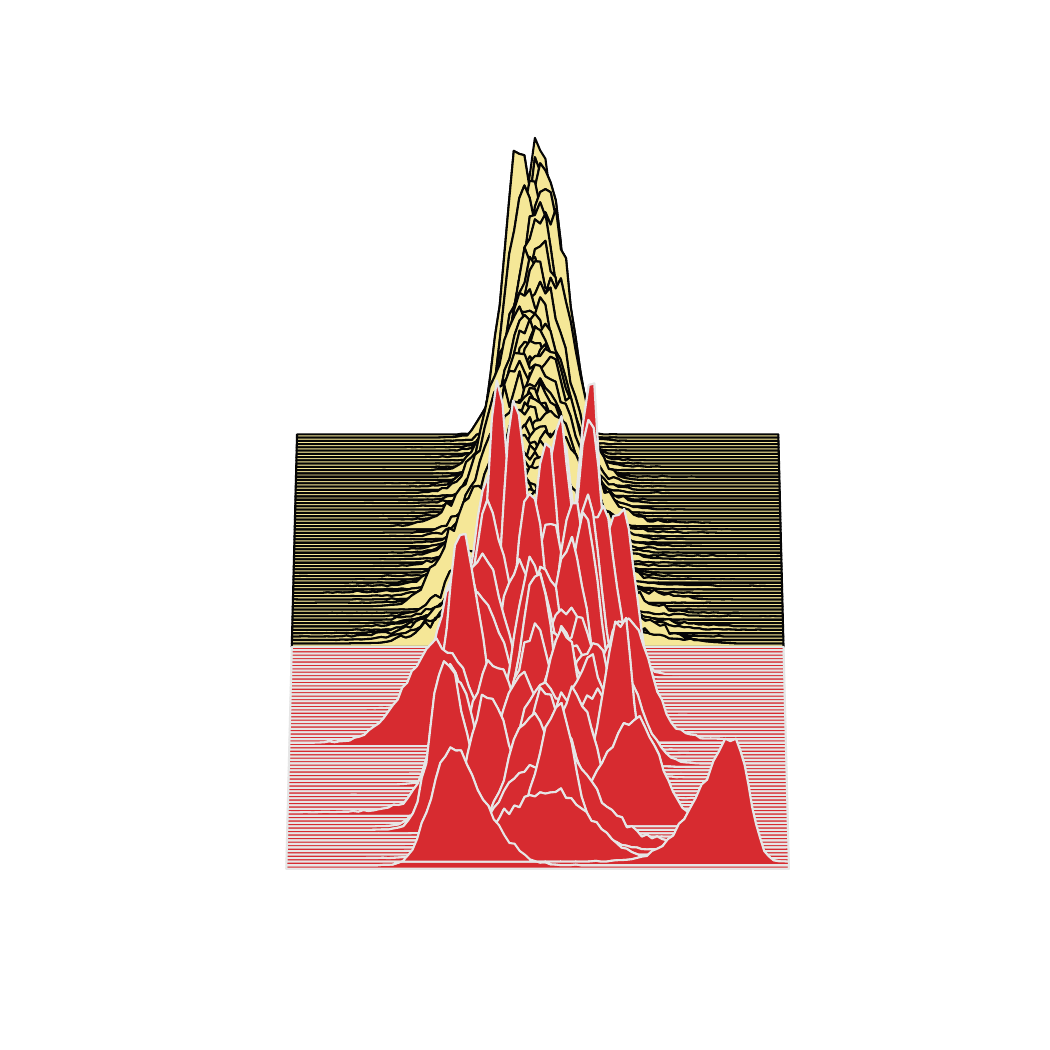} &
\includegraphics[trim=55 55 55 55, clip, keepaspectratio=false, width=222mm, height=111mm]{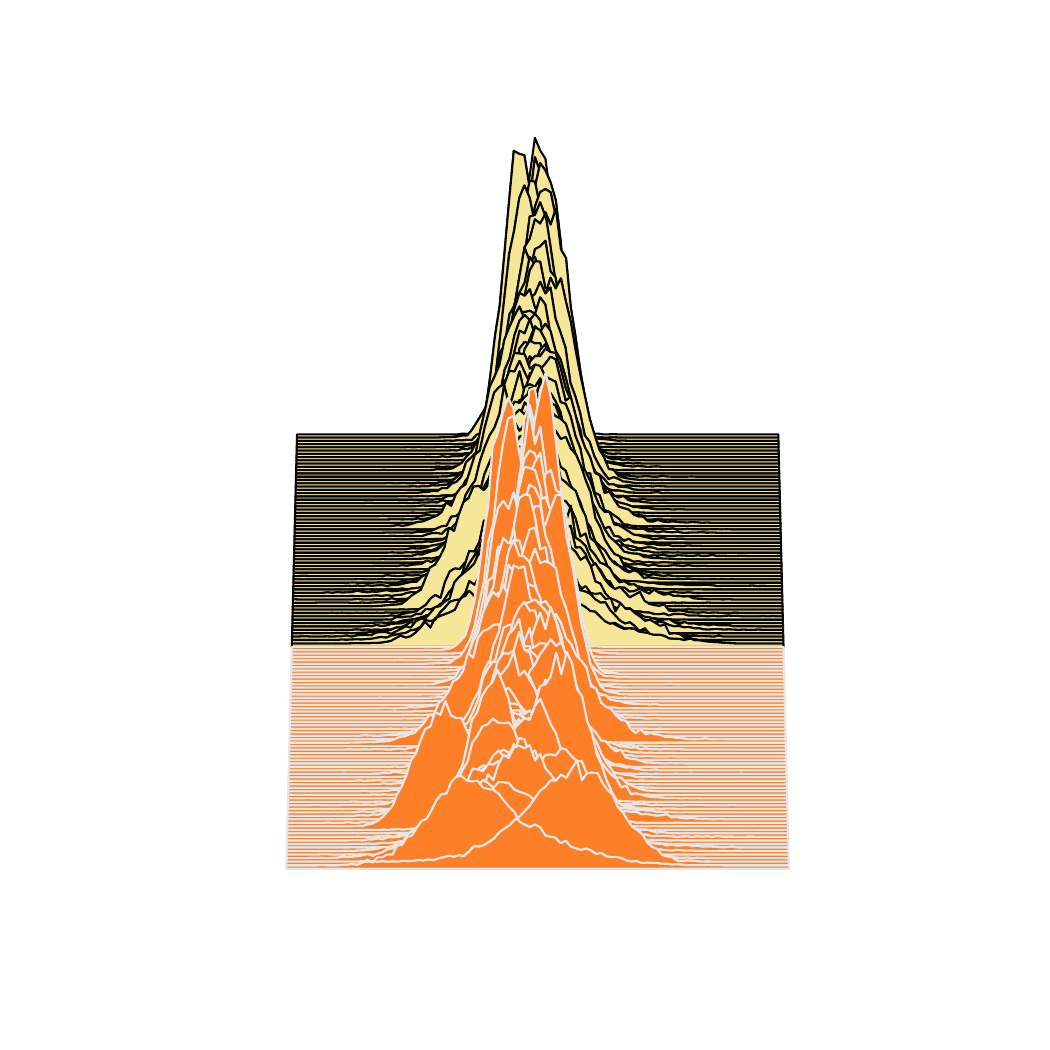} &
\includegraphics[trim=55 55 55 55, clip, keepaspectratio=false, width=222mm, height=111mm]{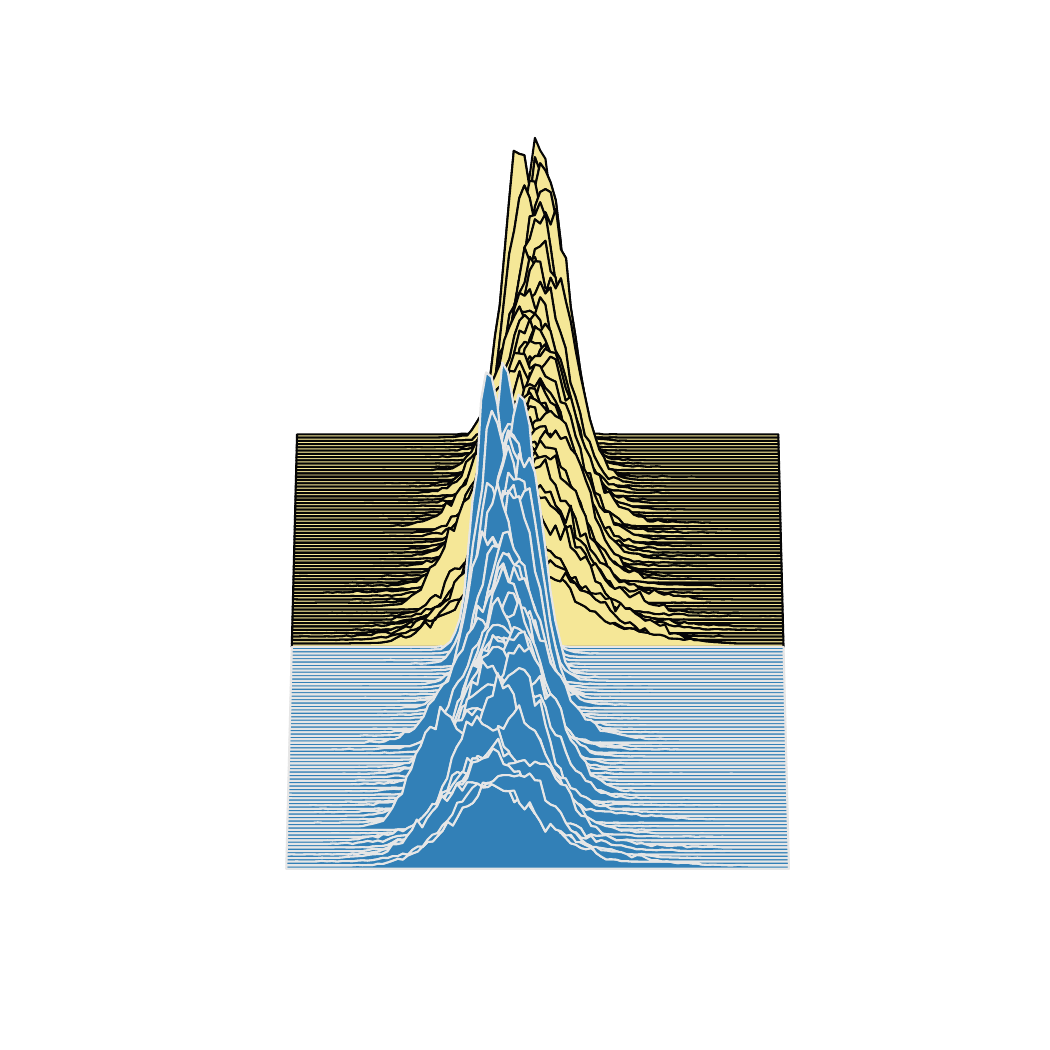} &
\includegraphics[trim=55 55 55 55, clip, keepaspectratio=false, width=222mm, height=111mm]{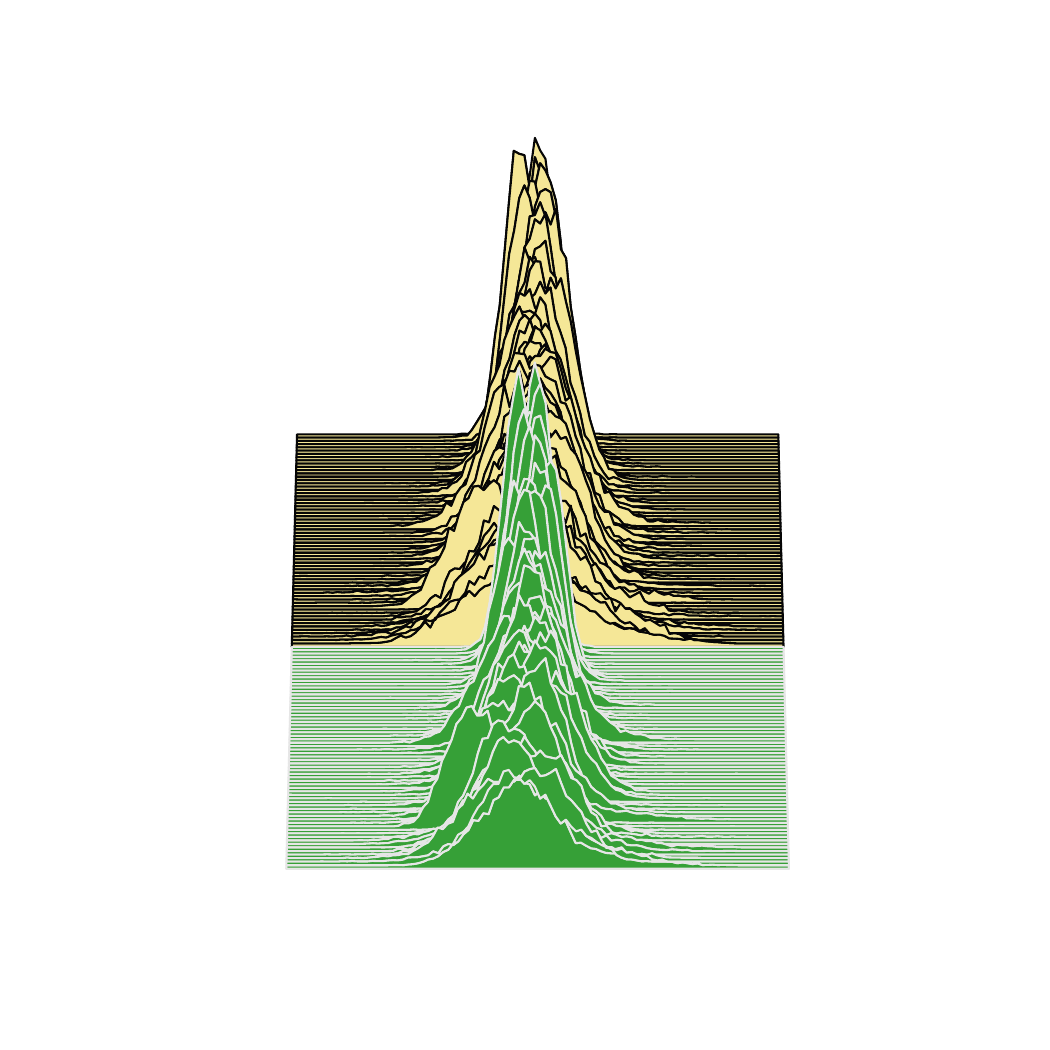} \\
\resizebox{0.12\textwidth}{!}{\rotatebox{90}{ Layer 23}} &
\includegraphics[trim=55 55 55 55, clip, keepaspectratio=false, width=222mm, height=111mm]{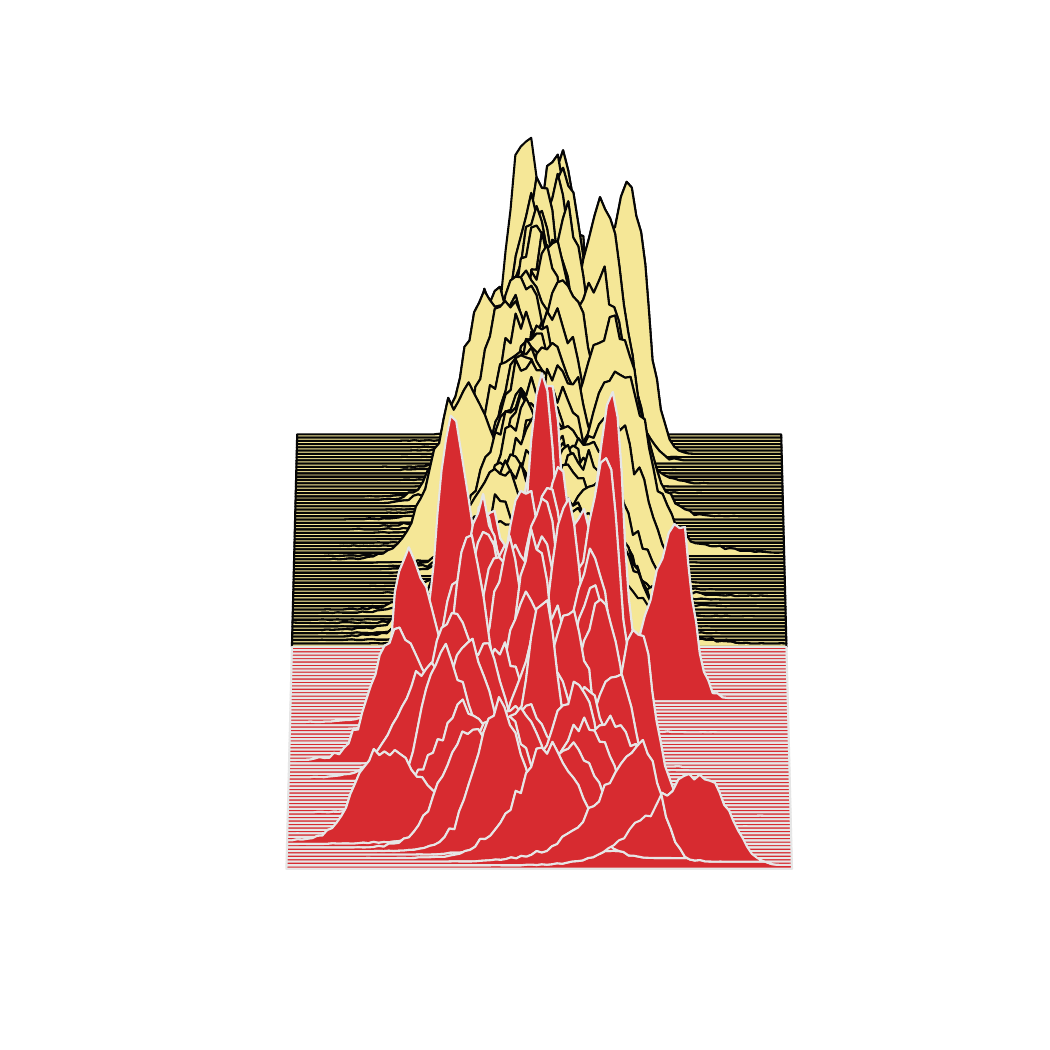} &
\includegraphics[trim=55 55 55 55, clip, keepaspectratio=false, width=222mm, height=111mm]{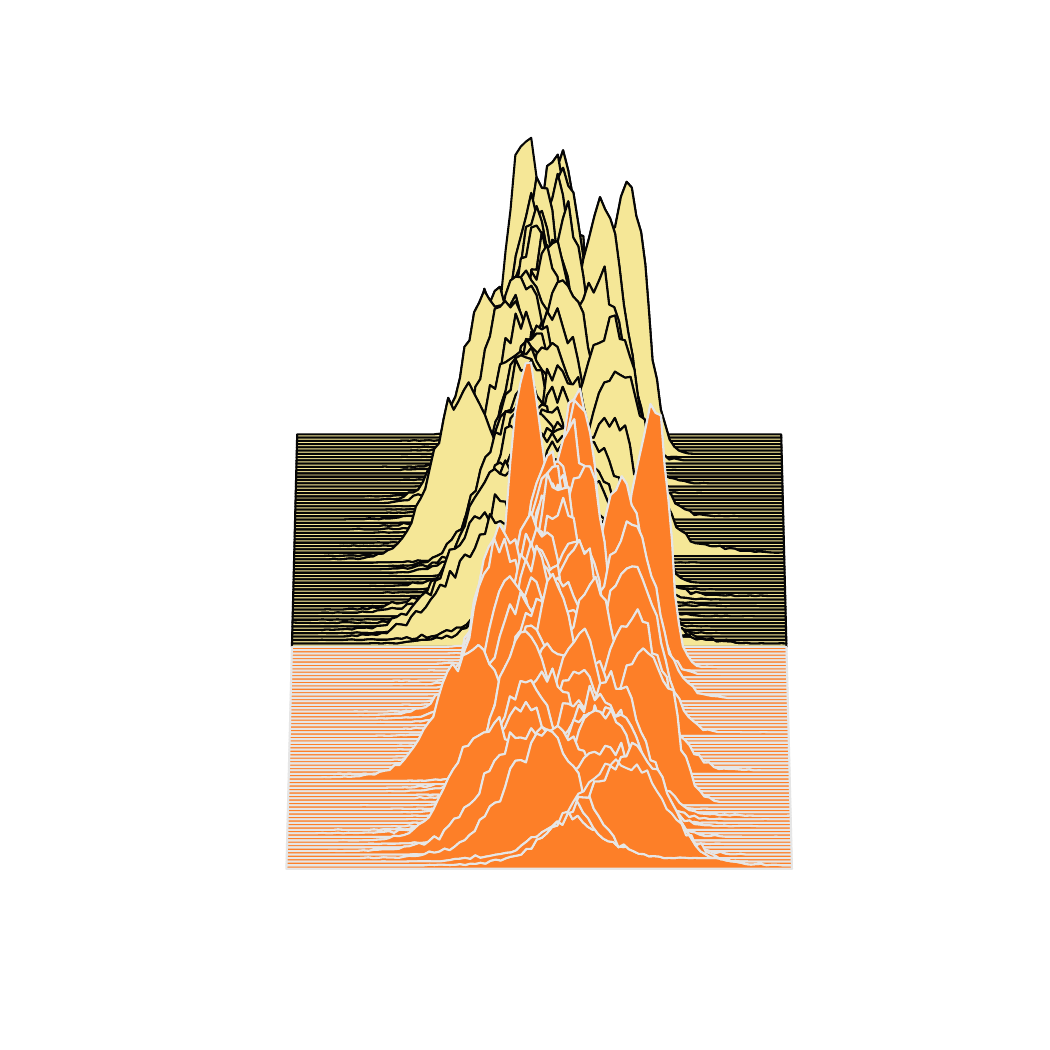} &
\includegraphics[trim=55 55 55 55, clip, keepaspectratio=false, width=222mm, height=111mm]{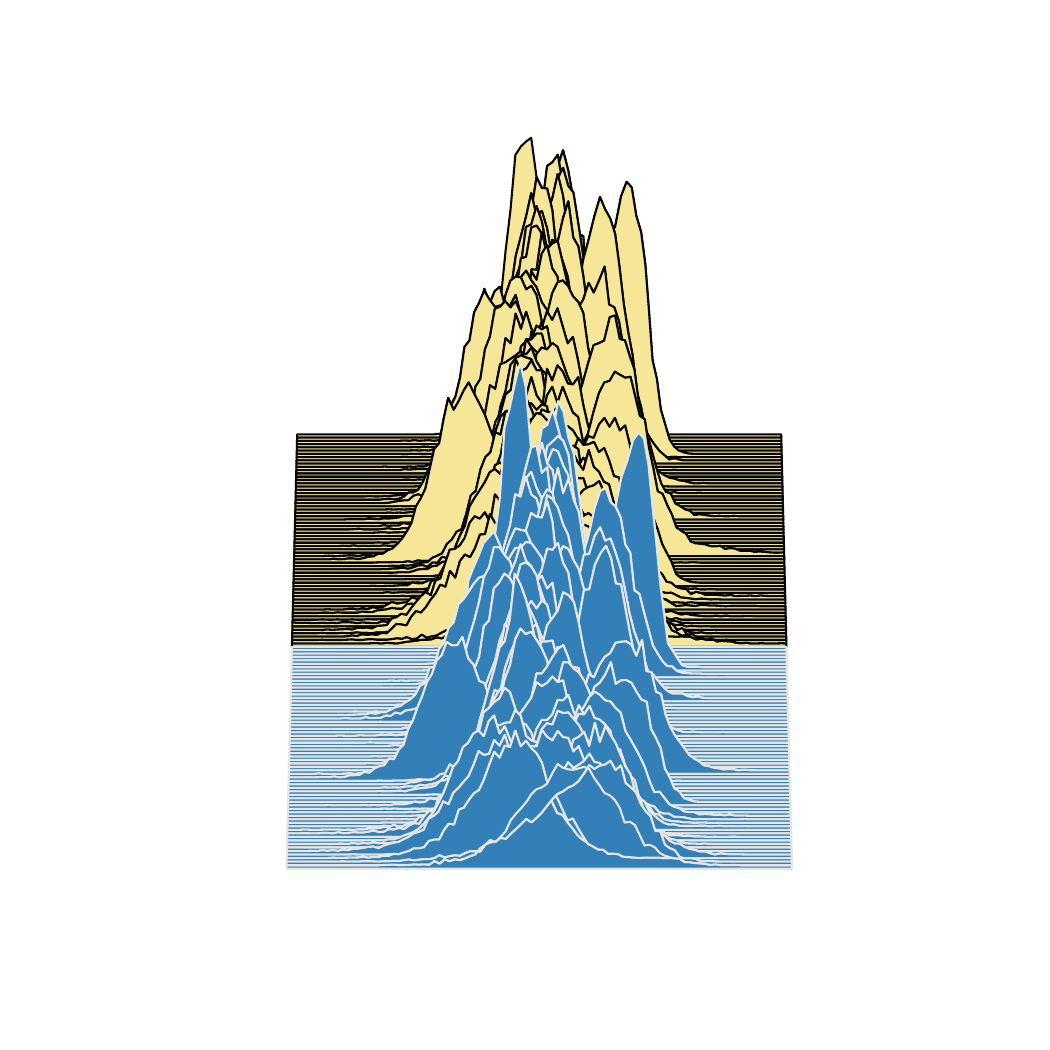} &
\includegraphics[trim=55 55 55 55, clip, keepaspectratio=false, width=222mm, height=111mm]{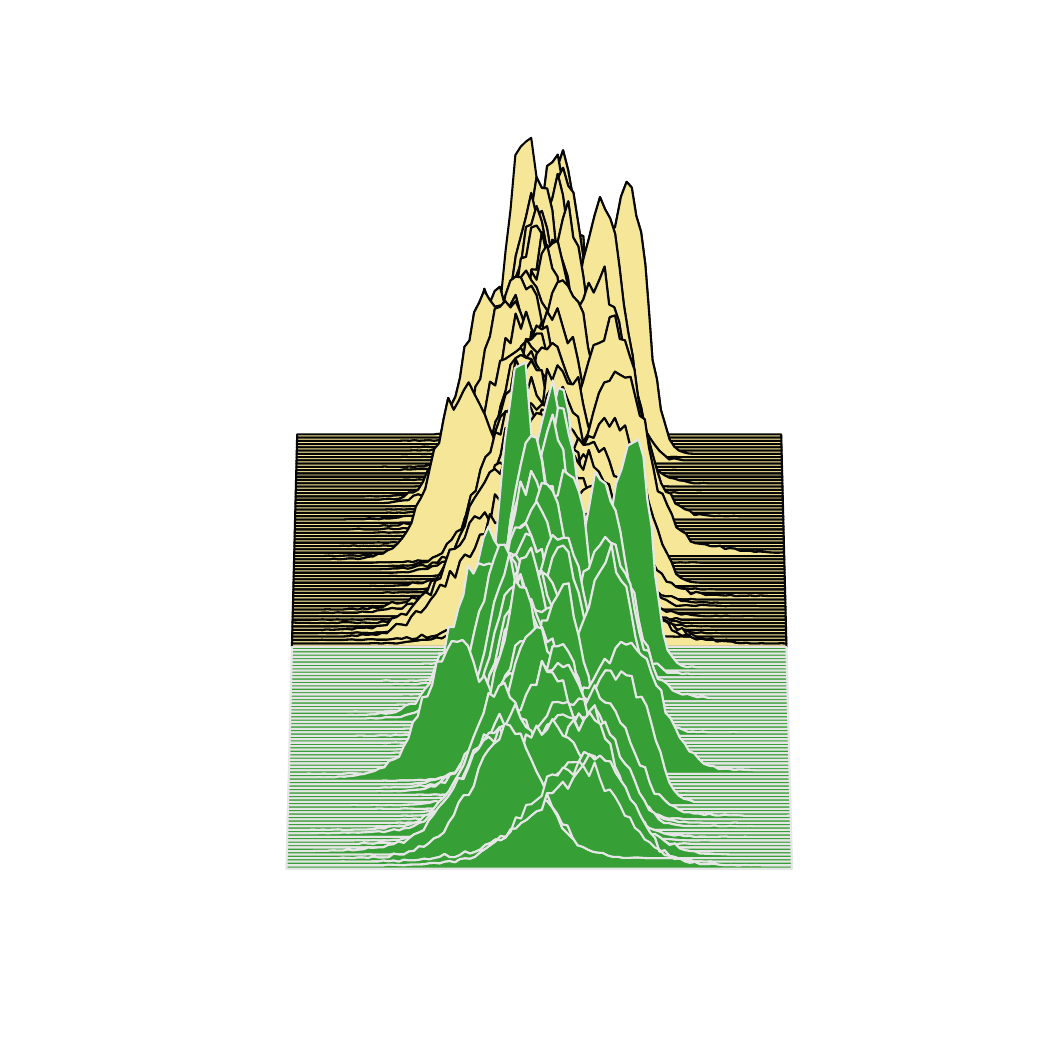} \\
\resizebox{0.12\textwidth}{!}{\rotatebox{90}{ Layer 26}} &
\includegraphics[trim=55 55 55 55, clip, keepaspectratio=false, width=222mm, height=111mm]{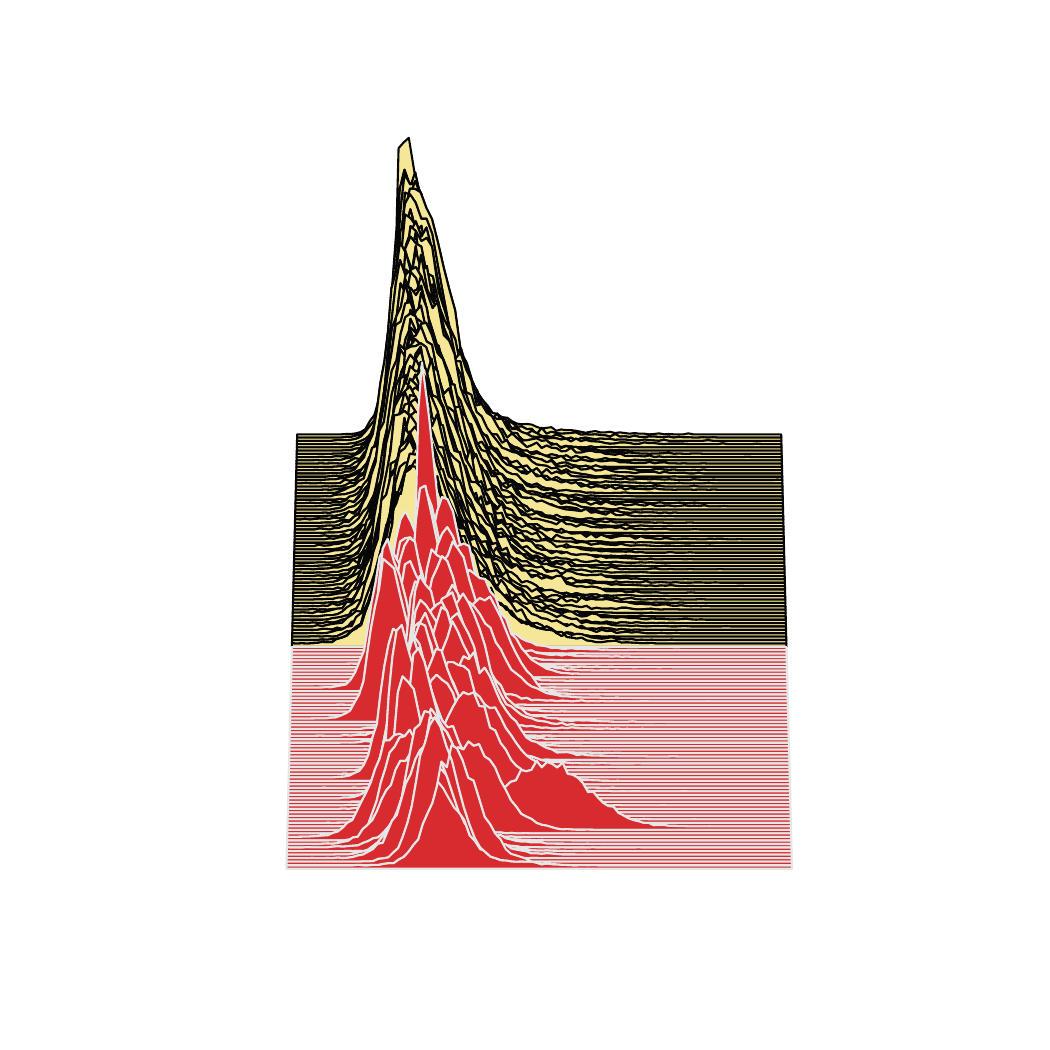} &
\includegraphics[trim=55 55 55 55, clip, keepaspectratio=false, width=222mm, height=111mm]{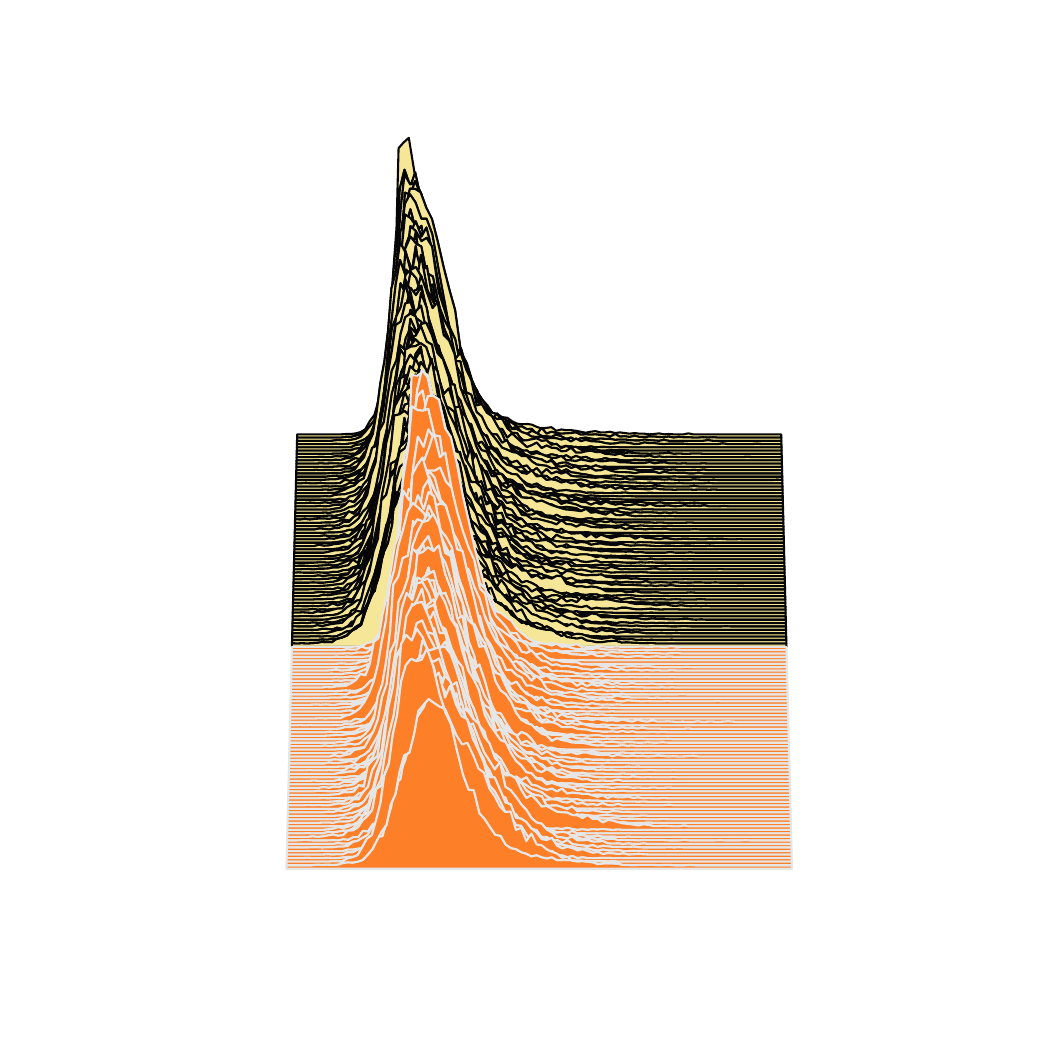} &
\includegraphics[trim=55 55 55 55, clip, keepaspectratio=false, width=222mm, height=111mm]{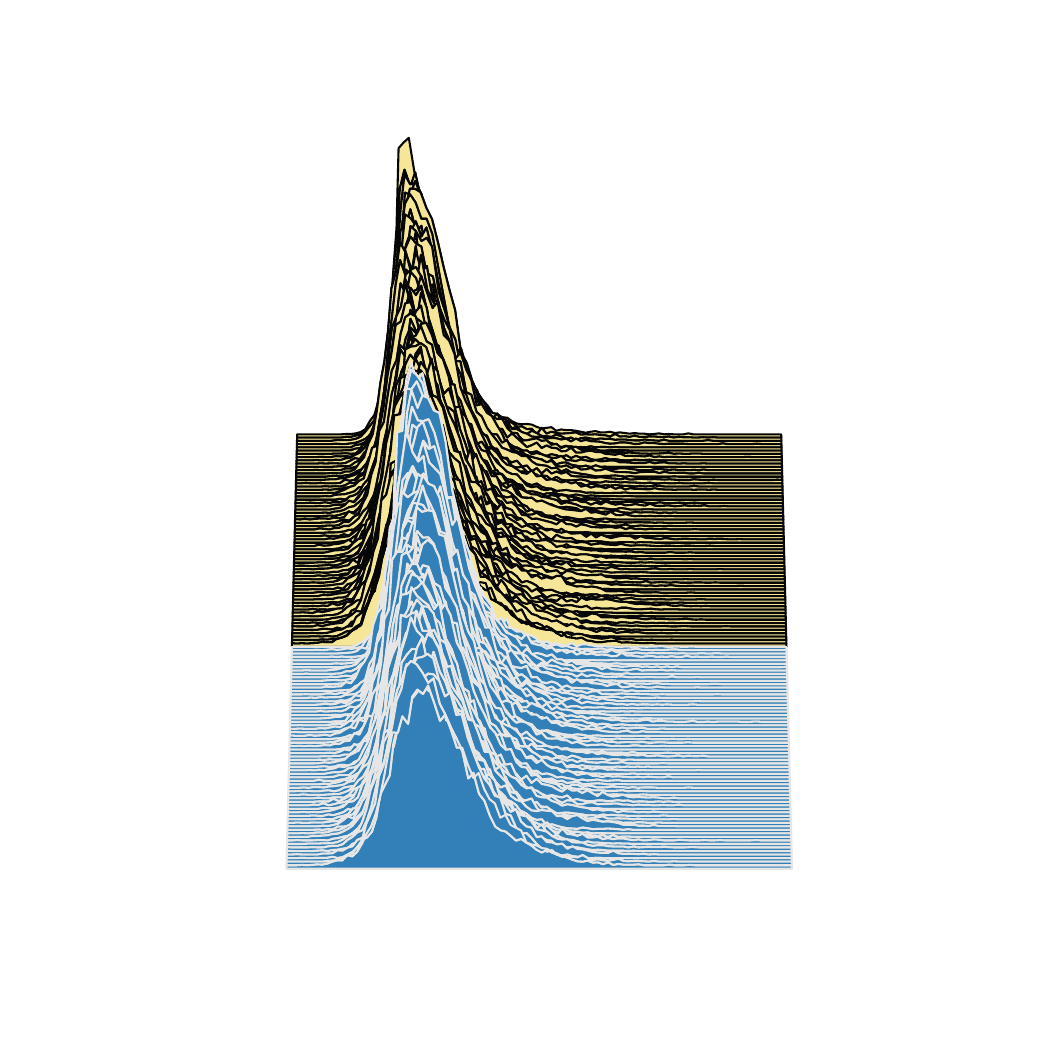} &
\includegraphics[trim=55 55 55 55, clip, keepaspectratio=false, width=222mm, height=111mm]{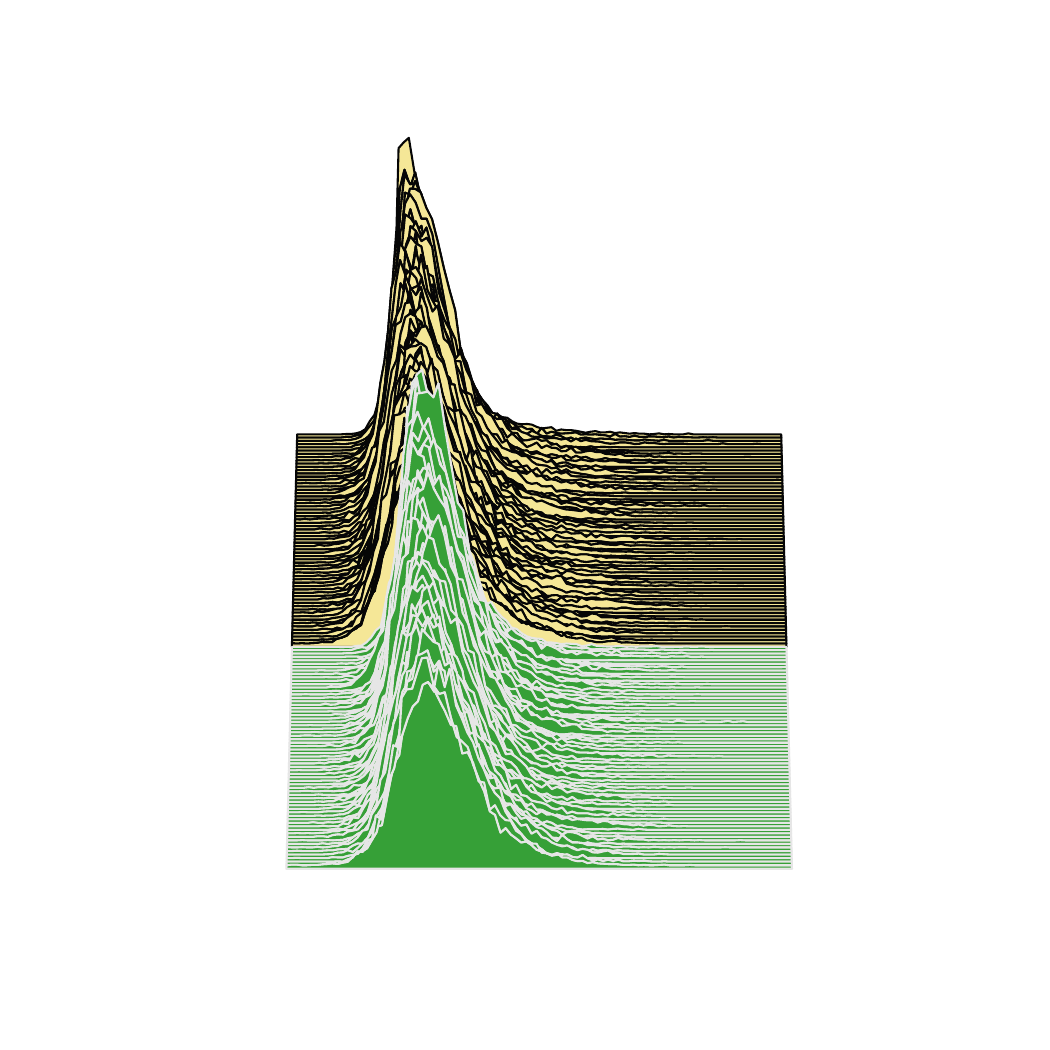} \\
\end{tabular}}
\caption{%
Adapted features on CIFAR-100-C with Gaussian noise (front) and reference features without corruption (back).
Corruption shifts the source features from the reference.
BN shifts the features back to be more like the reference.
Tent shifts features to be less like the reference, and more like an oracle that optimizes on target labels.
}
\label{fig:more-features}
\end{figure}

\end{document}